\documentclass[twocolumn]{article}
\date{}
\usepackage[letterpaper, margin=0.5in]{geometry}
\usepackage{authblk}
\usepackage{credits}
\usepackage{hyperref} 
\hypersetup{colorlinks=true,urlcolor=blue,linkcolor=blue} 
\usepackage{amsmath} 
\usepackage{amssymb} 
\usepackage[rightcaption]{sidecap} 
\usepackage[T1]{fontenc} 
\usepackage{lmodern} 
\usepackage{titlesec}
\usepackage{abstract} 
\usepackage{caption} 
\usepackage[
    backend = biber, 
    style   = numeric-comp, 
    sorting = none,         
    url     = false         
]{biblatex}
\DeclareCaptionFormat{arXivformat}{\normalfont\sffamily\scriptsize\selectfont\textbf{#1#2}#3}
\title{\vspace{-1.0em}Dissociating Artificial Intelligence from Artificial Consciousness \vspace{-0.5in}}
\newcommand{\titlefont}{\fontfamily{lmss}\bfseries\fontsize{32pt}{38pt}\selectfont}

\author[a,1]{Graham Findlay}
\author[c,1]{William Marshall}
\author[a]{Larissa Albantakis}
\author[a,b]{Isaac David}
\author[a]{William GP Mayner}
\author[d]{Christof Koch}
\author[a,2]{Giulio Tononi}

\affil[a]{Department of Psychiatry, University of Wisconsin, Madison, Wisconsin, United States of America}
\affil[b]{Neuroscience Training Program, University of Wisconsin, Madison, Wisconsin, United States of America}
\affil[c]{Department of Mathematics and Statistics, Brock University, St. Catharines, Ontario, Canada}
\affil[d]{Allen Institute, Seattle, WA, United States of America}
\affil[1]{G.F.(Graham Findlay) and W.M. (William Marshall) contributed equally to this work.}
\affil[2]{Corresponding author: \href{gtononi@wisc.edu}{gtononi@wisc.edu}}

\credit{GF}{1,0,1,0,1,1,0,0,1,0,0,1,1,1}
\credit{WM}{1,0,1,0,1,1,0,0,1,1,0,0,1,1}
\credit{LA}{1,0,0,0,0,1,0,0,0,1,0,0,0,1}
\credit{ID}{0,0,1,0,1,0,0,0,1,0,0,0,0,1}
\credit{WGPM}{0,0,0,0,0,0,0,0,1,0,0,0,0,0}
\credit{CK}{1,0,0,0,0,0,0,0,0,0,0,1,0,1}
\credit{GT}{1,0,0,1,0,1,1,1,0,1,0,1,1,1}

\titleformat{\section}
  {\Large\sffamily\bfseries}
  {\thesection.}
  {0.5em}
  {}
\titleformat{\subsection}[runin]
  {\sffamily\bfseries}
  {\thesubsection.}
  {0.5em}
  {}
  [.]
\titleformat{\subsubsection}[runin]
  {\sffamily\small\bfseries\itshape}
  {\thesubsubsection.}
  {0.5em}
  {}
  [.]

\titleclass{\minisection}{straight}[\section]
\titleformat{\minisection}[runin]
   {\sffamily\small\bfseries}
   {\theminisection.}
   {0.5em}
   {}
   [.]
\titlespacing{\minisection}
	{0pt}
	{2ex}
	{2ex}

\newcommand{\ts}{\textsuperscript}

\usepackage{amsthm} 
\usepackage{makecell} 
\usepackage{comment} 
\usepackage{bm} 
\usepackage{mathtools} 
\usepackage{graphicx} 

\DeclareMathOperator{\ii}{\mathrm{ii}}
\newcommand{\D}{D}
\newcommand{\T}{\mathcal{T}}
\newcommand{\inp}{\leftarrow}
\newcommand{\out}{\rightarrow}
\renewcommand{\sc}{s}
\newcommand{\pospart}[1]{\left\vert\,{#1}\,\right\vert_+}

\DeclareMathOperator{\WCC}{WCC}
\DeclareMathOperator{\SCC}{SCC}
\newcommand{\REG}[2]{\mathrm{R}_{#1}^\mathrm{#2}}
\newcommand{\PRG}[2]{\mathrm{P}_{#2}^{#1}}
\newcommand{\IR}[1]{\mathrm{IR}_{#1}}
\newcommand{\MUX}[1]{\mathrm{M}_{#1}}
\newcommand{\MO}{\mathrm{MO}}
\newcommand{\BUF}[1]{\mathrm{B}_{#1}}
\newcommand{\CO}{\mathrm{C}_0}
\newcommand{\CX}[1]{\mathrm{X}_{#1}}
\newcommand{\CA}[1]{\mathrm{A}_{#1}}
\DeclareMathOperator{\ISO}{ISOLATED}
\DeclareMathOperator{\NONT}{NON-T}

\newtheorem{claim}{Claim}
\theoremstyle{definition}
\newtheorem{definition}{Definition}
\newtheorem*{remark}{Remark} 

\begin{document}
\twocolumn[ 
    \begin{@twocolumnfalse} 
        {\titlefont \maketitle \vspace{-1in}} 
        
        \begin{abstract}
        \normalfont\sffamily\bfseries\fontsize{10}{12}\selectfont 
        Developments in machine learning and computing power suggest that artificial general intelligence is within reach. This raises the question of artificial consciousness: if a computer were to be functionally equivalent to a human, being able to do all we do, would it experience sights, sounds, and thoughts, as we do when we are conscious? Answering this question in a principled manner can only be done on the basis of a theory of consciousness that is grounded in phenomenology and that states the necessary and sufficient conditions for any system, evolved or engineered, to support subjective experience. Here we employ Integrated Information Theory (IIT), which provides principled tools to determine whether a system is conscious, to what degree, and the content of its experience. We consider pairs of systems constituted of simple Boolean units, one of which---a basic stored-program computer---simulates the other with full functional equivalence. By applying the principles of IIT, we demonstrate that (i) two systems can be functionally equivalent without being phenomenally equivalent, and (ii) that this conclusion is not dependent on the simulated system's function. We further demonstrate that, according to IIT, it is possible for a digital computer to simulate our behavior, possibly even by simulating the neurons in our brain, without replicating our experience. This contrasts sharply with computational functionalism, the thesis that performing computations of the right kind is necessary and sufficient for consciousness.
        \end{abstract}
    \end{@twocolumnfalse} 
] 

Artificial intelligence (AI) is progressing rapidly, fueled by theoretical and hardware developments in machine learning \cite{LeCun2015, Schmidhuber2022, Bengio2024}. Though large language models such as ChatGPT and vision-language-action models (implementing robotically embodied AI) still lack several human capabilities, they perform spectacularly on a wide array of tasks, demonstrating abilities that were once the stuff of science fiction \cite{Brohan2023, Srivastava2023, Bubeck2023}. Consequently, a majority of AI researchers expect that artificial general intelligence is within reach, leading to systems that perform at least as well as humans in most and, eventually, in all cognitive domains \cite{Grace2024}. Such prospects raise the following question: would machines with human-like intelligence also have human-like consciousness? More precisely, would they have our subjective experiences---``what it is like'' to perceive a scene, endure a pain, or entertain a thought in one's mind \cite{Nagel1974}? Because of the torrential growth of the size and power of AI systems, these questions are shifting from purely hypothetical to practically significant: how we interact with AI, and what rights we accord to them, will hinge on whether we consider them as conscious entities who share our subjective experiences, including pleasure and pain \cite{Long2024, Metzinger2021, Chalmers2022, Shevlin2024}.

When Alan Turing proposed what is now known as the Turing test in 1950, he emphasized that his ``imitation game'' was a test of intelligence, not of consciousness \cite{Turing1950}. Nevertheless, many have since embraced computational functionalism: ``the thesis that performing computations of the right kind is necessary and sufficient for consciousness'' \cite{Putnam1967, Butlin2023}. So far, there is little agreement on what constitutes a computation ``of the right kind.'' Most functionalists seem to agree that if computers were to replicate our cognitive functions in every respect, they would also enjoy our subjective experiences. Others propose stricter definitions of functional equivalence: to be conscious like us, computers should simulate not only our cognitive functions but the detailed causal interactions that occur in our brain---say at the level of individual neurons \cite{Clark1989}.


But is functional equivalence really sufficient for phenomenal equivalence? To address this question, it is critical to rely on a general theory of consciousness, one that states the necessary and sufficient conditions for any system, evolved or engineered, to support subjective experience \cite{Ellia2021}. In this paper, we address the issue of computer consciousness on the basis of Integrated Information Theory (IIT) \cite{Albantakis2023, Tononi2016, Hendren2024}, which offers a principled explanation and a mathematical framework to determine whether and how a system is conscious. 

Crucially, IIT does not start from neural correlates of consciousness, such as neural activity patterns \cite{Lamme2010, Koch2016, Crick1990}, nor from cognitive functions often associated with consciousness, such as broadcasting or monitoring information, or adjusting an inferential model of the world and our actions \cite{Dehaene2018, Graziano2015, Rosenthal2008, Fleming2020, Wiese2021}. Instead, IIT starts with the essential properties of consciousness itself---those that are irrefutably true of every conceivable experience. These are IIT's five \emph{axioms of phenomenal existence}: 
every experience is intrinsic (for itself), specific (this one), unitary (a whole, irreducible to separate experiences), definite (this whole, containing all it contains, neither less nor more), and structured (composed of distinctions bound by relations that make it feel the way it feels).

The theory then formulates these five essential phenomenal properties in terms of corresponding causal properties---IIT's five \emph{postulates of physical existence}---which a substrate must have to support consciousness. These five postulates---intrinsicality, information, integration, exclusion, and composition---can be formulated mathematically and assessed algorithmically for any substrate, given the system's state and complete causal model (see \hyperref[subsec:causal_models]{Causal models} below) \cite{Albantakis2023, Hendren2024}. The analysis identifies systems that can support consciousness, called \emph{complexes}. The causal powers of a complex are then fully \emph{unfolded}, yielding a \emph{cause--effect structure} \cite{Albantakis2023, Hendren2024}. Finally, IIT claims that the quality of an experience---``what it is like to be'' in a specific phenomenal state---is fully accounted for, with no additional ingredients, by that complex's cause--effect structure \cite{Albantakis2023, Hendren2024}.

IIT is supported by growing scientific evidence about our own consciousness. For example, IIT provides a self-consistent account of why the cerebral cortex is important for consciousness but the cerebellum is not, despite having four times more neurons; why consciousness fades in deep slow-wave sleep, even though cortical neurons remain active; and why consciousness might have evolved \cite{Tononi2016}. IIT also makes several experimental predictions that have been validated in humans \cite{Tononi2016, Sasai2016, Haun2017}. Measures reflecting integrated information break down when consciousness vanishes in deep sleep, under general anesthesia, or after widespread brain damage, and recover during dreaming sleep, dreaming under ketamine anesthesia, or after recovery from brain injury \cite{Casarotto2016}. Indeed, a crude-but-practical proxy for IIT's integrated information measure, explicitly designed to test the theory's core tenets, has been successfully employed to detect covert consciousness in behaviorally unresponsive patients \cite{Edlow2023, Fecchio2023}.

To demonstrate that functional equivalence does not imply equivalence of cause--effect structure, and therefore does not imply equivalence of phenomenology, we first apply IIT's mathematical framework to a simple target system (the \textit{simulandum}) and to a computer that simulates it (the \textit{simulans}). We show that even though the computer is functionally equivalent to the target system, it is not a complex and fails to replicate the target system's cause--effect structure. Furthermore, we show that treating the computer's units and states at \emph{macro grains} \cite{Marshall2024} does not change this conclusion. We extend these results to non-trivial cases, including a Turing-complete version of the computer that can simulate arbitrary systems with arbitrary functions. It follows that such a computer could, for example, simulate a human brain in microphysical detail, without replicating its experiences. 
Our results challenge the default, dominant assumption of computational functionalism: that traditional computer systems replicating our cognitive functions or simulating the causal interactions occurring within our brain, will necessarily replicate our experiences, or have substantial experience of any kind. 

\section*{Theory}

This section provides a high-level overview of IIT and highlights aspects of the theory that are relevant to the results presented here. For a more detailed and complete exposition of IIT and its justification, see \cite{Albantakis2023, Hendren2024}.

\subsection*{Causal models}
\label{subsec:causal_models}

The starting point of IIT's analysis is a causal model of a physical system that captures all the interactions within it. Here, ``physical'' is meant operationally, indicating that the individual units can be manipulated and observed. The causal model is constituted of \emph{micro units}, each having two internal states as well as inputs and outputs. The units are micro units in the sense that no finer detail about them or their function is included in the model. Given a causal model, any set of micro units can be considered as a \emph{candidate system}. Any units not included in a candidate system are referred to as that system's \emph{background conditions}. The causal model of a candidate system is fully characterized by its \emph{transition probability matrix (TPM)}, causally conditioned over its background conditions \cite{Marshall2024}. 

Here we consider models with synchronously updating units implementing Boolean logic, which can be thought of as standing for transistors driven by a system-wide clock. While our idealized units ignore most physical details, they are sufficient to make our points about functional and phenomenal equivalence. Practically, the spatiotemporal grain of transistors and their interconnections is unambiguous, well-demarcated from other grain sizes, and is thought to capture the fundamental causal interactions within engineered silicon circuits \cite{McCulloch1943, Neumann1958}. 

\subsection*{Identifying complexes}

According to IIT, a substrate can support consciousness---it is a complex---if and only if it fulfills the five postulates \cite{Albantakis2023, Tononi2016, Hendren2024}. To determine whether a system of units is a complex, one evaluates its \emph{system integrated information} ($\varphi_s$)---the extent to which the system's intrinsic information (capturing the postulates of intrinsicality and information) \cite{Barbosa2020} is affected by its minimum partition into causally independent parts (capturing the integration postulate). To satisfy the exclusion postulate, a candidate system must specify a maximum of integrated information ($\varphi_s^*$) compared to all competing candidate systems with overlapping units and grains. This can be determined by evaluating $\varphi_s$ for every possible set of units within the causal model. Moreover, every system is evaluated at multiple grains, by exhaustively grouping subsets of its micro units into \emph{macro units}, and mapping states of the constituent micro units to states of the resulting macro units. This procedure, called \emph{macroing}, is done because a system's intrinsic causal powers ($\varphi_s$) may be higher at a coarser than at a finer grain, depending on its internal organization \cite{Marshall2018, Marshall2024, Hoel2016}. 

Once the maximum of integrated information within a causal model has been identified, all units of that complex are excluded from participating in other complexes. The search for complexes is then repeated recursively over the remaining units of the system until all non-overlapping complexes are identified.

\subsection*{Unfolding the cause--effect structure of a complex} 

Finally, composition is evaluated by assessing how a complex's causal power is structured. Briefly, \emph{distinctions} capture the causal powers of subsets of units, while \emph{relations} between distinctions characterize how their powers overlap. Distinctions and relations are both associated with measures of their irreducibility ($\varphi_d$ and $\varphi_r$, respectively). Together, the totality of a system's distinctions and relations composes its \emph{cause--effect structure} (also called a $\Phi$\emph{-structure}). The cause--effect structure fully characterizes what a complex in a state specifies about itself through the congruent causes and effects of its various subsets. The process of identifying all of a complex's distinctions and relations, and thereby obtaining its cause--effect structure, is referred to as \emph{unfolding}. According to IIT, the cause--effect structure of a complex in a state accounts for the quality of its experience in full, with no additional ingredients needed (``quality is structure,'' \cite{Albantakis2023, Hendren2024, Haun2019, Comolatti2024}). The quantity of consciousness associated with a cause--effect structure is measured by its \emph{structure integrated information} ($\Phi$)---the total irreducibility of its distinctions and relations ($\sum\varphi_d + \sum\varphi_r$). If the state of a complex changes, its cause--effect structure may change as well, and therefore its quality of consciousness.

\section*{Results}

Below, we apply IIT's causal powers analysis to a target system constituted of four micro units (PQRS) and to a functionally-equivalent four-bit computer, constituted of 117 micro units, able to simulate PQRS indefinitely. We first demonstrate that functional equivalence does not imply equivalence of cause--effect structures at the grain of micro units. We then demonstrate that there is no function-relevant macroing of the computer compliant with IIT's postulates that replicates the cause--effect structure of the target system. Finally, we extend the four-bit computer to be Turing-complete and demonstrate that the previous results do not depend on the complexity of the function being implemented.

\subsection*{A target system (PQRS) and the cause--effect structure it specifies} 

\begin{figure*}[h]
\centering
\includegraphics[width=7.5in]{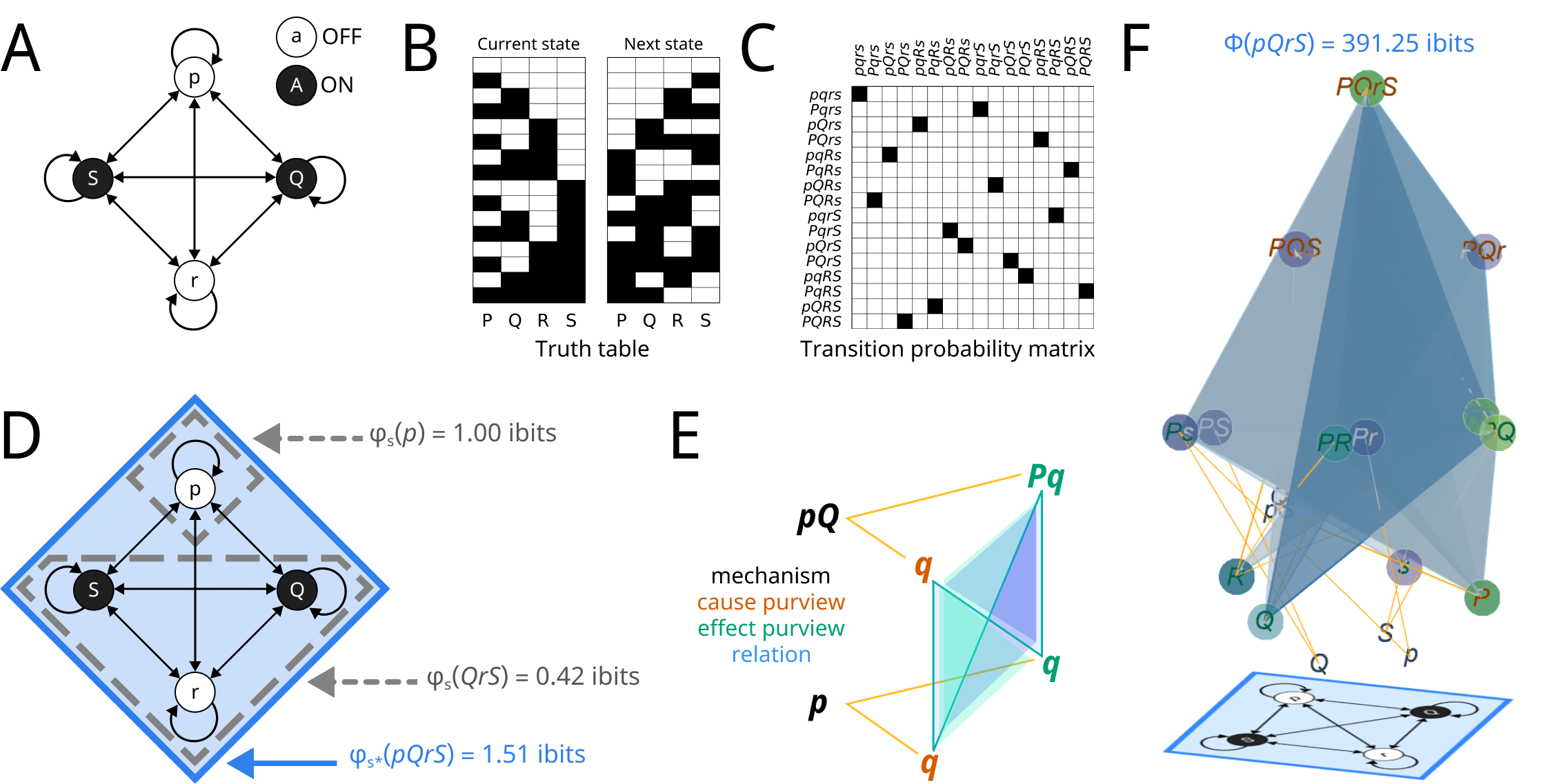}
\caption{\textbf{A target system for simulation.}
(\textbf{A}) PQRS is a substrate of four units with binary states that update synchronously at discrete intervals. The system is fully connected: each unit receives input from every unit, including itself. Here, the system is shown in state 0101, also written as $pQrS$.
(\textbf{B}) The units P, Q, R, and S implement custom boolean functions of four inputs which, despite not resembling familiar logic gates, are described by a truth table. The truth table yields, for any possible state of PQRS (left), what its subsequent state would be (right). 
(\textbf{C}) Since all of PQRS's state transitions are deterministic, this truth table is also equivalent to PQRS's Transition Probability Matrix (TPM; right). 
(\textbf{D}) Applying IIT's postulates to $pQrS$ and its TPM reveals a single, maximally irreducible complex ($\varphi_s(pQrS) = 1.51$ ibits). This complex is, by definition, more irreducible than all other candidate systems over the same substrate, including $QrS$ ($\varphi_s(QrS) = 0.42$ ibits; grey dashes) and $p$ ($\varphi_s(p) = 1.00$ ibits; grey dashes). All other candidate systems (e.g., $pQr$, $pQS$, $prS$, $pQ$, $pr$, etc.; not shown) are fully reducible, with $\varphi_s = 0$. 
(\textbf{E--F}) The unfolded cause--effect structure of $pQrS$ consists of 13 distinctions ($p$, $Q$, $S$, $pQ$, $pS$, $Qr$, $QS$, $rS$, $pQr$, $pQS$, $prS$, $QrS$, $pQrS$) and 8184 relations among these 13 distinctions. Relations bind state-congruent overlaps among the causes and effects of distinctions. Only 2-relations (edges) and 3-relations (triangles) are shown here. For example (\textbf{D}) , mechanism $pQ$ specifies a maximally irreducible cause $P$ and a maximally irreducible effect $Rs$, yielding a distinction. Distinctions $pQ$ and $pS$ jointly specify overlapping and congruent causes or effects (in this case, a cause $P$), yielding a relation. According to IIT, the  cause--effect structure (\textbf{E}) corresponds to the content, or quality, of the experience supported by $pQrS$. The quantity of consciousness is measured by $\Phi$ (here equal to 391.25 ibits), a non-negative number equal to the sum of the irreducibility values ($\varphi$) of all distinctions and relations in its cause--effect structure \cite{Albantakis2023}.}
\label{fig:pqrs_intro}
\end{figure*}

Fig. \ref{fig:pqrs_intro}A shows PQRS, the target system to be simulated, comprising a set of four binary units whose dynamics are defined by a truth table or, more generally, a transition probability matrix (Fig. \ref{fig:pqrs_intro}B--C). For simplicity, we consider PQRS's causal model in isolation, without background conditions. The system PQRS in state 0101 (also written as $pQrS$) satisfies intrinsicality, information, integration, and exclusion, and is thus a complex, with $\varphi_s = 1.51$ \emph{intrinsic bits} (Fig. \ref{fig:pqrs_intro}D), also called \emph{ibits} \cite{Barbosa2020}. Its cause--effect structure, unfolded according to \cite{Albantakis2023, Mayner2018}, is composed of 13 distinctions and 8184 relations and has $\Phi = 391.25$ ibits (Fig. \ref{fig:pqrs_intro}E--F). The system will transition through a repeating cycle of nine states, with varying complexes, cause--effect structures, and $\Phi$ values (Supplementary Fig. \ref{fig:pqrs_cycle}). This minimal system, though it contains > 8,000 relations, was chosen to reduce the computational requirements of analyzing the cause--effect structures specified by the computer simulating it, without loss of generality, as shown below.

\subsection*{A computer constituted of 117 units that can simulate PQRS indefinitely} 

\begin{figure*}[hp]
\centering
\includegraphics[width=6.95in]{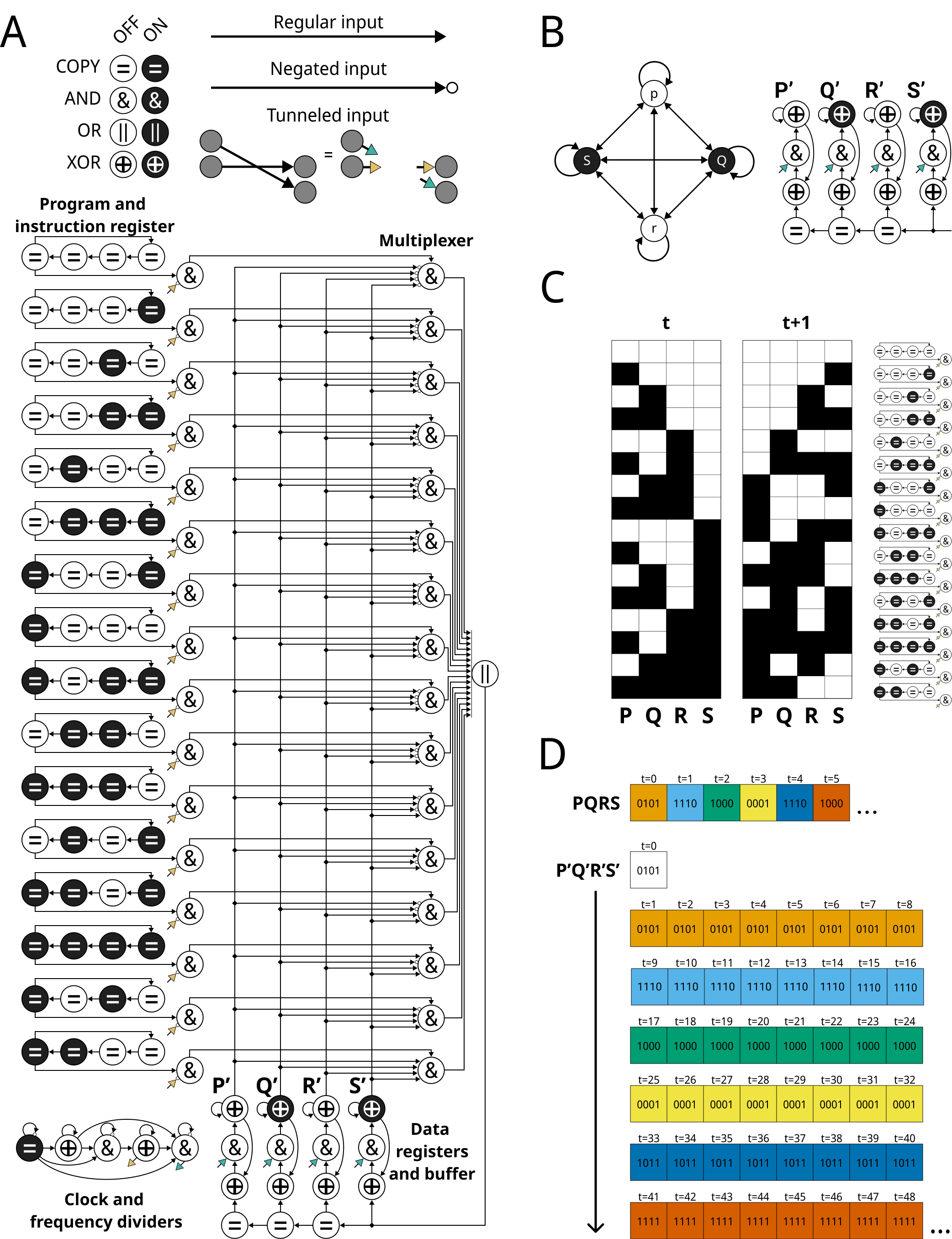}
\caption{\textbf{Four-bit computer that simulates PQRS indefinitely.}
(\textbf{A}) A simple computer can be used to simulate PQRS for an arbitrary number of time steps. The computer comprises a clock with frequency dividers, a program that encodes PQRS's transition rules, four one-bit data registers that store PQRS's state, and a multiplexer-like processing unit. Each of the computer's 117 units implements a Boolean function (COPY, AND, OR, or XOR) over its inputs, and is either OFF (white) or ON (black). To reduce visual clutter, colored arrows (teal or tan) are used in place of black arrows to indicate certain connections. For example, the rightmost AND gate in the clock and frequency dividers outputs directly to each AND gate in the data registers. 
(\textbf{B}) The computer can be programmed by setting the initial state of its data registers P', Q', R', and S' to encode the current state of PQRS. 
(\textbf{C}) The initial states of the program units are set to encode the state transition rules of PQRS. The data that the program operates on are the current states of P', Q', R', and S'.
(\textbf{D}) Each register updates its state every 8\ts{th} timestep, when the computer begins the next iteration of its simulation. For detailed information about how the computer works, see the \hyperref[sec:step_by_step]{``Step-by-step guide to the computer's operation''} supplement.}
\label{fig:computer_intro}
\end{figure*}

Fig. \ref{fig:computer_intro}A shows a basic four-bit computer capable of simulating PQRS. This simple digital computer with a Harvard-like, stored-program architecture has recognizable components: a clock, frequency dividers, program memory, an instruction register, data registers, and a multiplexer that combines instructions with data to serve as a minimal processor. The data registers store the state of the system until instructed to update; the program memory stores the function of the units the computer is simulating; the multiplexer's processing units combine this information about state and function to compute the next state of the target system; the clock synchronizes these processes and tells the registers when to update their state. The clock and program are read-only components of the computer; they output to the data registers and multiplexer but receive no feedback connections. For a detailed explanation of the computer's operation, and for a comparison to common computer architectures, see the \hyperref[sec:step_by_step]{``Step-by-step guide to the computer's operation''} supplement.

When initialized appropriately, this computer simulates the behavior of PQRS. The initialization procedure is simple: First, the states of the program units are set in such a way that they encode PQRS's state transition rules (Fig. \ref{fig:computer_intro}B). Next, the states of the units P', Q', R', and S' are set to reflect the initial state of PQRS (Fig. \ref{fig:computer_intro}C). Finally, the state of the clock is set as shown in Fig. \ref{fig:computer_intro}A. The processing units may be initialized to any state. After eight updates, the states of P', Q', R', and S' reflect a single state transition of P, Q, R, and S (Fig. \ref{fig:computer_intro}D), and the two systems will continue to produce homomorphic sequences of states indefinitely. This is true regardless of the initial states of PQRS and P'Q'R'S', hence the two systems---the 4-unit simulandum and the 117-unit simulans---are functionally equivalent (\textit{modulo} eight updates).

\subsection*{The computer fragments into multiple complexes, none of which specifies a cause--effect structure identical to that of PQRS} 

\begin{figure*}[h]
\includegraphics[width=7.5in]{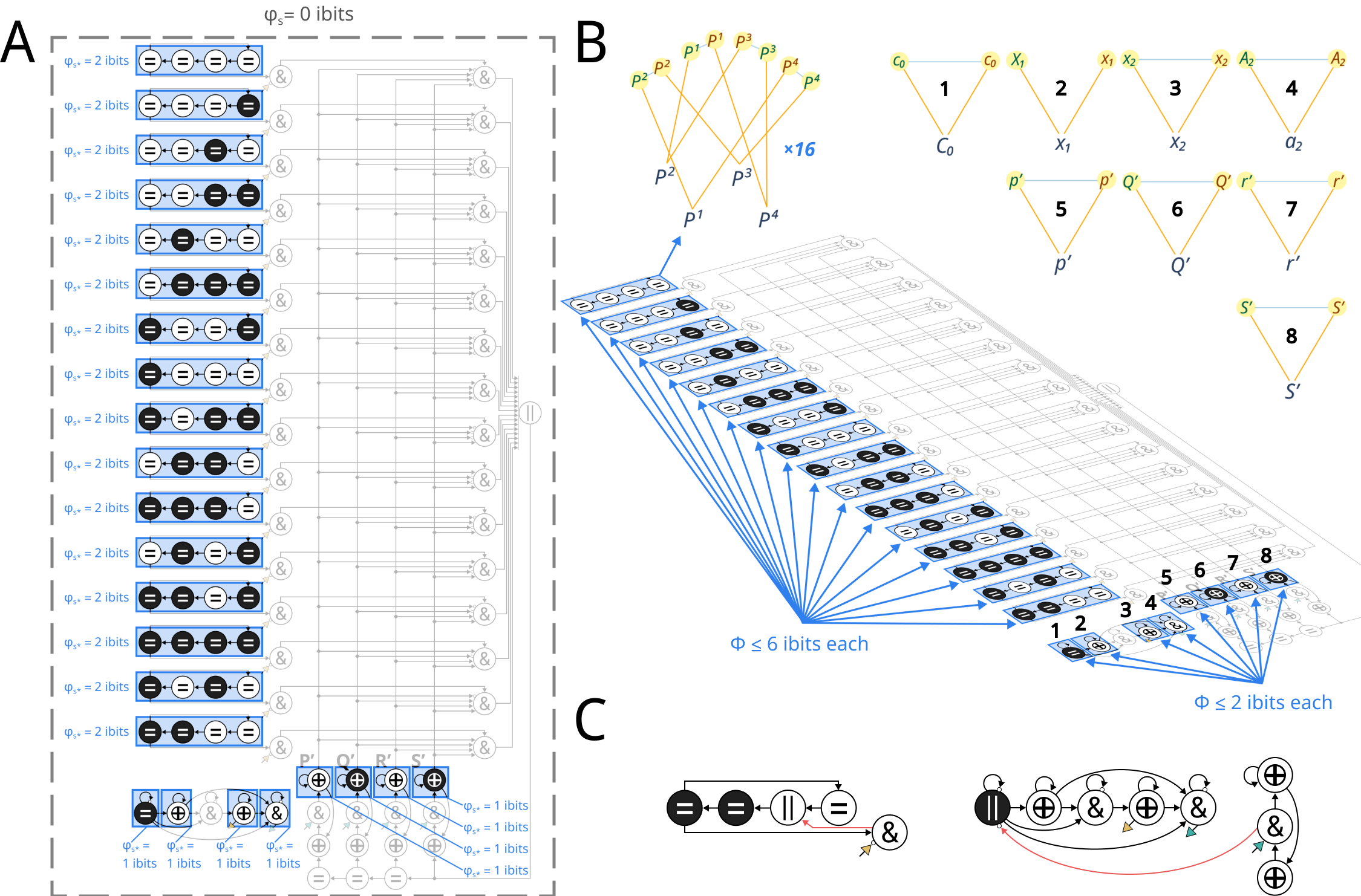}
\caption{\textbf{Identifying the computer's complexes and unfolding their cause--effect structures.}
(\textbf{A}) The computer as a whole is not integrated, regardless of its state ($\varphi_s = 0$ ibits), due in part to the presence of purely feedforward modules. Therefore, the computer is not a complex and has no cause--effect structure. Instead, it fragments into twenty-four disjoint small complexes, each one of which is maximally irreducible (shown in blue). None of these complexes specify a cause--effect structure equivalent to that specified by $pQrS$. 
(\textbf{B}) Shown here for illustrative purposes are the cause--effect structures of nine complexes depicted in panel A. Due to lack of space, the cause--effect structures of fifteen complexes in the program memory are not shown, but they are essentially identical to the one shown (top left). The cause--effect structures of the eight complexes in the clock and data registers are labeled (1-8). All complexes specify cause--effect structures that are simple, with only first-order relations, and not as rich as that of $pQrS$ (see Fig. \ref{fig:pqrs_intro}F). 
(\textbf{C}) Feedback connections can be added to the computer without disrupting its function; the computer's ability to indefinitely simulate any four-unit Boolean system is unimpaired. Each line of the program has a single COPY unit replaced by an OR unit, to which a feedback connection from the instruction register is wired (red; left). Similarly, the clock has a single COPY unit replaced by an OR unit, to which a negated feedback connection from each of the data registers is wired (red; right).}
\label{fig:ces_inequivalence}
\end{figure*}

Does the computer, when programmed to be functionally equivalent to $pQrS$, also replicate its cause--effect structure? By applying IIT's causal powers analysis to the computer as a whole, we find that, unlike $pQrS$, the computer has $\varphi_s = 0$ ibits (Fig. \ref{fig:ces_inequivalence}A, grey). This is because the computer contains modules that are connected to the rest of the system in a purely feedforward manner. This violates the requirement for integration, which can be assessed by partitioning the set of units that constitute a system into separate parts. In an integrated system ($\varphi_s > 0$ ibits), every part will both make a difference to-- and take a difference from the rest of the system \cite{Marshall2023}. 

Most subsets of units also have $\varphi_s = 0$ ibits. In fact, the complete IIT analysis reveals that the computer contains 24 individual complexes, each one constituted of between one and four units, each with $\Phi \leq 6$ ibits (Fig. \ref{fig:ces_inequivalence}A, blue; \hyperref[subsubsec:micro_analysis_of_wcc]{``Micro-grain analysis of the weakly connected computer''} supplement). Thus, the entire computer is never a single, large complex, no matter the state, but merely an aggregate of smaller complexes, each of which has its own tiny cause--effect structure. Importantly, each of the 24 complexes specify cause--effect structures that differ from the one specified by $pQrS$ (Fig. \ref{fig:ces_inequivalence}B). The number of distinctions are smaller (1--4 distinctions, whereas $pQrS$ has 13). The computer has only first-order distinctions (specified by individual units), whereas $pQrS$ also has higher-order distinctions (specified by two or more units). The number of relations is also vastly different (at most four relations for the computer, whereas $pQrS$ has 8184), as are relation order and degree \cite{Albantakis2023}. Moreover, these overall disparities hold true for other states of PQRS (Supplementary Fig. \ref{fig:pqrs_cycle}) and all other states the computer can support (see \hyperref[subsubsec:micro_analysis_of_wcc]{``Micro-grain analysis of the weakly connected computer''} supplement). Thus, despite being functionally equivalent, the computer and PQRS do not support equivalent cause--effect structures. 

\subsection*{The computer's lack of integration is not merely due to lack of feedback}

It is possible to modify the computer by introducing feedback connections that alter its TPM, but do not impair its ability to simulate $pQrS$ (Fig. \ref{fig:ces_inequivalence}C). This version of the computer is strongly connected in the graph-theoretical sense, yet as a whole remains reducible, with $\varphi_s = 0$ ibits. In fact, the same subsystems that were maximally irreducible without feedback connections (Fig. \ref{fig:ces_inequivalence}A--B) remain maximal even with such feedback in place (see \hyperref[subsubsec:micro_analysis_of_scc]{``Micro-grain analysis of the strongly connected computer''} supplement).  This is because, despite the presence of feedback connections, the computer has many \emph{fault lines}, such as sparse connectivity and bottlenecks, along which it segregates into smaller, non-overlapping and maximally integrated pieces \cite{Marshall2023}. In other words, the computer with or without feedback is an aggregate of about two dozen tiny complexes, rather than a single, larger complex in its own right.

\subsection*{The computer's failure to replicate the cause--effect structure of PQRS cannot be rescued by treating its units or states at macro grains} 

\begin{figure*}[h]
\centering
\includegraphics[width=7.5in]{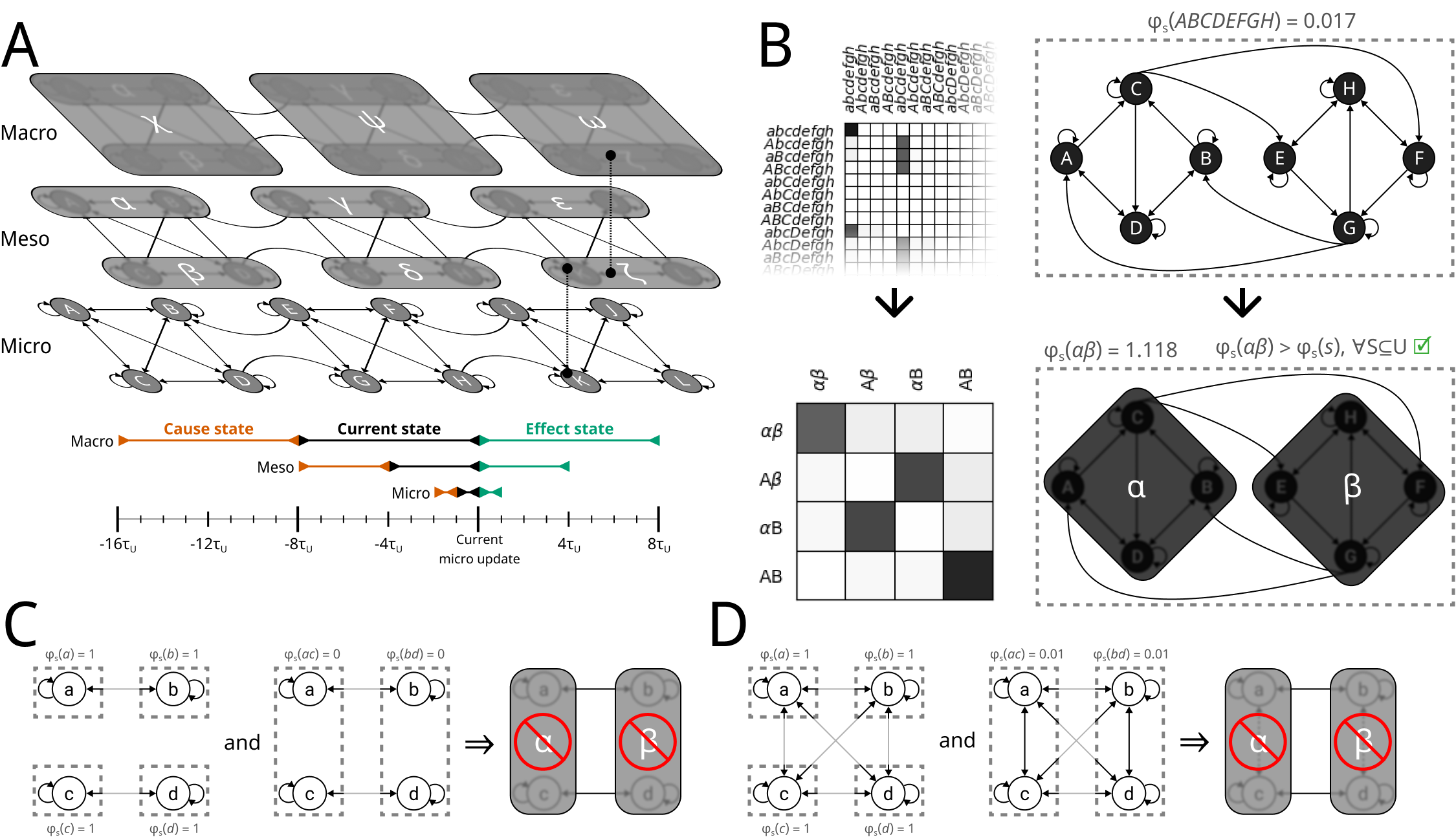}
\caption{\textbf{Identifying a system's intrinsic units based on maximally irreducible cause--effect power.}
(\textbf{A}) Identifying a system's intrinsic units by macroing over units and updates. Macro states can be defined not only over sets of units (top), but also over sets of updates (bottom). Macroing can be performed recursively across ``meso'' units and updates. 
(\textbf{B}) An example of a system whose intrinsic cause--effect power is maximal at a macro grain. There are combinatorially many ways to macro the eight micro units $ABCDEFGH$. Shown here is one possible macroing into two macro units  $\alpha\beta$ over two updates. The corresponding macro TPM captures the intrinsic causal powers of $\alpha\beta$ at this grain, and is analyzed using the same formalism for micro systems. $\alpha\beta$ in state $11$ has system integrated information $\varphi_s = 1.118$ ibits, greater than all of the micro-grain candidate systems (max $\varphi_s = 0.135$ ibits, not shown), including the system $ABCDEFGH$ ($\varphi_s = 0.017$). 
(\textbf{C}) A complex and its intrinsic units must comply with IIT's postulates of physical existence. 
Consider a system of four units $abcd$, in which there are no connections between vertical neighbors. Each micro unit has $\varphi_s = 1$ ibits on its own (left), while each pair of vertical neighbors has $\varphi_s = 0$ ibits (middle). Because each pair of vertical neighbors is reducible, they do not satisfy the integration postulate, and are not valid macro units (right).
(\textbf{D}) Macro units must also be definite, satisfying the exclusion postulate. This translates into the requirement that a macro unit must be ``maximally irreducible within'' (have greater $\varphi_s$ than any subset of its constituents) \cite{Marshall2024}. Consider $abcd$ when weak connections are introduced between vertical neighbors (left). Although each pair of vertical neighbors is now weakly integrated with $\varphi_s = 0.01$ ibits (middle), they are not maximally irreducible within (e.g., $\varphi_s(ac) < \varphi_s(a)$). Thus, the vertical neighbors are not valid macro elements (right). 
IIT's postulates of physical existence do not allow for something (a unit) to be built out of nothing (panel \textbf{C}) or nearly nothing (panel \textbf{D}). 
Adapted from \cite{Marshall2024}.}
\label{fig:micro_macro}
\end{figure*}

In IIT, the units that constitute a complex, called \emph{intrinsic units,} \cite{Marshall2024}  are those that maximize the complex's existence, as measured by $\varphi_s$. In principle, a complex's intrinsic units are established by evaluating the system at all possible grains, exhaustively grouping subsets of its micro units into macro units, and mapping states of the constituent micro units to states of the resulting macro units (Fig. \ref{fig:micro_macro}A) \cite{Marshall2018, Marshall2024, Hoel2016, Comolatti2022}. This procedure, called macroing, is done because a system's intrinsic causal powers may be higher at a coarser than at a finer grain, depending on its internal organization (Fig. \ref{fig:micro_macro}B). In such cases, a system's intrinsic units are macro units, rather than micro units. 

A complex and its intrinsic units must comply with IIT's postulates of physical existence \cite{Albantakis2023}. Thus, like complexes, macro units can only qualify as units if their cause--effect power is irreducible to that of independent subsets (satisfying integration). Otherwise, one could literally build something (a macro unit) out of nothing (non-interacting micro units; Fig. \ref{fig:micro_macro}C). And, just as complexes must be definite, which translates into maximally irreducible cause--effect power (satisfying exclusion), so must macro units. Otherwise, one could build something (a macro unit) out of nearly nothing (weakly interacting sets of micro units; Fig. \ref{fig:micro_macro}D). Several other constraints on what constitutes a valid macro unit follow from the postulates. For example, intrinsic units should not overlap (satisfying exclusion) \cite{Marshall2024}.

From the extrinsic perspective of a computer engineer, certain macroing schemes may provide a natural way to construct a high-level characterization of the computer. For example, it is useful to abstract away low-level digital logic into high-level components with precise input--output behavior, such as multiplexers, adders, bit shifters, and registers. It is also helpful to chunk together multiple updates of the state of such components into interpretable computational steps. 

\begin{figure}[ht!]
\centering
\includegraphics[width=\linewidth]{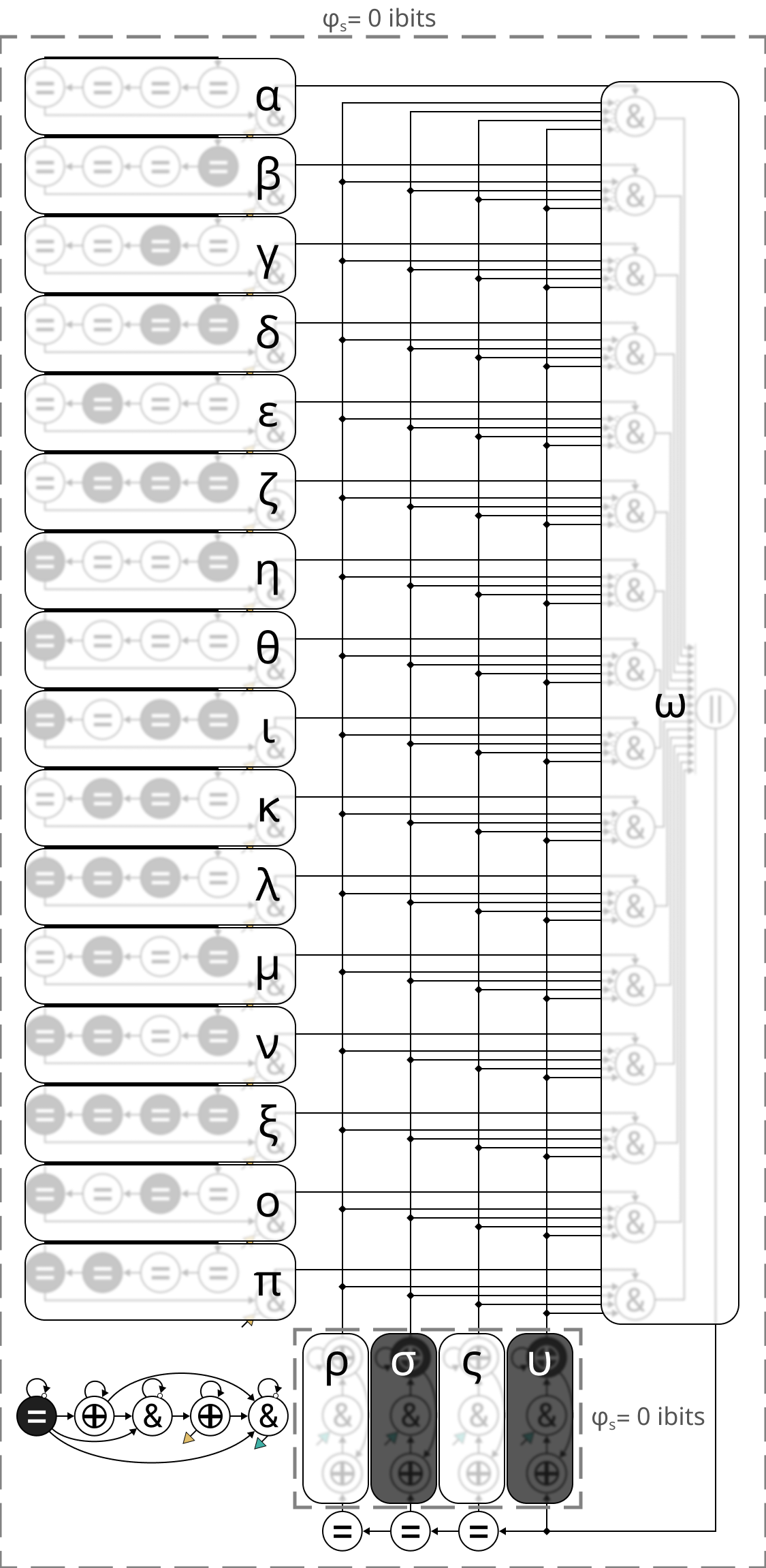}
\caption{\textbf{The computer does not replicate the target's cause--effect structure at any macro grain.}
One might consider macroing the computer as shown: Each line of the stored program and its corresponding instruction register unit is grouped into its own unit ($\alpha$-$\pi$), the multiplexer units are grouped together ($\omega$), and each data register is grouped into its own unit ($\rho$, $\sigma$, $\varsigma$, $\upsilon$). However, not all of these units qualify as intrinsic units (see text). Moreover, integration occurring in the background conditions, extrinsic to a candidate complex, does not contribute to its irreducibility \cite{Marshall2024}. The functional integration among the units constituting the candidate complex $\rho\sigma\varsigma\upsilon$ is entirely mediated by the multiplexer, and must be discounted (see text).} 
\label{fig:macro_computer}
\end{figure}

Fig. \ref{fig:macro_computer} shows one of many ways in which one might consider macroing the computer from an extrinsic, computational-functionalist perspective. Each line of the program, together with its corresponding unit in the instruction register, is grouped into a macro unit ($\alpha$ through $\pi$). The state of each macro unit might be defined as the state of its constituent instruction register unit, abstracting away all the sub-macro interactions within. One might also group the multiplexer units together ($\omega$) and define the macro state as the state of the multiplexer's output. Finally, one might consider each data register as a macro unit ($\rho$, $\sigma$, $\varsigma$, $\upsilon$) whose state is defined by the state of its output (P', Q', R', S'). This way of macroing results in a clear mapping between macro units $\rho\sigma\varsigma\upsilon$ and PQRS that captures the functional equivalence of the two systems.

However, the macroing shown in Fig. \ref{fig:macro_computer} is inconsistent with IIT's postulates of physical existence. For example, the requirement that macro units be integrated (Fig. \ref{fig:micro_macro}C--D) means that $\omega$ is not a valid unit, as it is fully reducible (for further examples, see \hyperref[sec:applying_postulates_to_macro_computer]{``Applying IIT's postulates to candidate macro units within the computer''} supplement). In fact, as shown in the \hyperref[subsubsec:macro_analysis_of_scc]{``Macro grain analysis of the strongly connected computer''} supplement, there is no macroing of the computer reflecting its function as a simulans that can replicate the cause--effect structure of PQRS. On the basis of IIT, this means that functional equivalence is not accompanied by phenomenal equivalence. 

While one example is enough to make the case, we conjecture that this result will generalize to many standard computer architectures. The underlying reason is that IIT is concerned with the complete set of causal powers intrinsic to a system, no more (i.e., no powers extrinsic to a mechanism or system may be borrowed, and no intrinsic powers may be multiplied) and no less (i.e., one cannot selectively pick the units or interactions within the system that are of interest to an external observer, and ignore the rest). For the very general computer architecture explored below, macroing of micro units into functionally relevant macro units necessarily involves either omission of, or causal overlap over, a set of shared constituents essential to each unit's function. These two facts together make it extremely unlikely that any similar computer could precisely replicate a target system's cause--effect structure (see \hyperref[sec:modern_architectures]{``How the computer compares to modern computer architectures''} and \hyperref[subsubsec:macro_likely_low_phi]{``Why the computer is unlikely to support a complex with high structure integrated information ($\Phi$)''}). Regardless, the point is that, in general, functional equivalence does not imply cause--effect structure equivalence, and therefore does not imply phenomenal equivalence on the basis of IIT. 

\subsection*{Cause--effect structures specified by the computer can be dissociated from those of target systems it is simulating} 

The above example shows that while a simple computer can be functionally equivalent to a system with a radically different substrate---the complex PQRS---it cannot specify an equivalent cause--effect structure. Owing to its architecture, the computer fragments into many small complexes, each of which specifies a trivial cause--effect structure, rather than the single, much larger cause--effect structure specified by PQRS. We now show that the computer fragments into the same small complexes that specify trivial cause--effect structures regardless of the target system it is simulating. 

\begin{figure}[ht]
\centering
\includegraphics[width=\linewidth]{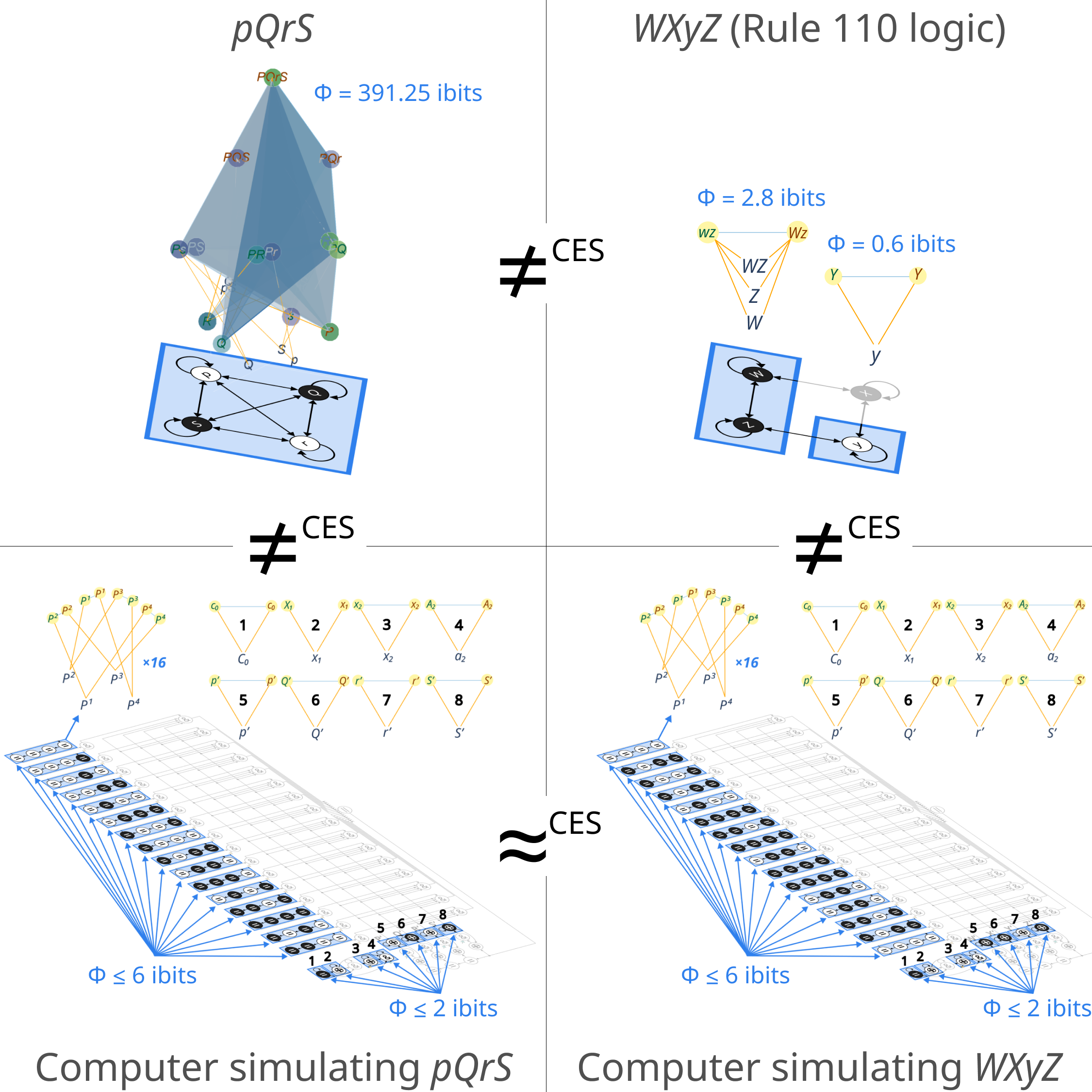}
\caption{\textbf{Dissociation between function and cause--effect structure.}
The single, large cause--effect structure specified by $pQrS$ (top left) and the two, smaller cause--effect structures specified $WXyZ$ (top right) bear little resemblance to each other. However, the fragmented cause--effect structures specified by the computer at the micro grain are insensitive this difference. In fact, the cause--effect structures supported by the computer as it simulates $pQrS$ (bottom left; see Fig. \ref{fig:ces_inequivalence}B) are nearly identical to those supported by the computer as it simulates $WXyZ$ (bottom right; see \hyperref[subsec:ring_motif]{``Analysis of the ring motif''} supplement). Thus, the cause--effect structures of the computer and those of the target systems it simulates are not just different, but dissociated.}
\label{fig:micro_ces_dissociation}
\end{figure}

Consider another four-unit system, WXYZ, an elementary cellular automaton that happens to implement Wolfram's Rule 110 (Supplementary Fig. \ref{fig:WXyZ}) \cite{Wolfram1983}. Cellular automata of any size can be made to implement Rule 110, which completely specifies that system's causal model and TPM. Furthermore, systems of arbitrary, finite size that implement Rule 110 are Turing complete \cite{Wolfram1983, Cook2004}. If placed into state 1101, $WXyZ$ fragments into two complexes, whose cause--effect structures are markedly different from that specified by $pQrS$ (Fig. \ref{fig:micro_ces_dissociation}, upper quadrants). The general-purpose four-bit computer described here can simulate $WXyZ$ just as well as $pQrS$. Moreover, analysis of the computer at the micro grain reveals that it again fragments into 24 complexes, specifying similarly trivial cause--effect structures, no matter whether it is simulating $pQrS$, $WXyZ$ (Fig. \ref{fig:micro_ces_dissociation}, lower quadrants), or any other four-unit system. This shows that the fragmented cause--effect structures supported by the computer are insensitive to major changes in the cause--effect structure supported by different target systems.

\subsection*{The dissociation between input--output functions and cause--effect structures can be extended by induction to arbitrarily large computers simulating arbitrarily complex behaviors} 

The computer shown in Fig. \ref{fig:computer_intro} is only capable of simulating four-bit/four-unit systems, but none of the qualitative findings above depend on the computer's size---only on its architecture. So long as this architecture is preserved, similar results will hold for arbitrarily large computers.

\begin{figure*}[h]
\centering
\includegraphics[width=5.45in]{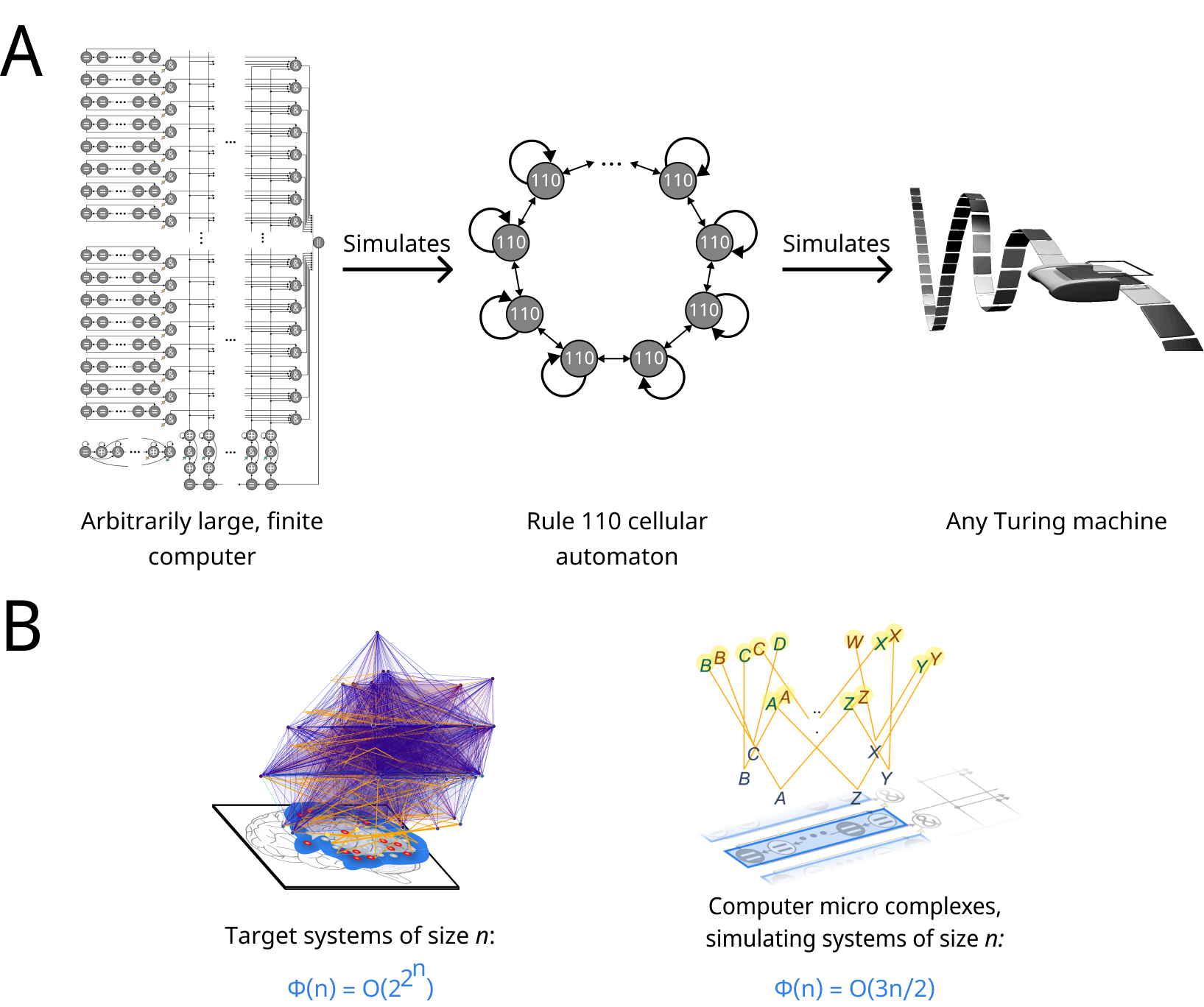}
\caption{\textbf{Inductive extension to large computers simulating arbitrarily complex systems.}
(\textbf{A}) The computer can be incrementally extended to simulate systems of arbitrary size (left), including biological brains or the Rule 110 cellular automaton (middle), which is Turing-complete (right). By transitivity, the programmable computer is Turing-complete too, if extended to arbitrary sizes.
(\textbf{B}) By induction on the size of the simulated system, it can be shown that the cause--effect structures specified by subsystems within the computer will not change qualitatively at the micro grain and can only grow linearly with system size (see \hyperref[subsec:ring_motif]{``Analysis of the ring motif''} supplement), whereas the cause--effect structures of the simulated target systems may grow double-exponentially in size and richness, resulting in an increasing phenomenal dissociation between the computer and the systems it simulates.}
\label{fig:induction}
\end{figure*}

To increase the capacity of the computer to $n = 2^k$ bits ($k \in\ \mathbb{N}$) requires increasing the number of program instructions to $2^n$, each with length $n$, appending a frequency divider to the existing clock chain (for a total of $2k + 1$ timekeeping units), increasing the number of data registers to $n$, extending the multiplexer to have $2^n$ inputs (one corresponding to each possible state of the system), and extending the buffer (chain of copy units extending from the processor) to $2n - 5$ units. In this way, the expansion process can be continued to create an arbitrarily large computer; an $n$-bit version of the computer would require $\Theta(n2^n)$ units and $2n$ updates to complete one simulation iteration (Fig. \ref{fig:induction}A, left).

Like the computer shown in Fig. \ref{fig:computer_intro}, such a system will have $2^n + n + k + 1$ complexes: one $n$-unit complex for each line of the program, and one single-unit complex for each unit with a self-loop in the clock chain and data registers. Each complex will continue to specify the same simple cause--effect structures, with the largest having $\Phi\leq\frac{3n}{2}$ ibits (\hyperref[subsec:ring_motif]{``Analysis of the ring motif''} supplement). As before, adding feedback connections will not change these results. Again, macroings of the system following from its function as simulans should not be expected to replicate the cause--effect structure(s) of its simulanda, and there will always be a class of systems for which replication is provably impossible.

Such an arbitrarily large computer is capable of simulating arbitrarily complex behaviors. For example, an $n$-bit version of a digital computer can simulate any $n$-variable logical system, including all cellular automata. An arbitrarily large elementary cellular automaton implementing Rule 110 is Turing-complete \cite{Cook2004}, and since the computer is at least as powerful as this cellular automaton, the computer is itself Turing-complete. By the Church--Turing thesis \cite{Copeland2008}, then, our simple computer could perform any computable function.

The dissociation shown above (Fig. \ref{fig:micro_ces_dissociation}) between the cause--effect structures specified by the computer at the micro grain and those specified by the target systems it is simulating can be exacerbated if the targets are large and have high $\Phi$. In fact, the magnitude of this dissociation can be a double-exponential function of the size of the simulated system \cite{Albantakis2015}. A system of $n$ units has $2^n - 1$ possible distinctions with $\sum\varphi_d = \mathcal{O}(2^n)$ ibits \cite{Zaeemzadeh2024b}. Only $n$ of these distinctions capture causal constraints exerted by first-order mechanisms; the vast majority of possible distinctions will be higher-order, corresponding to causal constraints exerted by combinations of units. In turn, the number of possible relations is given by the powerset of distinctions, and is therefore double-exponential as a function of $n$ for rich systems, with $\sum\varphi_r = \mathcal{O}(2^{2^n})$ ibits. In other words, the structure integrated information of a rich system can be as high as $\Phi = \sum\varphi_d + \sum\varphi_r = \mathcal{O}(2^{2^n})$ ibits \cite{Zaeemzadeh2024b}. There is no reason to expect that the computer, no matter how large its size and how complex its program, will specify anything beyond small cause--effect structures, each with just first-order mechanisms and linearly increasing $\Phi$ (\hyperref[subsubsec:macro_likely_low_phi]{``Why the computer is unlikely to support a complex with high structure integrated information ($\Phi$)''}).   

\section*{Discussion}

IIT starts from the essential properties of experience and formulates them in causal, operational terms. This leads to the conclusion that to support consciousness, a system must have cause--effect power that is intrinsic, specific, irreducible, definite (maximally irreducible), and structured. The particular cause--effect structure of distinctions and relations specified by the system accounts for the quality of a particular experience---``what it is like to be'' that system in its current state \cite{Nagel1974}. The amount of structure integrated information ($\Phi$) of the cause--effect structure corresponds to the quantity of consciousness.

Here we applied the mathematical framework of IIT to the causal model of a simple target system, constituted of four units implementing Boolean functions, as well as to an explicit model of a simple, stored-program, four-bit computer, constituted of 117 units and programmed to be functionally equivalent to the target. Although the computer replicated the input--output functions of the target, neither the computer as a whole, nor subsets of its units, could replicate the cause--effect structure of the target. Furthermore, the latter could not be replicated by macroing the units of the computer according to the postulates of IIT. Finally, these results held for a Turing-complete version of the computer which, in principle, can simulate arbitrarily large systems. 

The target system employed here was extremely small and specified a single cause--effect structure that was comparatively simple (for the state shown in Fig. \ref{fig:pqrs_intro}, composed of 13 distinctions and 8184 relations, with $\Phi = 391.25$ ibits). The functionally equivalent computer, considered at the micro grain, fragmented into 24 small complexes, each of which specified trivial cause--effect structures (with a maximum of 4 distinctions and 4 relations). The substrate of consciousness in the brain can be assumed to be a complex constituted of a much larger number of units that specifies a cause--effect structure of extraordinary richness and high $\Phi$ \cite{Zaeemzadeh2024b, Haun2019}. Based on the principles of IIT and the results obtained here, we conjecture that computers with architectures sufficiently similar to the one shown here will fragment into many small complexes that specify trivial cause--effect structures (see \hyperref[sec:modern_architectures]{``How the computer compares to modern computer architectures''} and \hyperref[subsubsec:macro_likely_low_phi]{``Why the computer is unlikely to support a complex with high structure integrated information ($\Phi$)''}). If they were to support some non-trivial cause--effect structures, it would \emph{not} be in virtue of their computational power, functional sophistication, or which simulations they perform. We argue that if standard computers implementing general AI were to replicate our behaviors and cognitive functions---even if they did so by simulating, neuron by neuron, the functioning of our brain---they would generally have negligible consciousness and would not replicate our experiences.  

The issue of computer consciousness has been considered before, but not on the basis of a comprehensive, quantitative theory of consciousness. Some have argued that computers are unlikely to be conscious, motivated by various intuitions. For example, consciousness might depend on some critical biological attribute, such as homeostasis, emotion, or embodiment \cite{Aru2023, Damasio2023, Dreyfus1992, Sayre1993, Seth2024}. However, it is not clear why such attributes would be critical for consciousness, nor why they would be beyond the reach of computers \cite{Churchland1990, Pinker2009}. Other have suggested that computers may lack some critical ``physical-chemical'' ingredient, though it is not clear which or why \cite{Searle1980, Penrose1989}. 

On the other side of the fence, a common assumption is that consciousness should be identified with some function or computation, whether language, attention, self-monitoring, ``global broadcasting'' of information, ``recurrent processing,'' or ``predictive processing'' \cite{Dehaene2018, Graziano2015, Wiese2021, Fodor1979, Blum2022, Brockman2015}. For some, consciousness would also require embodiment and interaction with the physical world, including sensing, acting, and autonomy \cite{Prinz2009, Pezzulo2016}, or the replication of some neuronal microfunctions \cite{Clark1989}. Functions are multiply realizable, and mental states, if seen as intermediate computational states between inputs and outputs, can in principle be implemented by both brains and computers. However, there is strong evidence that consciousness can be dissociated from a variety of cognitive functions \cite{Koch2016}. We can be conscious even when we are not doing, thinking, or saying anything, as in ``pure presence'' meditation \cite{Forman1997, Boly2024}; in dreaming sleep, when we are not interacting with the environment \cite{Siclari2017}; under certain psychedelics \cite{Timmermann2023}, or under ketamine anesthesia \cite{Sarasso2015}. 

Moreover, some have argued that computation is an implausible foundation for consciousness precisely because it is substrate-independent and observer-dependent \cite{Blackmon2013, Putnam1987, Brette2019}. Turing completeness can be achieved using radically different substrates \cite{Turing1937, Moore1956, Chomsky1959, Rabin1959}, whether simple \cite{Cook2004, Schoenfinkel1924} or complex, feedforward \cite{Hornik1991} or recurrent \cite{Siegelmann1995, Krohn1965}. Also, any program could in principle be mapped by an observer onto almost any thermodynamically open substrate through an appropriate encoding of inputs and outputs \cite{Blackmon2013, Putnam1987} (but see \cite{Chalmers1996, Shiller2024}). Why would one assume that a simulation of what something \textit{does} should be equivalent to what something \textit{is}? After all, we do not expect that running a simulation of a rainstorm will make a computer wet \cite{Searle1980, Block1978} or that a simulation of a black hole will bend space-time around the computer. Above all, why should certain functions or computations (but not others) be associated with experience, and why would they feel the way they do (and not some other way)?

On the basis of IIT and in sharp contrast to computational functionalism, we argue that what matters for consciousness, its presence, quality, and quantity, is a system's intrinsic causal structure---what the system \textit{is}---rather than its extrinsic functions---what the system \textit{does} \cite{Albantakis2023, Hendren2024, Tononi2015, Zaeemzadeh2024a, Tononi2023}. Because internal organization cannot generally be inferred from overall system dynamics or input--output behavior \cite{Oizumi2014, Albantakis2019, Grasso2021a, Albantakis2023}, functions are not reliable proxies for the properties of consciousness. In short, consciousness is about being, not doing.

Of course, the conclusions derived in this paper can only be considered valid to the extent that IIT itself can be considered valid. Future experiments may show that, contrary to the predictions of IIT, our own consciousness is not associated with a maximum of intrinsic cause--effect power, or that the quality of specific experiences (say, the experience of visual space) is not consistent with the properties of the associated cause--effect structures, \cite{Albantakis2023, Tononi2016, Haun2019, Tononi2015a}. In that case IIT would be invalidated, and with it the present results concerning machine consciousness. So far, however, IIT provides a rational basis for making inferences about the presence of consciousness in biological or artificial systems other than ourselves. IIT is grounded in the essential properties of our own consciousness---the only case in which we have immediate, irrefutable proof that subjective experience exists. Its predictions can be assessed on our own brain, so far with consistent results \cite{Tononi2016}. It can explain disparate facts in a coherent manner---for example, why certain brain substrates are important for consciousness while others are not \cite{Tononi2016, Koch2016}, why consciousness vanishes during sleep despite ongoing neural activity \cite{Massimini2005}, whether a stimulus will be perceived or remain unconscious during psychophysical manipulations \cite{Haun2017}, why the stream of consciousness may split under some concurrent tasks \cite{Sasai2016}, why specific contents of consciousness such as the visual field, temporal flow or mental objects may feel the way they do \cite{Haun2019, Comolatti2024, Grasso2021b, Haun2024, Balduzzi2009}, and why consciousness may have evolved due to selective pressure for information integration \cite{Albantakis2014, Edlund2011, Mayner2024}. Finally, IIT provides a precise mathematical formulation of the requirements for consciousness. Crucial for the present purposes, this formulation can be applied, in principle, whenever an adequate causal model of a system is available, irrespective of what it is made of---neurons or transistors---and irrespective of its functions.

\begin{figure}[ht]
\includegraphics[width=\linewidth]{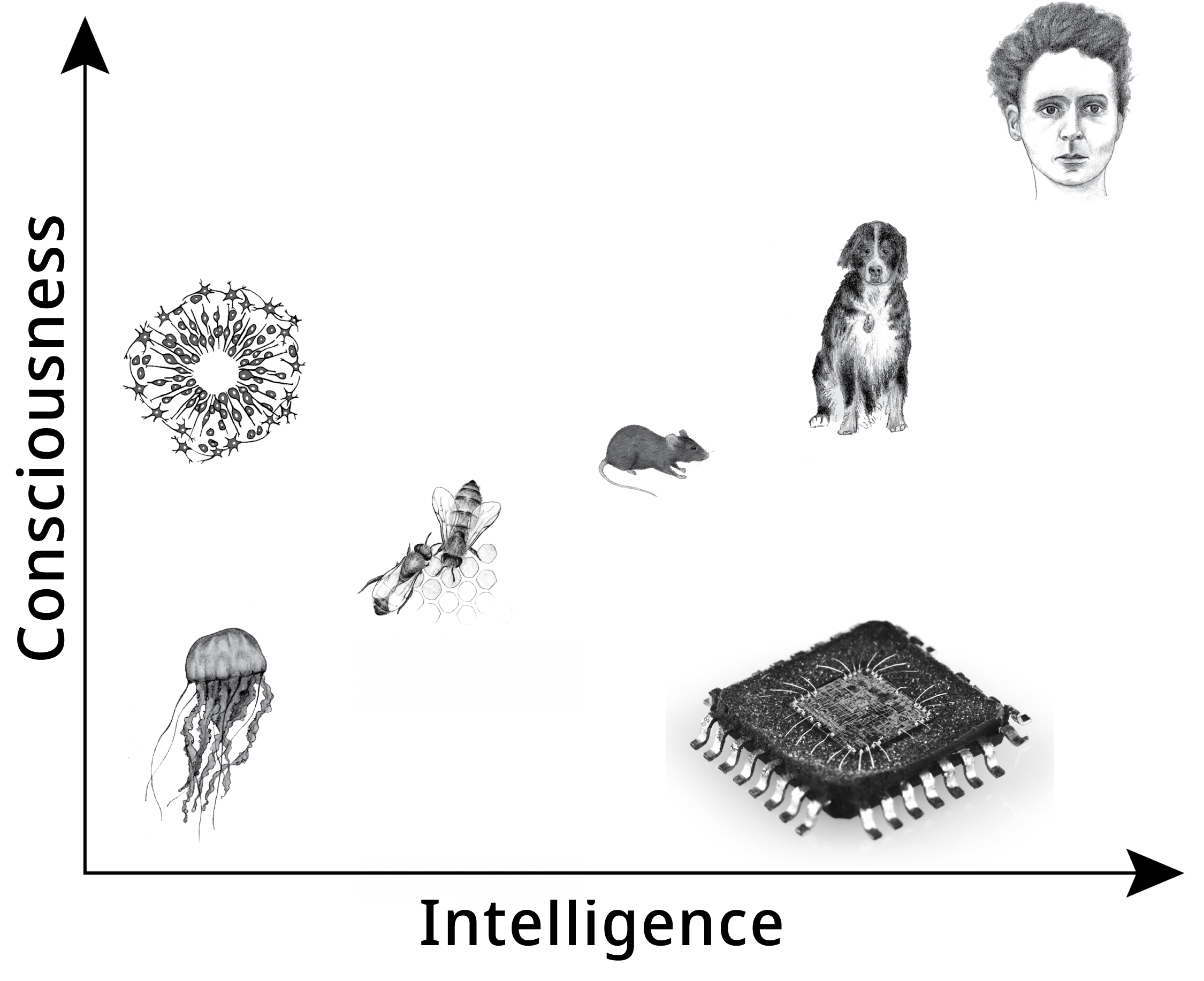}
\caption{\textbf{A double dissociation between consciousness and intelligence.}
In the biological systems to which we are accustomed, consciousness and intelligence are often assumed to go hand in hand. IIT offers a possible explanation for this co-evolution \cite{Albantakis2014}. However, although our instinctive attribution of consciousness based on intelligent behavior may be justified in some cases, it is unlikely to be justified in artificially designed systems---either silicon or organic. We have shown here that IIT implies the distinct possibility of artificial intelligence without consciousness (bottom right). On the other hand, our analysis does not rule out artificial consciousness altogether. An intriguing question is whether truly neuromorphic computers that mimic the physical organization of our brain and do not rely on a CPU could be designed to achieve both functional and phenomenal equivalence to human brains. Similarly, it is an open question whether artificial organic systems, such as cerebral organoids (top left), may lead to forms of consciousness without intelligence. Modified from \cite{Koch2019}.}
\label{fig:double_dissociation}
\end{figure}

An important issue is the extent to which these results can be generalized. The stored-program computer model introduced here captures certain fundamental aspects of modern computer architectures, such as a central processing unit (CPU), registers, a clock, and separation between computation and memory, quite unlike nervous systems. By applying IIT's postulates to this particular computer-like system, we demonstrated that functional equivalence with the system it is simulating is not accompanied by equivalence in cause--effect structure, and thus by phenomenal equivalence. In principle, one could apply a similar approach to a computer architecture of choice \cite{Sampson2024}. However, the results obtained here depend primarily on the presence of computational bottlenecks, such as the processor/multiplexer, which force simulated constituents of the target system (say, virtual neurons) to overlap over the same physical substrate (say, a real register). Hence, our findings should extend to a broad class of machines that rely on CPUs or even graphical processing units (GPUs) (see \hyperref[sec:modern_architectures]{``How the computer compares to modern computer architectures''} supplement). They would also extend to brain--machine interfaces that were to replace the connections among neurons by multiplexing the output-input traffic through a CPU; or to replace the input--output transfer function of neurons, again multiplexed through a CPU, while leaving neuronal connections in place (see \hyperref[sec:replication_vs_simulation]{``Replication vs. Simulation''} supplement). On the other hand, our analysis does not rule out artificial consciousness in other contexts. For example, an intriguing question is whether neuromorphic computers that mimic the physical organization of our brain \cite{Ham2021, Serb2020, Schemmel2010, Mead1990, Young2019} could be designed to achieve both functional and phenomenal equivalence. This would imply that similar experiences can be supported by different kinds of substrates \cite{Putnam1967a}. A related question is whether quantum computers, with entangled qubits, might be engineered to possess the required cause--effect structures and high integrated information \cite{Albantakis2023-Quantum, Kleiner2021, Zanardi2018}. 

The identification of subjective experience with maxima of intrinsic cause--effect power also leads to a reconsideration of the relationship between intelligence and consciousness (Fig. \ref{fig:double_dissociation}) \cite{Koch2017}. Historically, we have been accustomed to intelligence and consciousness going hand in hand: to behave intelligently, we typically need to be conscious---in fact, we can hardly do anything when unconscious. Humans have an innate tendency to attribute agency and consciousness to other agents that behave intelligently. If general AI came about, it would probably feel unnatural, not to say narrow-mindedly bio-centric, to deny consciousness to machines that act, speak, and reason like us. We showed previously that IIT offers an explanation for why intelligence and consciousness may tend to co-evolve in biological organisms: selective pressure for intelligence, acting on resource-limited substrates such as brains, typically leads to an increase in integration among the parts of a system \cite{Albantakis2014, Edlund2011}. This is seldom the case for engineered systems, for which simplicity and modularity are favored \cite{Simon1996}. Thus, while our instinctive attribution of consciousness to biological organisms behaving intelligently may often be justified, in other cases IIT leads to inferences that may contrast with our intuitions and projections. For example, cerebral organoids sharing key anatomical features of brain regions that support consciousness, if kept in a dream-like state, might be vividly conscious despite lacking overt intelligence \cite{Birey2017, Farahany2018}. Conversely, as shown here, a traditional computer might exhibit human-like behavior and extraordinary intelligence yet experience nothing---it would not see, hear, feel, think, or understand anything. It would only act \textit{as if} it did---it would do everything and be nothing.

\minisection*{ACKNOWLEDGEMENTS} 
{\normalfont\sffamily\scriptsize\selectfont 
This project was made possible through the support of a grant from Templeton World Charity Foundation (TWCF0216, G.T.). In addition, this research was supported by the David P White Chair in Sleep Medicine at the University of Wisconsin-Madison (G.T.), by the Tiny Blue Dot Foundation (UW 133AAG3451; G.T.), and by the Natural Science and Engineering Research Council of Canada (NSERC; RGPIN-2019-05418; W.M.). L.A. acknowledges the support of a grant from the Templeton World Charity Foundation (TWCF-2020-20526). The funders had no role in study design, data collection and analysis, decision to publish, or preparation of the manuscript.
}

\minisection*{AUTHOR CONTRIBUTIONS} 
{\normalfont\sffamily\scriptsize\selectfont\insertcreditsstatement} 

\AtNextBibliography{\sffamily\scriptsize} 
\printbibliography 

\clearpage
\onecolumn

{\centering\Huge\sffamily\bfseries Supplementary Materials\par}
\vspace{0.5cm}
\hrule

\section*{Step-by-step guide to the computer's operation.}
\label{sec:step_by_step}

Let $U$ be the four-bit computer, which simulates any system of four Boolean units. The computer is composed of synchronously updating discrete units, each of which implements an elementary logic function. These units are organized into four main modules: the clock and frequency dividers, the program and instruction registers, the data registers, and the multiplexer/processing unit (Figure \ref{fig:computer_with_labeled_units}). The data registers store the state of the system until instructed to update; the program memory stores the function of the units that the computer is simulating; the processing unit combines this information about state and function to compute the next state of the data registers; the clock synchronizes these processes and tells the registers when to update their state. When initialized appropriately, the computer simulates the behavior of any four-unit Boolean network indefinitely. The system being simulated is referred to as the \emph{target system}.

A reference implementation with precise details of every unit and connection can be found at \href{https://github.com/CSC-UW/Findlay_et_al_2024a}{https://github.com/CSC-UW/Findlay\_et\_al\_2024a}. A diagram of the computer with all units labeled is shown in Fig. \ref{fig:computer_with_labeled_units}. Briefly, the four-bit computer has the following components:
\begin{itemize}
    \item Four data registers $\textbf{R} = \{\REG{1}{}, \REG{2}{}, \REG{3}{}, \REG{4}{}\}$, ordered from left to right. Each register $\REG{i}{}$ contains 3 units: $\REG{i}{} = \{\REG{i}{SIM}, \REG{i}{AND}, \REG{i}{XOR}\}$. 
    In the main text, $\REG{1}{SIM}$, $\REG{2}{SIM}$, $\REG{3}{SIM}$, and $\REG{4}{SIM}$ are referred to as P', Q', R', and S', respectively.
    
    \item Sixteen program lines $\textbf{P} = \{\PRG{}{1}, \PRG{}{2}, \ldots, \PRG{}{16}\}$, ordered from top to bottom. Each program line is a ring of 4 COPY units: $\PRG{}{i} = \{\PRG{1}{i}, \PRG{2}{i}, \PRG{3}{i}, \PRG{4}{i}\}$.
    
    \item An instruction register $\textbf{IR} = \{\IR{1}, \IR{2}, \ldots, \IR{16}\}$, consisting of sixteen AND units, ordered from top to bottom.
    
    \item A multiplexer $\textbf{M} = \{\MUX{1}, \MUX{2}, \ldots, \MUX{16}, \MO\}$, consisting of sixteen AND units, ordered from top to bottom, and a generalized OR unit ($\MO$).
    
    \item A timekeeping chain $\textbf{T} = \{\CO, \CX{1}, \CA{1}, \CX{2}, \CA{2}\}$, consisting of the core oscillator $\CO$ and two frequency dividers. 
    
    \item A buffer $\textbf{B} = \{\BUF{1}, \BUF{2}, \BUF{3}\}$, consisting of three COPY units between the multiplexer's output (MO) and the leftmost register $\REG{1}{}$. 
\end{itemize}

Once initialized (see below), the computer takes eight updates (nine transitions) to simulate a single state transition of the target system. After these eight updates, which we refer to as a single \emph{simulation iteration}, the states of units $\REG{1}{SIM}$, $\REG{2}{SIM}$, $\REG{3}{SIM}$, and $\REG{4}{SIM}$ (also referred to as P'Q'R'S', see main text Figure 2) update synchronously to reflect the state of the target system. The target and the computer are functionally equivalent in the sense that there exists a constant factor $C$ (here, $C=8$) such that the state of the target at update $i$ (in its own reference frame) is equal to the state of $\REG{1}{SIM}\REG{2}{SIM}\REG{3}{SIM}\REG{4}{SIM}$ at all updates $t \in \left(\thinspace iC, (i + 1)C \thinspace\right]$ (in the computer's reference frame). 

We now walk through the initialization procedure and an entire simulation iteration, as the computer simulates PQRS's transition from state 0101 to state 1110. 

\subsection*{Update 0: Initialization}
\label{subsec:update0}
Figure \ref{fig:update0} shows the initialized computer. First, the initial states of units $\REG{1}{SIM}$, $\REG{2}{SIM}$, $\REG{3}{SIM}$, and $\REG{4}{SIM}$ are set to reflect the initial state of the target (0101). We set the initial state of $\REG{1}{SIM}$ to OFF (0/white), the initial state of $\REG{2}{SIM}$ to ON (1/black), the initial state of $\REG{3}{SIM}$ to OFF, and the initial state of $\REG{4}{SIM}$ to ON (main text Figure 2C). The other units in the data registers are set to OFF. 

Next, the state of the clock is set so that the COPY unit ($\CO$) is ON, and all other units ($\CX{1}$, $\CA{1}$, $\CX{2}$, and $\CA{2}$) are OFF. $\CO$ with its negated self-loop forms the oscillating core of the clock; its state will flip every update. $\CX{1}$ and $\CA{1}$ form a frequency divider. After the first update $\CX{1}$ will be ON for two updates, then OFF for two updates, and will repeat this pattern indefinitely (i.e., $\CX{1}$ has a 50\% duty cycle and a period of four updates). $\CA{1}$ will turn ON for the last one of every four updates (i.e., a 25\% duty cycle and a period of four updates). $\CX{2}$ and $\CA{2}$ form a second frequency divider, and have the same duty cycles as their upstream counterparts, and the same initial phase, but double their periods. $\CX{2}$ is OFF for four updates, then ON for four updates. $\CA{2}$ is OFF for seven updates, then ON for one update. For example, the state of each unit over the first 16 updates is shown in Table \ref{table:timekeeping_units}. 

\begin{table}[h]
\centering
\begin{tabular}{| c || c c c c c |} 
 \hline
 Update & $\CO$ & $\CX{1}$ & $\CA{1}$ & $\CX{2}$ & $\CA{2}$ \\ [0.5ex] 
 \hline\hline
 0 & 1 & 0 & 0 & 0 & 0 \\ 
 \hline
 1 & 0 & 1 & 0 & 0 & 0 \\ 
 \hline
 2 & 1 & 1 & 0 & 0 & 0 \\ 
 \hline
 3 & 0 & 0 & 1 & 0 & 0 \\ 
 \hline
 4 & 1 & 0 & 0 & 1 & 0 \\ 
  \hline
 5 & 0 & 1 & 0 & 1 & 0 \\ 
  \hline
 6 & 1 & 1 & 0 & 1 & 0 \\ 
  \hline
 7 & 0 & 0 & 1 & 1 & 1 \\ 
  \hline
 8 & 1 & 0 & 0 & 0 & 0 \\ 
  \hline
 9 & 0 & 1 & 0 & 0 & 0 \\ 
   \hline
10 & 1 & 1 & 0 & 0 & 0 \\ 
   \hline
11 & 0 & 0 & 1 & 0 & 0 \\ 
   \hline
12 & 1 & 0 & 0 & 1 & 0 \\ 
   \hline
13 & 0 & 1 & 0 & 1 & 0 \\ 
   \hline
14 & 1 & 1 & 0 & 1 & 0 \\ 
   \hline
15 & 0 & 0 & 1 & 1 & 1 \\ 
 \hline
\end{tabular}
\caption{\textbf{The state of each timekeeping unit over the course of the first 16 updates}}
\label{table:timekeeping_units}
\end{table}

Next, the states of the computer's program units are set in such a way that they encode the target's state transition rules (main text Figure 2B). The program has 16 lines, each implemented by a ring of four COPY units. The top ring encodes the state that PQRS would transition to from state 0000. The second ring from the top encodes the state that PQRS would transition to from state 1000. The third from 0100, the fourth from 1100, the fifth from 0010, the sixth from 1010, the seventh from 0110, the eighth from 1110, and so on (in little-endian like fashion, with leftmost bits changing faster than rightmost bits). The leftmost unit in each ring is set to the state that P would transition to, the second from the left is set to the state that Q would transition to, the third from the left is set to the state that R would transition to, and the fourth from the left is set to the state that S would transition to. Finally, the instruction register units ($\textbf{IR}$) are initialized OFF. 

Because we want to program the computer with the information that PQRS in state 0101 transitions to state 1110, we need to set the units in the eleventh ring from the top to ON, ON, ON, OFF, and ON when read from left to right. If we would like the computer's simulation to continue indefinitely, we should also initialize the other fifteen program instructions (main text Figure 2B). 

\subsection*{Transition from update 0 to update 1} 
Figure \ref{fig:update1} shows the first of nine transitions involved in a single simulation iteration. During this step, the computer fetches the data that it needs to compute the future state of P (equivalently, $\REG{1}{SIM}$). The two important elements of this calculation are the simulated current state of PQRS and, given that state, the rule describing what P would do. 

Recall that the states of the computer's program units were set in such a way that they encoded the target's state transition rules. The leftmost unit in each program line was initially set to represent a post-transition state of P, with each line of the program representing a different pre-transition state of PQRS. At update 1, the program and instruction registers have updated so that the initial states of these leftmost units are now the states of the instruction register units. Thus, each instruction register unit is set to a post-transition state of P, with each unit representing a different pre-transition state of PQRS. 

The states of $\REG{1}{SIM}$, $\REG{2}{SIM}$, $\REG{3}{SIM}$, and $\REG{4}{SIM}$---which reflect the simulated current state of PQRS---have not changed (by design, no $\REG{i}{SIM}$ will change unless its respective $\REG{i}{AND}$ is ON). The clock updates (Table \ref{table:timekeeping_units}).

\subsection*{Transition from update 1 to update 2} 
Figure \ref{fig:update2} shows the second of nine transitions involved in a single simulation iteration. During this step, the next state of P is computed, and the computer becomes ready to compute the next state of Q. 

Each AND unit in the multiplexer receives four inputs from the data registers and one input from the instruction register. We say that one of these AND units is enabled at update $t$ when all four of its inputs (after any negation) from the data registers were ON at update $t - 1$. Because no two multiplexer units receive the same four inputs from the data registers, at most one AND unit in the multiplexer can be enabled at any given update. In fact, the state of the data register outputs at any update $t$ enables exactly one multiplexer unit at update $t + 1$. 

Invariably, the topmost multiplexer unit will be enabled when the data register outputs are in state 0000. The second multiplexer unit from the top will be enabled when the data register outputs are in state 1000. The third in 0100, the fourth in 1100, the fifth in 0011, the sixth in 1010, the seventh in 0110, and the eighth in 1110, and so on. Notice how this mirrors the way in which the program memory is structured. When a particular AND unit in the multiplexer is enabled, its next state is determined by the content of the corresponding instruction register unit. This is the means by which we combine the current state of the target (represented by which multiplexer unit is enabled) with the desired behavior of the target in that state (represented by the corresponding instruction register unit). 

At update 1, the eleventh multiplexer unit from the top ($\MUX{11}$) was enabled. Whether it turns ON is determined by the eleventh line of the program ($\PRG{}{11} = \{\PRG{0}{11}, \PRG{1}{11}, \PRG{2}{11}, \PRG{3}{11}\}$) and instruction register ($\IR{11}$), which says what PQRS would do in state 0101. More precisely, the eleventh line's instruction register unit ($\IR{11}$) at update 1 was in the same state that P would transition to when PQRS is in state 0101. All together, this means that at update 2, the eleventh multiplexer unit ($\MUX{11}$) will be ON if and only if the next state of P is ON. All other AND units in the multiplexer will be OFF. 

To reiterate, there are two possibilities at update 2: either exactly one AND unit in the multiplexer is ON, in which case P should transition to ON in our simulation, or exactly no AND units in the multiplexer are ON, in which case P should transition to OFF in our simulation. At this point we say that the next state of P is computed, since an extrinsic observer of the system can tell what the next state of P should be without looking ahead into the computer's future (i.e., at updates $t > 2$). 

As all of this is happening, the computer is also becomes ready to compute the next state of Q. The state of the data register outputs has not changed since they were initialized. This means that the same multiplexer unit which was enabled at update 1 ($\MUX{11}$) will still be enabled at update 2, and the same program line ($\PRG{}{11}$) and instruction register ($\IR{11}$, though its output will of course change) will determine that multiplexer unit's state. Finally, the instruction register updates so that each unit is set to represent a post-transition state of Q, again with each line representing a different pre-transition state of PQRS. 

The clock updates (Table \ref{table:timekeeping_units}). 

\subsection*{Transition from update 2 to update 3} 
Figure \ref{fig:update3} shows the third of nine transitions involved in a single simulation iteration. During this step, the next state of P propagates towards the registers, the next state of Q is computed, and the computer becomes ready to compute R. 

At update 2, either exactly one AND unit in the multiplexer was ON, in which case P would transition to ON in our simulation, or exactly no AND units in the multiplexer were ON, in which case P would transition to OFF in our simulation. The multiplexer's generalized OR unit ($\MO$, which turns ON whenever one of its 16 inputs is ON and is OFF otherwise) at update 3 is therefore ON if the next state of P is ON, and OFF if the next state of P is OFF. In our example, the OR unit is ON. 

The next state of Q is now computed exactly as the next state of P was. Because Q transitions to ON when PQRS is 0101, exactly one AND unit in the multiplexer is ON. 

The state of the data register outputs has not changed. This means that the same AND unit ($\MUX{11}$) which was enabled at update 2 will still be enabled at update 3, and the same program line ($\PRG{}{11}$) and instruction register ($\IR{11}$, though its output will of course change) will determine that multiplexer unit's state. Finally, the instruction register updates so that each unit is set to represent a post-transition state of R, again with each line representing a different pre-transition state of PQRS. 

The clock updates (Table \ref{table:timekeeping_units}). 

\subsection*{Transition from update 3 to update 4} 
Figure \ref{fig:update4} shows the fourth of nine transitions involved in a simulation iteration. During this step, P and Q's next state propagate towards the registers, the next state of R is computed, and the computer becomes ready to compute S. 

At update 3, the multiplexer's OR unit ($\MO$) represented the next state of P, and was therefore ON. At update 4 the buffer unit which the OR outputs to ($\BUF{3}$) will therefore be ON. 

The multiplexer's OR now updates to reflect the next state of Q, just as it did for P at update 3. In our case, the OR is ON. 

The next state of R is now computed exactly as the next state of P and Q were. Because R transitions to ON when PQRS is 0101, exactly one AND unit in the multiplexer ($\MUX{11}$) is ON. 

The state of the data register outputs has not changed. This means that the same AND unit which was enabled at update 3 ($\MUX{11}$) will still be enabled at update 4, and the same program line ($\PRG{}{11}$) and instruction register ($\IR{11}$, though its output will of course change) will determine that multiplexer unit's state. Finally, the instruction register updates so that each unit is set to represent a post-transition state of S, again with each line representing a different pre-transition state of PQRS. 

The clock updates (Table \ref{table:timekeeping_units}). Unit $\CX{2}$, which has been OFF since initialization, is now ON. This disables the instruction register. 

\subsection*{Transition from update 4 to update 5} 
Figure \ref{fig:update5} shows the fifth of nine transitions involved in a simulation iteration. During this step, P, Q, and R's next state propagate towards the registers, and the next state of S is computed. 

At update 4, the COPY unit immediately downstream of the multiplexer ($\BUF{3}$) represented the next state of P, and was therefore ON. At update 5 this state will propagate to the next unit in the buffer chain ($\BUF{2}$). Similarly, at update 4, $\MO$ represented the next state of Q, and was therefore ON. At update 5, this state will propagate to $\BUF{3}$.  

$\MO$ now updates to reflect the next state of R, just as it did for Q at update 4. In our case, $\MO$ is ON. 

The next state of S is now computed exactly as the next state of P, Q, and R were. Because S transitions to OFF when PQRS is 0101, exactly zero AND units in the multiplexer will be ON at update 5. 

The clock updates (Table \ref{table:timekeeping_units}).

\subsection*{Transition from update 5 to update 6} 
Figure \ref{fig:update6} shows the sixth of nine transitions involved in a simulation iteration. 

At update 6, the inputs to each data register's $\REG{}{XOR}$ unit (i.e., $\BUF{1}$ to $\REG{1}{XOR}$, $\BUF{2}$ to $\REG{2}{XOR}$, $\BUF{3}$ to $\REG{3}{XOR}$, and $\MO$ to $\REG{4}{XOR}$) will hold that register's future state. In our case, this means that $\BUF{1}$, $\BUF{2}$, $\BUF{3}$, and $\MO$ will be ON, ON, ON, and OFF, respectively. 

The clock updates (Table \ref{table:timekeeping_units}).

\subsection*{Transition from update 6 to update 7} 
Figure \ref{fig:update7} shows the seventh of nine transitions involved in a simulation iteration. During this step, the registers determine whether their outputs need to change state. 

Each register's lower XOR unit ($\REG{1}{XOR}$, $\REG{2}{XOR}$, $\REG{3}{XOR}$, $\REG{4}{XOR}$) receives two inputs: one from its upper XOR ($\REG{1}{SIM}$, $\REG{2}{SIM}$, $\REG{3}{SIM}$, $\REG{4}{SIM}$ respectively), which holds the simulated current state of P, Q, R, or S, and one from another unit which holds the simulated next state of P, Q, R, or S ($\BUF{1}$, $\BUF{2}$, $\BUF{3}$, $\MO$ respectively). The XOR performs a ``same-or-different'' operation on these two inputs, effectively deciding whether a register's currently stored value is the same as its desired future value, or different. This is the same as deciding whether or not a register's current value needs to change. 

In our case, we are simulating a transition from state 0101 to state 1110. The state of each register's lower XOR is therefore ON (``change''), OFF (``no change''), and ON (``change''), and OFF (``change'') respectively. 

The clock updates (Table \ref{table:timekeeping_units}). $\CA{2}$ is now ON, which enables the AND units in each data register ($\REG{1}{AND}$, $\REG{2}{AND}$, $\REG{3}{AND}$, $\REG{4}{AND}$). 

\subsection*{Transition from update 7 to update 8} 
Figure \ref{fig:update8} shows the eighth of nine transitions involved in a simulation iteration. During the step, each register's ``change'' signal is allowed to influence its stored value. 

Normally, $\CA{2}$ is OFF. As a consequence, the data registers' AND units ($\REG{1}{AND}$, $\REG{2}{AND}$, $\REG{3}{AND}$, $\REG{4}{AND}$) are usually disabled. At update 8, however, the $\CA{2}$ was ON. This means that each register's AND unit was enabled, and is now ON if and only if its register's lower XOR was ON at update 7. In our case, these AND units are ON, OFF, ON, and ON respectively. The ``change'' signals from update 7 have thus propagated to these AND units at update 8, where they are able to influence the what the states of the data register outputs will be at update 9. 

The clock updates (Table \ref{table:timekeeping_units}). $\CA{2}$ returns to being OFF. $\CX{2}$ also switches OFF, which enables the instruction register to preemptively begin loading values for the next simulation iteration. 

\subsection*{Transition from update 8 to update 9} 
Figure \ref{fig:update9} shows the last transition involved in a simulation iteration. During this transition, the states of $\REG{1}{SIM}$, $\REG{2}{SIM}$, $\REG{3}{SIM}$, and $\REG{4}{SIM}$ update synchronously to become 1110, and the next simulation iteration begins. 

$\REG{1}{SIM}$, $\REG{2}{SIM}$, $\REG{3}{SIM}$, and $\REG{4}{SIM}$ each receive two inputs: one from itself, which carries its own current state, and one from the AND unit which determines whether that state needs to change. If the XOR's current state is ON and it gets a ``change'' signal (ON), then it turns OFF. If the XOR's current state is ON and it gets a ``no change'' signal (OFF), then it remains ON. If the XOR's current state is OFF and it gets a ``change'' signal (ON), then it turns ON. If the XOR's current state is OFF and it gets a ``no change'' signal (OFF), then it remains OFF. Because the AND units are normally all OFF (``no change''), the XORs normally hold their state. At update 8, however, some of these AND units may have been ON, and therefore the data register outputs at update 9 have changed their state, if necessary. 

In our case, the data register outputs have become 1110. Our first simulation iteration is complete, and the next one has begun. The process continues indefinitely.

\clearpage
\section*{Results}
\label{sec:results}
All notation and terminology below follows \cite{Albantakis2023} and \cite{Marshall2024}, which should be consulted for details of IIT's framework and formalism. In keeping with \cite{Albantakis2023} and \cite{Marshall2024}, $\log$ is always the base 2 logarithm. Results are organized as follows:
\renewcommand\contentsname{\vspace*{-30pt}}
\tableofcontents
\vspace*{\baselineskip}

The definition and results below build up to three key claims concerning (1) the potential complexes and bounds on $\Phi$ in the weakly connected computer at the micro grain (see Claim \ref{claim:wcc_complexes}); (2) the potential complexes and bounds on $\Phi$ in the strongly connected computer at the micro grain (see Claim \ref{claim:scc_complexes}); and (3) the inability of function-relevant macroings to replicate the cause--effect structure of the target system (see Claim \ref{claim:macro-result}).

\addcontentsline{toc}{subsection}{General notation, definitions, and results}
\subsection*{General notation, definitions, and results}
\label{subsec:notation_definitions_results}

\begin{definition}[Micro connectivity]
    For a model $U = (U_1, U_2, \ldots, U_n)$ with transition probabilities $p(u \mid \bar{u})$ for $u, \bar{u} \in \Omega_U$, the micro connectivity matrix, denoted $\mathcal{A}^{\mu}(U)$, is an $n \times n$ matrix where the $(i, j)$ cell ($a^{\mu}_{ij}$) indicates whether $U_i$ has an effect on $U_j$. Specifically, $a^{\mu}_{ij} = 1$ if there exists $u, \bar{u}$ and $\tilde{u}$ with $\bar{u}_k = \tilde{u}_k$ for $k \neq i$ and $\bar{u}_i \neq \tilde{u}_i$ such that $p(u_j \mid \bar{u}) \neq p(u_j \mid \tilde{u})$. Essentially, there is a directed connection from $U_i$ to $U_j$ if the current state of $U_i$ has an effect on $U_j$, for any fixed state of the rest of the units. 
\end{definition}

\begin{definition}[Paths]
     For a model $U = (U_1, U_2, \ldots, U_n)$ with micro connectivity matrix $\mathcal{A}^{\mu}(U)$ a path from $U_a$ to $U_b$ is a sequence of units $\wp(U_a, U_b) = (U_{c_1}, U_{c_2}, \ldots, U_{c_m})$ such that:  
    \begin{equation*}
        U_{c_1} = U_a, ~ U_{c_m} = U_b, ~ a^{\mu}_{c_ic_{i+1}} = 1 \quad \forall i = 1,\ldots, m-1.
    \end{equation*}  
\end{definition}

\begin{definition}[Generalized connectivity]
    A unit $J$ connects to a unit $K$ (denoted $J \out K$) if there exists a micro path between units $J$ and $K$, consisting entirely of micro units that are either constituents or apportionments of $J$. In other words
    \begin{equation*}
        J \out K \iff \exists \wp^{\mu}(U_a, U_b) : \wp(U_a, U_b) \setminus U_b \subseteq U^J \cup W^J ~ \wedge ~ U_a \in U^J ~ \wedge ~ U_b \in U^K.
    \end{equation*}
\end{definition}

\begin{definition}[Generalized connectivity matrix]
    For a system $S = (J_1, J_2, \ldots, J_{|S|})$ the generalized connectivity matrix of $S$, denoted $\mathcal{A}(S)$, is an $|S| \times |S|$ matrix where the $(i, j)$ cell ($a_{ik}$) indicates whether $J_i$ connects to $J_k$, 
    \begin{equation*}
        a_{ik} = \begin{cases} 1 & \text{ if } J_i \out J_k \\ 0 & \text{ otherwise.} \end{cases}
    \end{equation*} 
\end{definition}

\begin{definition}[Cut matrix]
    For a system $S = (J_1, J_2, \ldots, J_{|S|})$ and partition $\theta \in \Theta(S)$, the cut matrix $\mathcal{X}(\theta, S)$, is an $|S| \times |S|$ matrix where the $(i, k)$ cell ($x_{ik}$) indicates whether $\theta$ cuts the potential connection from $J_i$ to $J_k$. Let $S^{(i)}, S^{(k)} \in \theta$ be the parts containing $J_i$ and $J_k$ respectively, then  
    \begin{equation*}
        x_{ik} = \begin{cases} 0 \text{ if } S^{(i)}_{\leftarrow} \in \theta \text{ or } S^{(k)}_{\rightarrow} \in \theta \text{ or } S^{(i)} = S^{(k)} \\ 1 \text{ otherwise. } \end{cases}
    \end{equation*}
\end{definition}

\begin{definition}[$\Gamma$-notation]
    Given a system $S = (S_1, S_2, \ldots, S_n)$ with generalized connectivity matrix $\mathcal{A}(S)$, let $\Gamma^{\inp}_{S}(S_i) \subseteq S$ be unit $S_i$'s \emph{system predecessors}---the set of units that act upon $S_i$, with all background units removed: 
    \begin{equation*}
        \Gamma_S^{\inp}(S_i) = \{S_j : S_j \in S, a_{ji} = 1\}.
    \end{equation*}
    Similarly, let $\Gamma^{\out}_{S}(S_i)$ denote $S_i$'s \emph{system successors}:
    \begin{equation*}
        \Gamma_S^{\out}(S_i) = \{S_j : S_j \in S, a_{ij} = 1\}.
    \end{equation*}
    
    Finally, we can overload these operators to accept as arguments a set of units $V$. For example, let $\Gamma^{\inp}_{S}(V) \subseteq S$ be the \emph{system predecessor union} of $V$ within $S$:
    \begin{equation*}
        \Gamma^{\inp}_{S}(V) = \bigcup_{V_i \in V}\Gamma^{\inp}_{S}(V_i).
    \end{equation*}
\end{definition}

\begin{claim}
\label{claim:weakly_connected_reducible}
    Systems that are not strongly connected are reducible.
\end{claim}
\begin{proof}
    Let $S$ be a system in any state $s \in \Omega_S$. If $S$ is not strongly connected, then there exists $F \subset S$ such that $\Gamma^{\inp}_{S}(F) \cap (S \setminus F) = \varnothing$. Consider a partition $\theta = \{F^{\inp}, (S\setminus F)^{\out}\} \in \Theta(S)$, where we noise all incoming connections from $S\setminus F$ to $F$. Since there are no incoming connections to $F$ from $S \setminus F$, the system's transition probabilities remain the same before and after partitioning (i.e., $\T_e = \T_e^\theta$ and $\T_c = \T_c^\theta$; see Eqns. 16-18 in \cite{Albantakis2023}). Therefore, $\theta' = \{F_{\inp},~ {S \setminus F}_{\out}\}$ is a minimum partition, and the system is reducible: $\varphi_s(s, \theta') = 0$.
\end{proof}

\begin{definition}[Externally determined units]
    Let $S = (S_1, S_2, \ldots, S_m)$ be a system in state $s$, and $W = U \setminus U^S$ be background conditions in state $w$. A unit $s_i \in s$ is \emph{externally determined} if: 
    \begin{equation*}
        p_e(\bar{s_i} \mid s) = p_e(\bar{s_i}), \; \forall \bar{s}, s \in \Omega_S
        \quad \vee \quad
        p_c(s_i \mid \bar{s}) = p_c(s_i), \; \forall \bar{s}, s \in \Omega_S.
    \end{equation*} 
    
    Essentially, a unit is externally determined if either its current or future state are fully determined by the background conditions of the system.  Note that whether a unit is externally determined depends on the state of the background units $w$, but not on the system state $s$.

    An example is a deterministic AND unit $S_i$ receiving input from a background unit which is 0. It does not matter what $s$ is: the causal marginalization of $w$, conditional on $u$, guarantees that $\bar{s_i} = 0$.

    Another example is a deterministic OR unit $S_i$ receiving input from a background unit which is 1. It does not matter what $s$ is: the causal marginalization of $w$, conditional on $u$, guarantees that $\bar{s_i} = 1$. 

    These examples illustrate that it is not the actual value (0 or 1) of the next state $\bar{s_i}$ that defines whether or not a unit is externally determined---it is about whether anything with in the system can affect that state. 
\end{definition}

\begin{claim}
\label{claim:exdet_reducible}
    Systems with externally determined units are reducible.
\end{claim}
\begin{proof}
    Let $S$ be a system in any state $s \in \Omega_S$ with background conditions $w$. Let $S_i \in S$ be externally determined, then consider a partition of $S$ where the first part consists of the externally determined unit with its inputs cut, and the second part is its complement:
    \begin{equation*}
        \theta = \{\{S_i\}_{\inp},~ S \setminus \{S_i\}_{\out}\}
    \end{equation*}
    The partitioned probability matrix is unaffected by this partition since the probability distribution of $S_i$ does not depend on the state of $S$ (either $\T_c = \T_c^\theta$ or $\T_e = \T_e^\theta$, depending on whether the current or future state is determined; see Eqns. 16-18 in \cite{Albantakis2023}) and thus 
    \[ \min\{\varphi_c(s, \theta), \varphi_e(s, \theta)\} = 0. \] Therefore, $\theta' = \{\{S_i\}_{\inp},~ S \setminus \{S_i\}_{\out}\}$ is a minimum partition, and the system is reducible: $\varphi_s(s, \theta') = 0$.  
\end{proof}

\begin{claim}
\label{claim:single_in_single_out_no_hom}
    For any system $S = (S_1, S_2, \ldots, S_n)$, if unit $S_i$ has a single system successor $S_j$, and unit $S_j$ has a single system predecessor $S_i$, then neither $S_i$ nor $S_j$ can contribute to higher-order mechanisms. 
\end{claim}
\begin{proof}
    Assume $S_i$ has a single system successor $S_j$ (i.e., $\Gamma^{\out}_{S}(S_i) = S_j$) and $S_j$ has a single system predecessor $S_i$ (i.e., $\Gamma^{\inp}_{S}(S_j) = S_i$). Consider a higher-order mechanism $M \subseteq S$ in any state and a potential purview $Z \subseteq S$. If $S_i \in M$, we can define a disintegrating partition $\theta \in \Theta(M, Z)$ that cuts no connections from $M$ to $Z$. If $S_j \in Z$, then we can define 
    \[ \theta = \{(S_i, S_j), (M\setminus S_i, Z\setminus S_j)\} \]
    or if $S_j \notin Z$, then we can define 
    \[ \theta = \{ (S_i, \varnothing), (M\setminus S_i, Z) \}. \]
    In either case, we have a partition $\theta$ that does not cut any connections from $M$ to $Z$, and so $\pi_e(z\mid m) = \pi_e(z)$ for all $z \in \Omega_Z$ and $m \in \Omega_M$, and thus 
    \[ 
        \varphi_e(m) = \varphi_e(m,Z,\theta) = \pi_e(z\mid m)\pospart{\log\frac{\pi_e(z \mid m)}{\pi_e(z)}} = 0,
    \]
    and $M$ is reducible. Note that the above partition is only valid for $|M| > 1$, and thus it is still possible that $M = S_i$ forms a first-order mechanism. 

    Similarly, if $S_j \in M$, we can define a partition $\theta \in \Theta(M, Z)$ that cuts no connections from $Z$ to $M$. If $S_i \in Z$, then we can define 
    \[ \theta = \{ (S_j, S_i), (M\setminus S_j, Z\setminus S_i) \}, \]
    or if $S_i \notin Z$, then we can define 
    \[ \theta = \{ (S_j, \varnothing), (M\setminus S_j, Z) \}. \]
    In either case, we have a partition $\theta$ that does not cut any connections from $Z$ to $M$, and so $\pi_c(z\mid m) = \pi_c(z)$ for all $z \in \Omega_Z$ and $m \in \Omega_M$, and thus 
    \[ 
        \varphi_c(m) = \varphi_c(m,Z,\theta) = \pi_c^\leftarrow(z\mid m)\pospart{\log\frac{\pi_c(z \mid m)}{\pi_c(z)}} = 0, 
    \]
    and $M$ is reducible.
\end{proof}

\addcontentsline{toc}{subsection}{Analysis of the ring motif}
\subsection*{Analysis of the ring motif}
\label{subsec:ring_motif}

\begin{definition}[Rings]
    A system $S = (S_1, S_2, \ldots, S_n)$ is a ring if and only if all of the following conditions hold:
    \begin{enumerate}
        \item S is strongly connected
        \item For each $S_k \in S$, there exists a unique $i \neq k$  such that $\Gamma_{S}^{\inp}(S_k) = S_i$ 
    \end{enumerate}
\end{definition}

\begin{claim}
\label{claim:copy_ring_distinctions}
    Let $S$ be a ring in state $s$. The cause--effect structure of $s$ contains at most $|S|$ distinctions, each of which is a first order distinction with $\varphi_d \leq 1$. 
\end{claim}
\begin{proof}
    Let $m = s_j \in s$ be a potential first order mechanism. Because $S$ is a ring, $S_j$ has a single input $\Gamma_S^\inp(S_j)$ and a single output $\Gamma_S^\out(S_j)$. If $s_j$ does not specify the cause state of its input, then the ring can be cut $\theta = \{\{S_j\}_{\inp}, (S \setminus S_j)_{\out}\}$ and $\varphi_s(s, \theta) = 0$. A similar argument applies to the effect state of it's output. Therefore, if $\varphi_s(s) > 0$, then $s_j$ specifies the cause state of $\Gamma_S^\inp(S_j)$ and the effect state of $\Gamma_S^\out(S_j)$. Since $S_j$ has no other inputs or outputs, the only possible purviews are $Z_c^* = \Gamma_S^\inp(S_j)$ and $Z_e^* = \Gamma_S^\out(S_j)$. Thus, the distinction is   
    \[ d(m) = (m = s_j, z^*_c = \Gamma_S^\inp(s_j),  z^*_e = \Gamma_S^\out(s_j), \varphi_d(m)). \] 
    As demonstrated in \cite{Zaeemzadeh2024b}, since $|\Gamma_S^\inp(S_j)| = |\Gamma_S^\out(S_j)| = 1$, 
        \[ \varphi_d(m) \leq 1. \]
    Also note that the information specified by the system as a whole about each unit's cause and effect is congruent with the information specified by each unit about its maximal cause--effect state. Therefore, all $|S|$ first-order distinctions exist in the cause--effect structure.
    
    Finally, no higher-order distinctions exist. This is because no units in the ring have both shared inputs and shared outputs (see Claim \ref{claim:single_in_single_out_no_hom}). Therefore, the set of distinctions is exactly the set of first order distinctions just described.
\end{proof}

\begin{claim}
\label{claim:copy_ring_relations}
    Let $S$ be a ring in state $s$ with $\varphi_s(s) > 0$, and let $|S| \geq 4$. The cause--effect structure of $s$ contains at most $|S|$ unique sets of distinctions that are related and 
    \[ \sum_{\bm{d}} \varphi_r(\bm{d}) \leq \frac{|S|}{2}. \] 
 
\end{claim}
\begin{proof}
    Let $\D(s)$ be the set of distinctions in the cause--effect structure (see Claim \ref{claim:copy_ring_distinctions}), $\bm{d} \subseteq \D(s)$ be a \emph{set} of distinctions that may be related, $d \in \D(s)$ be a single distinction, and $M(d)$ the mechanism of $d \in D(s)$.  

    First, consider $|S| = 4$ as a unique case. By direct inspection of the potential distinctions, there are at most two possible relations, each with up to two faces and 
        \begin{equation*}
        \begin{split}
            \varphi_r(\bm{d}) &= \min_{d \in \bm{d}} \left\|\bigcup_{f \in \bm{f(d)}} o^*_f\right\|\frac{\varphi_d(d)}{|z^*_c(d) \cup z^*_e(d)} \\
            &= \min_{d \in \bm{d}} (2)\frac{\varphi_d(d)}{2} \\
            &= \min_{d \in \bm{d}} \varphi_d(d) \leq 1, 
        \end{split}
    \end{equation*}
    and thus $\sum_{\bm{d}} \varphi_r(\bm{d}) \leq 2 = |S|/2$.  
    
    Next, consider $|S| > 4$. Because of the ring connectivity of $S$, there are guaranteed to be exactly $|S|$ pairs of units that are ``nearly neighbors'' (i.e., separated from each other in the ring by a single, intervening unit). Call the set of the corresponding distinction pairs $\mathrm{NN}$:
    \begin{equation*}
            \mathrm{NN} = \{d_i, d_j \in D(s) ~:~ \Gamma_S^\out(M(d_i)) = \Gamma_S^\inp(M(d_j))\}  
    \end{equation*}
    For each of these pairs, the cause purview of $d_j$ will overlap with the effect purview of $d_i$. If $z_c^*(d_j) = z_e^*(d_i)$, the cause and effect are congruent, then $\bm{d} = \{d_i, d_j\}$ constitutes a 2\ts{nd}-degree relation (because 2 distinctions, $d_i$ and $d_j$, are bound) with a 2\ts{nd}-degree face (because 2 purviews, $z^*_c(d_j)$ and $z^*_e(d_i)$, are involved). 
    
    Whether or not all $|S|$ of the possible relations $r(\bm{d}),~ \bm{d} \in \mathrm{NN}$ exist depends on whether all pairs of distinctions $\{d_i,~ d_j\} \in \mathrm{NN}$ agree on the state of their overlapping purview, i.e., on whether $z^*_c(d_j) = z^*_e(d_i)$. This will generally not be the case, so $|S|$ is an upper bound on the possible number of relations. For each relation $r(\bm{d})$ that exists, $\varphi_r \leq 0.5$,
    \begin{equation*}
        \begin{split}
            \varphi_r(\bm{d}) &= \min_{d \in \bm{d}} \left\|\bigcup_{f \in \bm{f(d)}} o^*_f\right\|\frac{\varphi_d(d)}{|z^*_c(d) \cup z^*_e(d)|} \\
            &= \min_{d \in \bm{d}} (1)\frac{\varphi_d(d)}{2} \\
            &\leq \frac{1}{2}
        \end{split}
    \end{equation*}
    
    No other relations are possible (e.g., 3\ts{rd}-degree relations, 4\ts{th}-degree relations, etc.), because there are no higher-order purviews. A candidate relation involving 3 or more distinctions will have non-overlapping purviews. Thus
    \begin{equation*}
        \sum \varphi_r \leq \sum \frac{1}{2} \leq \frac{|S|}{2}.
    \end{equation*}
\end{proof}

\begin{claim}
\label{claim:copy_ring_bigphi}
    Let $S$ be a ring in state $s$ with $\varphi_s(s) > 0$, and let $|S| \geq 4$. The structure integrated information $\Phi(s)$ is bounded above by $\frac{3}{2}|S|$ ibits.
\end{claim}
\begin{proof}
    $\Phi(s)$ is the sum of the values of integrated information of a substrate's distinctions and relations (see Eqn. 59 in \cite{Albantakis2023}). 
    
    Per Claim \ref{claim:copy_ring_distinctions}, distinctions will always contribute a maximum of $|S|$ ibits to $\Phi(s)$:
    \begin{equation*}
        \sum_{d \in \D(s)} \varphi_d \leq \\
        \sum_{d \in \D(s)} 1 = \\
        |S| \text{ ibits.} \\
    \end{equation*}
    
    Per Claim \ref{claim:copy_ring_relations}, the contribution of relations to $\Phi(s)$ will be maximized when $s$ is homogenous (i.e., all 0 or all 1), so that all maximal cause--effect states are congruent with each other. In this case, relations will contribute at most $\frac{|S|}{2}$ ibits to $\Phi(s)$:
    \begin{equation*}
        \sum_{\bm{d} \in \mathrm{NN}} \varphi_r(\bm{d}) \leq \\
        \sum_{\bm{d} \in \mathrm{NN}} \frac{1}{2} = \\
        \frac{1}{2}|S| \text{ ibits.} \\
    \end{equation*}
    for a total of:
    \begin{equation*}
        \Phi(s) = \sum_{d \in \D(s)} \varphi_d + \sum_{\bm{d} \in \mathrm{NN}} \varphi_r(\bm{d}) \leq |S| + \frac{1}{2}|S| = \frac{3}{2}|S| \text{ ibits.}
    \end{equation*} 
\end{proof}

\begin{definition}[Imperfect Ring]
    An imperfect ring $S = S' \cup S_n$ is defined by starting from a ring $S' = (S_1, S_2, \ldots, S_{n-1})$ and modifying it in a specific way. Start with a ring $S'$ with $|S'| = n - 1$ and add an additional unit $S_n$ such that $S_n$ outputs to $S_1$ and receives input from $S_{n - 2}$: 
    \[ \Gamma_S^\out(S_n) = S_1, ~ \Gamma_S^\inp(S_n) = S_{n - 2}. \]
    The modification has the following impact on the existing ring units: 
    \[ \Gamma_S^\inp(S_1) = \{S_{n - 1}, S_{n}\}, ~ \Gamma_S^\out(S_{n - 2}) = \{S_{n - 1}, S_n\}. \]
\end{definition}

\begin{claim}
\label{claim:imperfect_copy_ring_distinctions}
    Let $S'$ be a ring with $|S'| = n - 1$, and $S = S' \cup S_n$ an imperfect ring in state $s$. The cause--effect structure of $s$ contains at most $|S| + 1$ distinctions, $|S|$ potential first-order distinctions, and one potential second-order distinction, each with $\varphi_d \leq 1$. 
\end{claim}

\begin{proof}
    The first-order distinctions specified by $S$ are those potentially specified by $S'$, with two possible differences: The distinction specified by $s_1$ now has the potential for a larger cause purview, $Z_c^*(m) \subseteq \{s_{n - 1}, s_n\}$, and the distinction specified by $s_{n - 2}$ now has the potential for a larger effect purview, $Z_e^*(s_{n-2}) \subseteq \{s_{n - 1}, s_n\}$. However, regardless of whether this actually occurs, and regardless of whether $m = s_1$ or $m = s_{n - 2}$: 
        \[ \min(|Z_c^*(m)|, |Z_e^*(m)|) = 1.\]
    This means that, for both $m = s_1$ and $m = s_{n - 2}$, $\varphi_d(m) \leq 1$.
    
    It remains to check for potential distinctions whose mechanism includes $S_n$. For the first order $m = s_n$, the potential cause purview is $Z_c^* = s_{n - 2}$ and the potential effect purview is $Z_e^* = s_1$, and thus $\varphi_d(m) \leq 1$. 

    Next, $S_{n - 1}$ and $S_{n}$ have common input ($S_{n - 2}$) and common output ($S_1$), so they can potential specify a higher order mechanism. If $m = \{s_{n - 1}, s_n\}$ then the potential cause purview is $Z_c^* = s_{n - 2}$ and the potential effect purview is $Z_e^* = s_1$, and thus $\varphi_d(m) \leq 1$. 
    
    Finally, no other higher-order distinctions exist. This is because no sets of units in the imperfect ring have both shared inputs and shared outputs (see Claim \ref{claim:single_in_single_out_no_hom}).
\end{proof}

\begin{claim}
\label{claim:imperfect_copy_ring_relations}
    Let $S'$ be a ring with $|S'| = n - 1 \geq 4$ and $S = S' \cup S_n$ an imperfect ring in state $s$. The cause--effect structure $C(s)$ contains at most $|S| + 15$ unique sets of distinctions relating distinctions in $D(s)$ and 
    \[ \sum_{\bm{d}} \varphi_r(\bm{d}) \leq \frac{|S| - 2}{2} + \frac{64}{6}. \]
\end{claim}
\begin{proof}
    Let $\D(s)$ be the set of distinctions in the cause--effect structure (see Claim \ref{claim:imperfect_copy_ring_distinctions}), $\bm{d} \subseteq \D(s)$ be a \emph{set} of distinctions that may be related, $d \in \D(s)$ be a single distinction, and $M(d)$ the mechanism of $d \in D(s)$.  

    First, consider $|S'| = 4$ as a unique case. By direct inspection of the potential distinctions, there are at most 11 possible relations, each with up to two faces and $\varphi_r \leq 1$. Thus, 
    \begin{equation*} 
        \sum_{\bm{d}} \varphi_r(\bm{d}) \leq 11 < \frac{|S| - 2}{2} + \frac{64}{6} = 12.17.
    \end{equation*}
    
    Next, consider $|S'| > 4$. Consider the distinctions in $D(s')$. Those distinctions are also in $D(s)$, with the exception being a potentially larger cause purview for $m = s_1$ and a potentially larger effect purview for $m = s_{n - 2}$. Thus, all relations specified in Claim \ref{claim:copy_ring_distinctions} are also relations among distinctions in $D(s)$, except for the relation among $\bm{d} = \{d(m = s_1), d(m = s_{n - 2})\}$. That is, there are potentially $n - 2$ 2\ts{nd}-degree relations with a single 2\ts{nd}-degree face and $\varphi_r \leq 0.5$.
    
    Any further relations must include either the cause of $d(m = s_1)$, the effect of $d(m = s_{n - 2})$, or one of the new distinctions $d(m = s_n)$, $d(m = \{s_{n - 1}, s_n\})$. The possible relations for such a small set of distinctions can be computed explicitly, and are summarized in Table \ref{tab:relations}. The faces and overlap can be extracted from the necessary features of the distinctions, and then the maximum possible value of $\varphi_r$ is obtained based on the maximum possible $\varphi_d$ of the corresponding distinctions. 

    Combined, there are at most $|S| - 2$ relations contributing at most $\varphi_r = \frac{1}{2}$ each, plus the $\varphi_r$ of the relations listed in Table \ref{tab:relations}:
    \begin{equation*}
        \sum_{\bm{d}} \varphi_r(\bm{d}) \leq \\
         \frac{|S|-2}{2} + \frac{64}{6} \text{ ibits.} \\
    \end{equation*}
\end{proof}

\begin{table}[!ht]
    \centering
    \begin{tabular}{|c|c|c|c|}
    \hline
         $\bm{d}$ & faces & overlap & $\max \varphi_r$ \\
         \hline
         $d(s_1), d(s_{n - 2})$ & cause--effect & $\{s_{n-1}, s_n\}$ & 2/3 \\ \hline
         $d(s_2), d(s_{n})$ & cause--effect & $s_1$ & 1/2 \\ \hline
         $d(s_{n}), d(s_{n-3})$ & cause--effect & $s_{n-2}$ & 1/2 \\ \hline
         $d(s_{2}), d(\{s_{n - 1}, s_n\})$ & cause--effect & $s_1$ & 1/2 \\ \hline
         $d(\{s_{n - 1}, s_n\}), d(s_{n - 3})$ & cause--effect & $s_{n-2}$ & 1/2 \\ \hline
         $d(s_{n-1}), d(s_{n})$ & \makecell{cause--cause \\ effect--effect} & \makecell{$s_{n - 2}$ \\ $s_1$} & 1 \\ \hline
         $d(\{s_{n - 1}, s_n\}), d(s_{n - 1})$ & \makecell{cause--cause \\ effect--effect} & \makecell{$s_{n-2}$ \\ $s_{1}$} & 1 \\ \hline
         $d(s_n), d(\{s_{n - 1}, s_n\})$ & \makecell{cause--cause \\ effect--effect} & \makecell{$s_{n-2}$ \\ $s_1$} & 1 \\ \hline
         $d(s_2), d(s_n), d(\{s_{n - 1}, s_n\})$ & cause--effect--effect & $s_1$ & 1/2 \\ \hline
         $d(s_2), d(s_n), d(s_{n - 1})$ & cause--effect--effect & $s_1$ & 1/2 \\ \hline
         $d(s_2), d(\{s_{n - 1}, s_{n}\}), d(s_{n - 1})$ & cause--effect--effect & $s_1$ & 1/2 \\ \hline
         $d(s_n), d(\{s_{n - 1}, s_n\}), d(s_{n - 3})$ & cause--cause--effect & $s_{n - 2}$ & 1/2 \\ \hline
         $d(s_n), d(s_{n-1}), d(s_{n-3})$ & cause--cause--effect & $s_{n-2}$ & 1/2 \\ \hline
         $d(s_{n - 1}), d(\{s_{n - 1}, s_n\}), d(s_{n-3})$ & cause--cause--effect & $s_{n-2}$ & 1/2 \\ \hline
         $d(s_{n-1}), d(s_n), d(\{s_{n - 1}, s_n\})$ & \makecell{cause--cause--cause \\ effect--effect--effect} & \makecell{$s_{n - 2}$ \\ $s_1$} & 1 \\ \hline
         $d(s_2), d(s_{n-1}), d(s_n), d(\{s_{n - 1}, s_n\})$ & cause--effect--effect--effect & $s_1$ & 1/2 \\ \hline
         $d(s_{n-1}), d(s_n), d(\{s_{n - 1}, s_n\}), d(s_{n-3})$ & cause--cause--cause--effect & $s_{n-2}$ & 1/2 \\ \hline
    \end{tabular}
    \caption{Potential relations involving $S_n$ for an imperfect ring of at least five units.}
    \label{tab:relations}
\end{table}

\begin{claim}
\label{claim:imperfect_copy_ring_bigphi}
    Let $S'$ be a ring with $|S'| = n - 1$ and $S = S' \cup S_n$ be an imperfect ring in state $s$, and let $|S| \geq 5$. The structure integrated information $\Phi(s)$ is bounded above by 
    \[ \frac{9n + 64}{6} \text{ ibits}. \]
\end{claim}

\begin{proof}
    $\Phi(s)$ is the sum of the values of integrated information of a substrate's distinctions and relations (see Eqn. 59 in \cite{Albantakis2023}). 
    
    Per Claim \ref{claim:imperfect_copy_ring_distinctions}, distinctions will always contribute a maximum of $|S|$ ibits to $\Phi(s)$:
    \begin{equation*}
        \sum_{d \in \D(s)} \varphi_d \leq \\
        \sum_{d \in \D(s)} 1 \leq \\
        |S + 1| \text{ ibits.} \\
    \end{equation*}
    
    Per Claim \ref{claim:imperfect_copy_ring_relations}, there are at most $n - 2$ relations contributing at most $\varphi_r = \frac{1}{2}$ each, plus the $\varphi_r$ of the relations listed in Table \ref{tab:relations}:
    \begin{equation*}
        \sum_{\bm{d}} \varphi_r(\bm{d}) \leq \\
         \frac{n-2}{2} + \frac{64}{6} \text{ ibits.} \\
    \end{equation*}
    for a total of:
    \begin{equation*}
        \Phi(s) = \sum_{d \in \D(s)} \varphi_d + \sum_{\bm{d}} \varphi_r(\bm{d}) \leq (n + 1) + \frac{n - 2}{2} + \frac{64}{6} = \frac{9n + 64}{6} \text{ ibits.}
    \end{equation*} 
\end{proof}

\addcontentsline{toc}{subsection}{Analysis of the monad motif}
\subsection*{Analysis of the monad motif}
\label{subsec:monad_motif}

\begin{definition}[Monad]
\label{def:monad}
    A system $S$ is a monad if and only if all of the following conditions hold:
    \begin{enumerate}
        \item $|S| = 1$ ($S$ is a single unit). 
        \item $S = \Gamma^\inp_S(S) = \Gamma^\out_S(S)$.
    \end{enumerate}
\end{definition}

\begin{claim}
\label{claim:monad_max_big_phi}
    For a monad $S$ in state $s$ with $\varphi_s(s) > 0$, the cause--effect structure of $s$ has the following properties:
    \begin{itemize}
        \item Exactly one distinction, of 1\ts{st} order, with $\varphi_d \leq 1$ ibit.
        \item At most one relation, of 1\ts{st}-degree with a 2\ts{nd}-degree face, and $\varphi_r = \varphi_d$ ibits. 
        \item $\Phi(s) \leq 2$ ibits. 
    \end{itemize}
\end{claim}
\begin{proof}
    Because $S$ is a monad it can have at most a single, first order distinction $d$, with mechanism $m$. As demonstrated in \cite{Zaeemzadeh2024b}, since $|\Gamma_S^\inp(m)| = |\Gamma_S^\out(m)| = 1$, 
        \[ \varphi_d(m) \leq 1. \]
        
    Because $S$ is a monad, $m = S$, and the information specified by the system as a whole about $m$'s cause and effect is necessarily congruent with the information specified by $m$ about its maximal cause--effect state. Therefore, $m$ must exist in the cause--effect structure. 

    The cause--effect structure of $S$ may contain no relations, for example if $d$ specifies a cause state that is incongruent with its effect state. If relations are present, then there can be at most a single relation $r$ between $\bm{d}=\{d\}$ ($r$ must be of 1\ts{st}-degree, with a 2\ts{nd}-degree face). Furthermore, it is trivially true that the union of unique purview units (see Eqn. 53 of \cite{Albantakis2023}) and the relation purview (see Eqn. 54 of \cite{Albantakis2023}) of $r$ cannot be more than a single unit. By Eqn. 55 in \cite{Albantakis2023},
    \begin{equation*}
        \max \varphi_r(\bm{d}) = \varphi_d(m)
    \end{equation*}
    So, if $r$ exists at all, $\Phi(s) = \varphi_d(m) + \varphi_r(\bm{d}) = 2\varphi_d(m)$. It follows from the upper bound on $\varphi_d(m)$ that $\Phi(s) \leq 2$ ibits. 
\end{proof}

\addcontentsline{toc}{subsection}{Analysis of the computer}
\subsection*{Analysis of the computer}
\label{sec:analysis_of_computer}

\begin{definition}[The weakly connected computer]
\label{def:wcc}
    Let $\WCC(k),~ k \in \{2, 3, 4, \ldots\}$ be the $k$\ts{th} \emph{weakly connected computer}, which simulates any system of $n = 2^k$ boolean units. A reference implementation with precise details of every unit and connection in $\WCC(k)$ can be found at \href{https://github.com/CSC-UW/Findlay_et_al_2024a}{https://github.com/CSC-UW/Findlay\_et\_al\_2024a}. The $2\ts{nd}$ (i.e., four-bit) weakly connected computer is shown in Fig. \ref{fig:computer_with_labeled_units}. Briefly, $\WCC(k)$ has the following components:
    \begin{itemize}
        \item $n = 2^k$ data registers $\textbf{R} = \{\REG{1}{}, \REG{2}{}, \ldots, \REG{n}{}\}$, ordered from left to right. Each register $\REG{i}{}$ contains 3 units: $\REG{i}{} = \{\REG{i}{SIM}, \REG{i}{AND}, \REG{i}{XOR}\}$. As useful shorthand, let $\REG{}{SIM} = \{\REG{i}{SIM}\}_{i=1}^{n}$, since we often need to refer to this set of units.  
        
        \item $2^n$ program lines $\textbf{P} = \{\PRG{}{1}, \PRG{}{2}, \ldots, \PRG{}{2^n}\}$, ordered from top to bottom. Each program line is a ring of $n$ COPY units: $\PRG{}{i} = \{\PRG{1}{i}, \PRG{2}{i}, \ldots, \PRG{n}{i}\}$.

        \item An instruction register $\textbf{IR} = \{\IR{1}, \IR{2}, \ldots, \IR{2^n}\}$, consisting of $2^n$ AND units, ordered from top to bottom.

        \item A multiplexer $\textbf{M} = \{\MUX{1}, \MUX{2}, \ldots, \MUX{2^n}, \MO\}$, consisting of $2^n$ AND units, ordered from top to bottom, and a generalized OR unit ($\MO$).
        
        \item A timekeeping chain $\textbf{T} = \{\CO, \CX{1}, \CA{1}, \CX{2}, \CA{2}, \ldots, \CX{k}, \CA{k}\}$, consisting of the core oscillator $\CO$ and $k$ frequency dividers $\{\CX{i}, \CA{i}\}$. 
        
        \item A buffer $\textbf{B} = \{\BUF{1}, \BUF{2}. \ldots, \BUF{2n-5}\}$, consisting of $2n-5$ COPY units between the multiplexer's output ($\MO$) and the leftmost register $\REG{1}{}$. 
    \end{itemize}

    Altogether, $\WCC(k) = \textbf{R} \cup \textbf{P} \cup \textbf{IR} \cup \textbf{M} \cup \textbf{T} \cup \textbf{B}$.  
\end{definition}

\begin{definition}[The strongly connected computer]
\label{def:scc}
    Let $\SCC(k),~ k \in \{2, 3, 4, \ldots\}$ be the $k\ts{th}$ \emph{strongly connected computer}, which simulates any system of $n = 2^k$ boolean units. A reference implementation with precise details of every unit and connection in $\SCC(k)$ can be found at \href{https://github.com/CSC-UW/Findlay_et_al_2024a}{https://github.com/CSC-UW/Findlay\_et\_al\_2024a}. Main text Fig. 3E illustrates how the $2\ts{nd}$ (i.e., four-bit) strongly connected computer is constructed from its weakly connected counterpart. Briefly, $\SCC(k)$ is constructed by making the following modifications to $\WCC(k)$:
    \begin{itemize}
        \item The second-to-last unit in each program line, $\PRG{n-1}{i},~i\in\{1,2,\ldots,2^n\}$ becomes an OR unit, and a new input to this unit from $\IR{i}$ is introduced. 

        \item The core oscillator $\CO$ becomes an OR unit, and a new negated input is added to this unit from every data register's AND unit, $\REG{j}{AND},~j\in\{1,2,\ldots,n\}$.
    \end{itemize}
\end{definition}

\begin{definition}[State sequences, feasibility, and reachability]
\label{def:reachable}
    For a model $U$ with transition probabilities $p(u \mid \bar{u})$ for $u, \bar{u} \in \Omega_U$. Following \cite{Marshall2024}, we define a sequence of micro updates, from $t = a$ to $t = b$ as  
    \begin{equation*}
        u_{[a, b]} = (u_{a}, u_{a + 1}, \ldots, u_{b - 1}, u_b) \qquad a,b \in \mathbb{Z},~ a < b.
    \end{equation*}
    The sequence's length is defined as:
    \begin{equation*}
        |u_{[a, b]}| = b - a + 1 = \tau \qquad \tau \in \mathbb{N}^+.
    \end{equation*}
    A sequence of updates is \emph{feasible} if 
    \begin{equation*}
        p(u_{i+1} \mid u_{i}) > 0 \quad \text{for all } u_i \in u_{[a, b]}, ~ a \leq i < b. 
    \end{equation*}
    A state $u_b$ is reachable from $u_a$ in exactly $\tau$ updates if there exists a feasible $u_{[a, b]}$ with $|u_{[a, b]}| = \tau$. The set of all states reachable from $u_a$ in exactly $\tau$ updates, and the set of all states generally reachable from $u_a$, are therefore given by
    \begin{align}
        R(u_a, \tau) &= 
            \{u_b \in \Omega_U : \text{there exists a feasible sequence } u_{[a, b]},~ |u_{[a, b]}| = \tau \} \\
        R(u_a) &= 
            \bigcup_{\tau}R(u_a, \tau) \qquad \tau \in \mathbb{N}^+.
    \intertext{We can also overload this operator to accept as its argument a set of states $\Omega_{U}' \subseteq \Omega_U$,}
        R(\Omega_{U}', \tau) &= 
            \{u_b \in \Omega_U : \text{there exists a feasible sequence } u_{[a, b]},~ |u_{[a, b]}| = \tau,~ u_a \in \Omega_{U}'\} \\
        R(\Omega_{U}') &= 
            \bigcup_{\tau}R(\Omega_{U}', \tau) \qquad \tau \in \mathbb{N}^+.
    \end{align}
\end{definition}

In order for $\WCC(k)$ or $\SCC(k)$ to simulate a target system, they must be initialized properly. A short description of this procedure is given in \href{subsec:update0}{\emph{Step-by-step guide to the computer's operation: Initialization}}, and a complete reference implementation is found at the GitHub link above. Let $U \in \{\WCC(k), \SCC(k)\}$ be a computer of order $k$, and let $\Omega_0$ be that computer's set of valid initial states. For every conceivable target system $T$ of $n = 2^k$ boolean units, the proper initial state of $U$ to simulate $T$ is in $\Omega_0$. Because there are $2^{n2^n}$ possible systems $T$ (equivalently, there are this many possible programs), $2^n$ possible initial states of $T$, and one valid initial state of $U$ for each combination of these, $|\Omega_0| = 2^{n2^n+n}$. Following Definition \ref{def:reachable}, $R(\Omega_0)$ is the set of all states that $U$ could possibly visit in the course of any conceivable simulation. The set of all valid initial states, $\Omega_0$, and the set of all reachable states, $R(\Omega_0)$, are the same for both the strongly and weakly connected computers. 

\addcontentsline{toc}{subsubsection}{Micro grain analysis of the weakly connected computer}
\subsubsection*{Micro-grain analysis of the weakly connected computer}
\label{subsubsec:micro_analysis_of_wcc}

\begin{claim}
\label{claim:wcc_reducible}
    Let $U = \WCC(k)$ be the $k\ts{th}$ weakly connected computer with valid initial states $\Omega_0$. For every reachable state $u \in R(\Omega_0)$, $\varphi_s(u) = 0$.
\end{claim}
\begin{proof}
    $U$ is not strongly connected, because the clock and program are feedforward to the rest of $U$, so is reducible per Claim \ref{claim:weakly_connected_reducible}.
\end{proof}

\begin{claim}
\label{claim:wcc_program_complexes}
    Let $U = \WCC(k)$ be the $k\ts{th}$ weakly connected computer with valid initial states $\Omega_0$, and $n=2^k$. For every reachable state $u \in R(\Omega_0)$, $\textbf{P}$ forms $2^n$ separate $n$-unit complexes, each with $\Phi \leq \frac{3n}{2}$.
\end{claim}
\begin{proof}
    Per Claim \ref{claim:wcc_reducible}, the computer as a whole has $\varphi_s(u) = 0$. However, this does not mean that all subsystems $S \subset U$ must have $\varphi_s(s) = 0$. Any strongly connected subset of the computer could be a complex. The only strongly connected subsets of $U$ that include units from the program are the $2^n$ program lines in $\textbf{P}$, each of which is an $n$-unit COPY ring. It follows from Claims \ref{claim:weakly_connected_reducible} and \ref{claim:copy_ring_bigphi} that these form $2^n$ separate complexes, each with $\Phi < \frac{3n}{2}$ (e.g., main text Fig. 3C). 
\end{proof}

\begin{claim}
\label{claim:wcc_clock_complexes}
    Let $U = \WCC(k)$ be the $k\ts{th}$ weakly connected computer with valid initial states $\Omega_0$. For every reachable state $u \in R(\Omega_0)$, $\CO$, $\{\CX{i}\}_{i=1}^k$, and $\{\CA{i}\}_{i=1}^k$ form at most $2k + 1$ single-unit complexes.
\end{claim}
\begin{proof}
    Per Claim \ref{claim:wcc_reducible}, the computer as a whole has $\varphi_s(u) = 0$. However, this does not mean that all subsystems $S \subset U$ must have $\varphi_s(s) = 0$. Any strongly connected subset of the computer could be a complex. The only strongly connected subsets that include units from the timekeeping chain are the $2k + 1$ monads. 
\end{proof}
\begin{remark}
    Although not proven (or necessary) here, we could be more precise about the complexes that form: $\CO$ always, $\CX{1}$ through $\CX{k-1}$ always, $\CX{k}$ only if $\CA{k-1}$ is ON, and $\CA{i},~ i \in \{1, 2, \ldots, k\}$ only if all its inputs are ON. In other words, we are being charitable to the computer here and elsewhere, listing all the complexes that could form in any reachable state, even if they do not actually exist most of the time. 
\end{remark}

\begin{claim}
\label{claim:and_implies_ak}
    Let $U \in \{\WCC(k), \SCC(k)\}$ be an $2^k=n$-bit computer in a reachable state $u \in R(\Omega_0)$. Let $S \subseteq U$ be a system within $U$, in a reachable state $s \subseteq u$. If $S$ includes some data register's AND unit, then either it includes the output of the $k\ts{th}$ frequency divider or it is reducible:
    \begin{equation*}
        \exists i : \REG{i}{AND} \in S \implies \bigl(\CA{k} \in S \vee \varphi_s(s) = 0\bigr)
    \end{equation*}        
\end{claim}
\begin{proof}
    Let $S \subset U$ be a candidate system and without loss of generality, $\REG{1}{AND} \in S$. Assume $\CA{k} \notin S$. There are two cases:
    \begin{enumerate}
        \item Case 1: $\CA{k}$ is 0. Then $\REG{1}{AND}$ is externally determined. It follows from Claim \ref{claim:exdet_reducible} that $S$ is reducible with $\varphi_s(s) = 0$.  
        \item Case 2: $\CA{k}$ is 1. When evaluating causes, knowledge of the current state of $u$ is used to compute the probability distribution over potential prior states of the background units (Eqn. 5 in \cite{Albantakis2023}). The past state of $\CA{k}$ was 0 with probability one. Thus the current state of $\REG{1}{AND}$ is externally determined. It follows from Claim \ref{claim:exdet_reducible} that $S$ is reducible with $\varphi_s(s) = 0$. 
    \end{enumerate}
    Both cases lead to $\varphi_s(s) = 0$, therefore either $\CA{k} \in S$ or $\varphi_s(s) = 0$.
\end{proof}

\begin{claim}
\label{claim:wcc_no_and}
    Let $U = \WCC(k)$ be the $k\ts{th}$ weakly connected computer in a reachable state $u \in R(\Omega_0)$. Let $S \subseteq U$ be a system within $U$, in a reachable state $s \subseteq u$. If $S$ includes some data register's $\REG{}{AND}$ unit then $\varphi_s(s) = 0$:
    \begin{equation*}
        \not\exists i : \REG{i}{AND} \in S \quad \implies \quad \varphi_s(s) = 0.
    \end{equation*}        
\end{claim}
\begin{proof}
    Assume $\exists i : \REG{i}{AND} \in S$. It follows from Claim \ref{claim:and_implies_ak} that either $\CA{k} \in S$ or $\varphi_s(s) = 0$. But if $\CA{k} \in S$, $S$ is weakly connected and reducible per Claim \ref{claim:weakly_connected_reducible} with $\varphi_s(s) = 0$.  
\end{proof}

\begin{claim}
\label{claim:wcc_only_monads}
    Let $U = \WCC(k)$ be the $k\ts{th}$ weakly connected computer in a reachable state $u \in R(\Omega_0)$. The only candidate complexes that include a unit outside of the program or timekeeping chain are monads in $\REG{}{SIM}$.   
\end{claim}
\begin{proof}
    Let $\mathbb{S}^*$ be the set of complexes that exist within $U$. Let $S \in \mathbb{S}^*$ in a reachable state $s \subseteq u$ be a complex that includes at least one unit outside of the program or timekeeping chain. It follows from Claim \ref{claim:wcc_no_and} that $\not\exists i : \REG{i}{AND} \in S$. From Claim \ref{claim:weakly_connected_reducible}, $S$ must be strongly connected. The only strongly connected $S$ are monads $S = \{\REG{i}{SIM}\}$ for $i = 1, \ldots, n$.  
\end{proof}

\begin{claim}[Complexes of the weakly connected computer]
\label{claim:wcc_complexes}
    Let $U = \WCC(k)$ be the $k\ts{th}$ weakly connected computer with valid initial states $\Omega_0$, and $n=2^k$. For every reachable state $u \in R(\Omega_0)$, the following complexes may form:
    \begin{enumerate}
        \item $\CO$ forms at most one single-unit complex, with $\Phi \leq 2$. 
        \item $\{\CX{i}\}_{i=1}^k$ form at most $k$ separate single-unit complexes, each with $\Phi \leq 2$. 
        \item $\{\CA{i}\}_{i=1}^k$ form at most $k$ separate single-unit complexes, each with $\Phi \leq 2$. 
        \item $\REG{}{SIM}$ form at most $n$ separate single-unit complexes, each with $\Phi \leq 2$.
        \item $\textbf{P}$ form $2^n$ separate $n$-unit complexes, each a ring with $\Phi \leq \frac{3n}{2}$. 
    \end{enumerate}
\end{claim}
\begin{proof}
Items 1-3 follow from Claim \ref{claim:wcc_clock_complexes}. 
Item 4 follows from Claim \ref{claim:wcc_only_monads}. 
Item 5 follows from Claim \ref{claim:wcc_program_complexes}.
The bounds on $\Phi$ follow from Claim \ref{claim:monad_max_big_phi} and Claim \ref{claim:copy_ring_bigphi}.
\end{proof}

This suffices to show that the weakly connected computer at the micro grain---no matter its size or program---can only support cause--effect structures with first-order mechanisms and linearly increasing (or constant) $\Phi$.

\addcontentsline{toc}{subsubsection}{Micro-grain analysis of the strongly connected computer}
\subsubsection*{Micro-grain analysis of the strongly connected computer}
\label{subsubsec:micro_analysis_of_scc}

\begin{claim}
\label{claim:Ak_in_implies_C0_in_and_Ak_off}
    Let $U = \SCC(k)$ be the $k\ts{th}$ strongly connected computer in a reachable state $u \in R(\Omega_0)$. Let $S \subseteq U$ be a system within $U$, in a reachable state $s \subseteq u$. If $|S| > 1$, $\CA{k} \in S$ and $\varphi_s(s) > 0$, then $\CO \in S$ and $\CA{k} = 0$. 
\end{claim}
\begin{proof}
    Suppose $S \subseteq U$ is a candidate system with $\varphi_s(s) > 0$ and $\CA{k} \in S$. It follows from Claim \ref{claim:weakly_connected_reducible} that $S$ is a strongly connected, which implies that $\CO \in S$. 
    
    Assume $\CA{k} = 1$. Since $s \subseteq u,~ u \in R(\Omega_0)$, $\CA{k}=1$ implies that $\CO=0$ and $\CX{1}=0$. 
    \begin{itemize}
        \item Case 1: $\CX{1} \notin S$. Then $\{\CA{i}\}_{i=1}^k$ are externally determined by $\CX{1}$. By Claim \ref{claim:exdet_reducible}, $S$ is reducible. 
        \item Case 2: $\CX{1} \in S$. The future state of $\{\CA{i}\}_{i=1}^k$ are determined by $\CO$. We cut the outputs of $\CX{1}$ without loss,
            \[ \theta = \{{\CX{1}}_\out, (S \setminus \CX{1})_{\inp}\}, \]
        yielding $\varphi_s(s, \theta) = 0$ (i.e., $S$ is reducible). 
    \end{itemize} 
    The reducibility of $S$ contradicts that $\varphi_s(s) > 0$, and thus $\CA{k} = 0$. 
\end{proof}

\begin{claim}
\label{claim:scc_no_and_nor_xor}
    Let $U = \SCC(k)$ be the $k\ts{th}$ strongly connected computer in a reachable state $u \in R(\Omega_0)$. There are no complexes $S \subseteq U$ in a reachable state $s \subseteq u$ that include some data register's $\REG{}{AND}$ unit, as well as the $\REG{}{XOR}$ unit that inputs to it:
    \begin{equation*}
        \exists i : \{\REG{i}{AND}, \REG{i}{XOR}\} \subset S \implies \varphi_s(s) = 0
    \end{equation*}        
\end{claim}
\begin{proof}
    Consider $S \subseteq U$ such that $\exists i : \{\REG{i}{AND}, \REG{i}{XOR}\} \subset S$. From Claim \ref{claim:and_implies_ak}, $\CA{k} \in S$. From Claim \ref{claim:Ak_in_implies_C0_in_and_Ak_off}, $\CO \in S$ and $\CA{k} = 0$. We can cut the outputs of $\REG{i}{XOR}$ without loss,
        \[ \theta = \{{\REG{i}{XOR}}_\out, (S \setminus \REG{i}{XOR})_{\inp}\}, \]
    yielding $\varphi_s(s, \theta) = 0$ (i.e., $S$ is reducible).
\end{proof}

\begin{claim}
\label{claim:scc_no_mux_or_buf}
    Let $U = \SCC(k)$ be the $k\ts{th}$ strongly connected computer in a reachable state $u \in R(\Omega_0)$. Any candidate system $s \subseteq u$ that includes a unit from the multiplexer or buffer is reducible:
    \begin{equation*}
        S \cap (\textbf{M} \cup \textbf{B}) \neq \varnothing \implies \varphi_s(s) = 0. 
    \end{equation*}          
\end{claim}
\begin{proof}
    For $S$ to have $\varphi_s(s) > 0$, it follows from Claim \ref{claim:weakly_connected_reducible} that $S$ must be strongly connected. If $S$ includes a unit from \textbf{M} or \textbf{B}, then $S$ must include some register's $\REG{}{AND}$ unit, as well as the $\REG{}{XOR}$ unit that inputs to it, in order to be strongly connected. That is, $\exists i : \{\REG{i}{AND}, \REG{i}{XOR}\} \subset S$. By Claim \ref{claim:scc_no_and_nor_xor}, $S$ is reducible with $\varphi_s(s) = 0$.
\end{proof}

\begin{definition}[$L_i^{\blacklozenge}$, $L_i^{\blacktriangle}$, and $L_i^{\blacktriangledown}$]
    Let $U = \SCC(k)$ be the $k\ts{th}$ strongly connected computer. Let $L_i = \PRG{}{i} \cup \{\IR{i}\}$ be the $i\ts{th}$ line of the program, plus its corresponding line of the instruction register. We define the three strongly connected subsets of $L_i \subset U$ as follows:
    \begin{align*}
        L_i^{\blacklozenge} &= L_i = \{\PRG{1}{i}, \PRG{2}{i}, \ldots, \PRG{n}{i}, \IR{i}\} \\
        L_i^{\blacktriangle} &= L_i \setminus \{\IR{i}\} = \PRG{}{i} =\{\PRG{1}{i}, \PRG{2}{i}, \ldots, \PRG{n}{i}\} \\
        L_i^{\blacktriangledown} &= L_i \setminus \{\PRG{n}{i}\} = \{\PRG{1}{i}, \PRG{2}{i}, \ldots, \PRG{n-2}{i}, \IR{i}\}
    \end{align*}  
\end{definition}

\begin{claim}
\label{claim:scc_program_candidates}
    Let $U = \SCC(k)$ be the $k\ts{th}$ strongly connected computer in a reachable state $u \in R(\Omega_0)$. Let $S \subseteq U$ be a system within $U$, in a reachable state $s \subseteq u$. If $S$ is irreducible and includes some unit in the program or instruction register, then $S$ is a strongly connected subset of that unit's line. 
    \begin{equation*}
        \bigl( \varphi_s(s) > 0  \quad \wedge \quad S \cap L_i = \varnothing \bigr) \implies S \in \{L_i^{\blacklozenge}, L_i^{\blacktriangle},L_i^{\blacktriangledown}\}
    \end{equation*}
\end{claim}
\begin{proof}
    Since $S$ is irreducible, it follows from Claim \ref{claim:weakly_connected_reducible} that that $S$ is a strongly connected. Claim \ref{claim:scc_no_and_nor_xor} and Claim \ref{claim:scc_no_mux_or_buf} eliminate all strongly connected systems that intersect with $L_i$, except for $\{L_i^{\blacklozenge}, L_i^{\blacktriangle},L_i^{\blacktriangledown}\}$. Therefore, $S$ must be one of these candidates.
\end{proof}

\begin{claim}
\label{claim:scc_state5}
    Let $U = \SCC(k)$ be the $k\ts{th}$ strongly connected computer in a reachable state $u \in R(\Omega_0)$, and $n=2^k$. Let $S \subseteq U$ be a system within $U$, in a reachable state $s \subseteq u$. 
    \\ \\
    If all of the following are true:
    \begin{itemize}
        \item $S$ is irreducible $(\varphi_s(s) > 0)$
        \item $S$ includes any unit outside of the program and instruction register ($S \cap \{\textbf{R},  \textbf{M}, \textbf{T}, \textbf{B}\} \neq \varnothing$)
        \item $S$ is not a monad ($|S| > 1$).
    \end{itemize}
    then the following are implied:
    \begin{itemize}
        \item $u \in R(\Omega_0, 2nj-3)$ for some $j \in \mathbb{N}^+$.
        \item $S = \{\CO, \CA{k}, \REG{i}{AND}\}$ for some $i \in \{1, 2, \ldots, n\}$.
    \end{itemize}
\end{claim}
\begin{proof}
    Assume all of the premises. Claims \ref{claim:weakly_connected_reducible}, \ref{claim:scc_no_and_nor_xor}, \ref{claim:scc_no_mux_or_buf}, and \ref{claim:scc_program_candidates} eliminate all candidates for $S$ except those where $\{\CO, \CA{k}, \REG{i}{AND}\} \subseteq S$ for some $i \in \{1, 2, \ldots, n\}$. The following must also be true:
    
    \begin{enumerate}
        \item $\CA{k}=0$. This follows from Claim \ref{claim:Ak_in_implies_C0_in_and_Ak_off}.
        \item $\CO=0$. If $\CO=1$, then it determines its own future state, and we can cut the outputs of any $\REG{i}{AND} \in S$ without loss, and $S$ is reducible. 
        \item $S \cap \{\CX{i}\}_{i=1}^{k} = \varnothing$. Otherwise we could cut the outputs of $S \cap \{\CX{i}\}_{i=1}^{k}$ without loss, since $\CO$ determines the future state of $\CA{k}$. 
        \item $S \cap \{\CA{i}\}_{i=1}^{k-1} = \varnothing$. Otherwise $S$ would be weakly connected and reducible. 
        \item $\{\CX{i}\}_{i=1}^{k} = 1$. Otherwise, $\CA{k}$ is externally determined, and $S$ is reducible. 
        \item $u \in R(\Omega_0, 2nj-3)$ for some $j \in \mathbb{N}^+$. This follows from the reachability of $s \subseteq u$ and the constraints on the states of units $\CO=0$ and $\CA{k}=0$ (both in $S$), and $\{\CX{i}\}_{i=1}^{k} = 1$ (not in $S$). \label{item:state5}
        \item Finally, $|S \cap \{\REG{i}{AND}\}_{i=1}^{n}| = 1$. Assume otherwise, i.e., $\exists i,j,~ i \neq j ~:~ \{\REG{i}{AND}, \REG{j}{AND}\} \subset S$. 
        
        For any $u \in R(\Omega_0, 2nj-3)$, the current state of $S$ is $s=(0,0,\ldots,0)$ (all OFF). What is the maximal cause state $s'_c$ (see Eqn. 12 in \cite{Albantakis2023})? Given the current state, $\CA{k}=0$ in the cause state, because this is sufficient to bring about $\REG{i}{AND}=0$ and $\REG{j}{AND}=0$ in the current state, regardless of the state of background units $W = U \setminus S$, whereas $\CA{k}=1$ would have no power to bring about the current state $s$. Moreover, the only candidate cause states $\bar{s}$ that have any power to bring about $s$ are those for which at least one of $\CO$, $\REG{i}{AND}, \REG{j}{AND}$ is 1. These cause states are sufficient to bring about $\CO=0$, whereas other cause states have no power to bring about $s$. 
        
        Let $\bar{s}_1$ be a candidate cause state such that $\CO$ is 1 and $\CA{k}$ is 0. Let $\bar{s}_2$ be any candidate cause state such that $\CO$ and $\CA{k}$ are 0 and at least one of $\REG{i}{AND}, \REG{j}{AND}$ is 1. Observe that $\bar{s}_2$ is sufficient to bring about $s$, including $\CA{k}=0$, regardless of the background state $w$. In contrast, $\bar{s}_1$ can only bring about $s$ for certain background conditions (e.g., when $\{\CX{i}\}_{i=1}^{k}=1$, $\bar{s}_1$ does not bring about $s$.) Thus, some degradation of the cause information is guaranteed for $\bar{s}_1$ relative to $\bar{s}_2$. More precisely, because $p_c(\sc \mid \bar{s}_2) > p_c(\sc \mid \bar{s}_1)$ and $p_c^{\leftarrow}(\bar{s}_2 \mid \sc) = p_c^{\leftarrow}(\bar{s}_1 \mid \sc)$, this implies $\ii_{c}(\sc, \bar{s}_2) > \ii_{c}(\sc, \bar{s}_1)$ (see Eqn. 7 in \cite{Albantakis2023}), and therefore $s'_c = \bar{s}_2$. 

        Without loss of generality, assume that $\REG{i}{AND}=1$ in $s'_c = \bar{s}_2$. Given this cause state, we can cut the outputs of $\REG{j}{AND}$ without loss (because $\CO$'s current state was fully determined by $\REG{i}{AND}$'s cause state). The reducibility of $S$ contradicts that $S$ is irreducible, and the claim follows. 
    \end{enumerate}
\end{proof}

\begin{claim}[Complexes of the strongly connected computer]
\label{claim:scc_complexes}
    Let $U = \SCC(k)$ be the $k\ts{th}$ strongly connected computer with valid initial states $\Omega_0$, and $n=2^k$. For every reachable state $u \in R(\Omega_0)$, the following complexes may form:
    \begin{enumerate}
        \item For each line $\{L_i\}_{i=1}^{2^n}$ of the program and instruction register, at most one of its strongly connected subsets $\{L_i^{\blacklozenge}, L_i^{\blacktriangle}, L_i^{\blacktriangledown}\}$ may form a complex, for a total of $2^n$ separate complexes. Each complex is either a ring of size $n$ with $\Phi \leq \frac{3n}{2}$ ibits, or an imperfect ring of size $n+1$ with $\Phi \leq \frac{9(n + 1) + 64}{6}$. \label{item:scc_complexes_program_lines}
        \item $\REG{}{SIM}$ form at most $n$ separate single-unit complexes, each with $\Phi \leq 2$. \label{item:scc_complexes_register_outputs}
        \item $\{\CX{i}\}_{i=1}^k$ form at most $k$ separate single-unit complexes, each with $\Phi \leq 2$. \label{item:scc_complexes_clk_xi}
        \item $\{\CA{i}\}_{i=1}^{k-1}$ form at most $k-1$ separate single-unit complexes, each with $\Phi \leq 2$. \label{item:scc_complexes_clk_ai}
        \item $\{\CO\}$ forms at most one single-unit complex, with $\Phi \leq 2$. \label{item:scc_complexes_clk_c0}
        \item $\{\CA{k}\}$ forms at most one single-unit complex, with $\Phi \leq 2$. \label{item:scc_complexes_clk_ak}
    \end{enumerate}
    For states $u \in R(\Omega_0, 2nj-3)$ for some $j \in \mathbb{N}^+$, the following complex may form:
    \begin{enumerate}
        \setcounter{enumi}{7}
        \item $\{\CO, \CA{k}, \REG{i}{AND}\}$ for some $i \in \{1, 2, \ldots, n\}$, each with $\Phi \leq \frac{9}{2}$. \label{item:scc_complexes_state5}
    \end{enumerate}
\end{claim}
\begin{proof}

Item \ref{item:scc_complexes_program_lines} follows from Claims \ref{claim:scc_program_candidates}, \ref{claim:copy_ring_bigphi}, and \ref{claim:imperfect_copy_ring_bigphi}. 

Per Claim \ref{claim:weakly_connected_reducible}, any complex needs to be strongly connected, and Claims \ref{claim:scc_no_mux_or_buf} and \ref{claim:scc_no_and_nor_xor} rule out most strongly connected subsets of $U$. Item \ref{item:scc_complexes_register_outputs} follows from the fact that the monads $\REG{}{SIM}$ are the only strongly connected subsets of $U$ that do not involve units in the timekeeping chain and have not already been eliminated by Claims $\ref{claim:scc_no_mux_or_buf}$ and \ref{claim:scc_no_and_nor_xor}, or covered by Claim \ref{claim:scc_program_candidates}. 

The only remaining complexes to consider are those that might form from units in the timekeeping chain and the AND units ($\{\REG{i}{AND}\}_{i=1}^{n}$) of each data register. Items \ref{item:scc_complexes_clk_xi}-\ref{item:scc_complexes_clk_ak} follow from Claim \ref{claim:scc_state5}, which details the only complex larger than a monad that might form (item \ref{item:scc_complexes_state5}), and implies that any other complex we have yet to consider must be a monad (items \ref{item:scc_complexes_clk_xi}, \ref{item:scc_complexes_clk_ai}, \ref{item:scc_complexes_clk_c0}, and \ref{item:scc_complexes_clk_ak}).  
\end{proof}

This suffices to show that the strongly connected computer at the micro grain---no matter its size or program---can only support cause--effect structures with first-order mechanisms (and potentially a single second-order mechanism) and linearly increasing (or constant) $\Phi$.


\addcontentsline{toc}{subsubsection}{Macro grain analysis of the weakly and strongly connected computers}
\subsubsection*{Macro grain analysis of the weakly and strongly connected computers}
\label{subsubsec:macro_analysis_of_scc}

In what follows, $U = \SCC(k)$ is always the $k\ts{th}$ strongly connected computer. Note that the results below also hold for the weakly connect computer. Following Marshall et al. \cite{Marshall2024}, $\mathbb{P}(u)$ is the set of all valid systems (both micro and macro) that can be defined from $U$ in state $u$. System $S = \{J_1, \ldots, J_{|S|}\}$ in state $s \in \mathbb{P}(u)$, with micro constituents $U^S$ in (micro) state $u^S \subset u$, is a valid \emph{macro system} if there exists at least one unit $J \in S$, such that $J$'s update grain constitutes more than one micro update (i.e., $\tau_J > 1$), or $J$ contains more than one micro constituent (i.e., $|U^J| > 1$). $S$ has background units $W = U \setminus U^S$ in state $w \subset u$. The background is always treated at the micro grain (every $W_i \in W$ is a micro unit). The background apportionments for the system, $W^S \subseteq W$, are simply the (non-overlapping) apportionments of its constituents: $W^S = \bigcup_{i = 1}^{|S|} W^{J_i}$.

Here, we relax the restrictions of IIT \cite{Marshall2024} in only one respect: we do not assume that $U$ is a system of micro units, and instead consider them to be meso units that implicitly satisfy the postulates. This counters an objection that our micro units are stand-ins for what are actually meso units (e.g., our AND unit could be considered a stand-in for an AND gate built from smaller components/transistors). The practical consequence of this perspective is that the individual units of $U$ can be considered as constituents of a macro unit with $\tau > 1$, even if they do not have a ``self-loop''. This gives the computer the best possible chance of forming complexes whose cause--effect structures replicate those of a simulated substrate, which strengthens the result. Note that all results below would hold (often trivially), if we treated the $U$ as a set of micro units. 

Let $\mathbb{P}^{\mathrm{SIM}}(u)$ be the set of all valid macro systems that include the outputs of all the data registers as constituents (i.e., $\REG{}{SIM} \subseteq U^S, \; \forall S \in \mathbb{P}^{\mathrm{SIM}}$). It follows from Claim \ref{claim:scc_complexes} and the requirement that macro units be maximally irreducible within \cite{Marshall2024} that $\REG{}{SIM}$ are not constituents of other units in $S$, but are themselves units of $S$ (i.e., $\REG{}{SIM} \subseteq S$), possibly at a macro update grain. We aim to show that there is no system $S$ in state $s \in \mathbb{P}^{\mathrm{SIM}}(u)$ that replicates the cause--effect structure $C(t)$ of the size-$n$ simulated target system $T$ in state $t$. The reason we do not show this for any $s \in \mathbb{P}(u)$ is that we are interested in knowing if the macro system replicates $C(t)$ \emph{in virtue of its function as a simulator}. After all, there are many ``pathological'' choices of $t$ that could result in spurious cause--effect structure replication. For example, $t$ could be designed to mimic precisely a subset of the computer that supports a complex, or to fragment into single-unit complexes whose cause--effect structures are trivial to reproduce. In fact, even the restriction that $s \in \mathbb{P}^{\mathrm{SIM}}(u)$ leaves the door open to some spurious replication of simple cause--effect structures, but it will suffice for us to ultimately demonstrate that simulation does not imply cause--effect structure replication. 

We first introduce two conditions related to the potential equivalence of cause--effect structure.
\begin{definition}[$\NONT$]
    Let $d = (m, z^*=\{z^*_c, z^*_e\}, \varphi_d) \in C(s)$ be a distinction in the cause--effect structure of $s$, with mechanism $M$ in state $m$, cause purview $Z^*_c$ in state $z^*_c$, effect purview $Z^*_e$ in state $z^*_e$, and irreducibility $\varphi_d$. 
    $\NONT(s)$ is the following proposition: 
    \begin{equation*} 
        \exists d \in C(s) : z^* \not\subseteq \REG{}{SIM} 
    \end{equation*}
\end{definition}

The $\NONT$ property indicates that a unit outside of $\REG{}{SIM}$ shows up as a purview in the cause--effect structure of $s$, thus precluding cause--effect structure equivalence with the target system. 

\begin{definition}[$\ISO$]
    Let $d = (m, z^*=\{z^*_c, z^*_e\}, \varphi_d) \in C(s)$ be a distinction in the cause--effect structure of $s$, with mechanism $M$ in state $m$, cause purview $Z^*_c$ in state $z^*_c$, effect purview $Z^*_e$ in state $z^*_e$, and irreducibility $\varphi_d$. 
    
    $\ISO^c(i)$ is the following proposition:
    \begin{equation*}
        \REG{i}{SIM} \in z^*_c \implies |z^*_c| = 1, \forall d \in C(s).
    \end{equation*}
    Similarly for $\ISO^e(i)$. Finally, let $\ISO(i) = \ISO^c(i) \vee \ISO^e(i)$.
\end{definition}

The $\ISO(i)$ property indicates that a unit in $\REG{}{SIM}$ is isolated within a cause--effect structure, meaning that it does not appear alongside other target units in a purview, and thus cannot form relations that bind it with other target units. If a unit is isolated, this precludes the possibility that the cause--effect structure of $s$ can be equivalent to the cause--effect structure of any target system that has relations to bind together units (including $PQRS$).

In what follows, we outline a sequence of claims that together demonstrate the desired result. Specifically, we aim to show:

\begin{claim}
\label{claim:macro-result}
For every system $s \in \mathbb{P}^{\mathrm{SIM}}(u)$, at least one of the following must be true:
    \begin{enumerate}
        \item $\varphi_s(s) = 0$
        \item $\NONT(s)$
        \item $\exists i \in \{1, \ldots, n\}$ such that $\ISO(i)$ 
    \end{enumerate}
\end{claim}
\begin{proof}
    The claim follows from Claims \ref{claim:macro1} - \ref{claim:macro7}, proven below. 
\end{proof}

This result suffices to show that the computer cannot replicate the cause--effect structure of an arbitrary target system (including $PQRS$) in virtue of its function as a simulator. 

\begin{claim}
\label{claim:macro1}
    Let $S$ in state $s \in \mathbb{P}^{\mathrm{SIM}}(u)$ be a macro system. If any register's AND gate is neither a constituent nor apportionment of the system, or is an apportionment of that register's SIM gate, then the system is reducible:
    \begin{equation*}
        \REG{i}{AND} \in W \setminus W^S ~ \vee ~ \REG{i}{AND} \in W^{\REG{i}{SIM}} \implies \varphi_s(s) = 0
    \end{equation*}
\end{claim}
\begin{proof}
    By definition, $\REG{}{SIM} \subseteq S, \; \forall S \in \mathbb{P}^{\mathrm{SIM}}(u)$. Assume $\exists i : \REG{i}{AND} \in W \setminus W^S ~ \vee ~ \REG{i}{AND} \in W^{\REG{i}{SIM}}$. Then we can cut the inputs to $\REG{i}{SIM}$ without loss, and $\varphi_s(s) = 0$.
\end{proof}

\begin{claim}
\label{claim:macro2}
    Let $S$ in state $s \in \mathbb{P}^{\mathrm{SIM}}(u)$ be a macro system. If the output of the multiplexer is neither a constituent of $S$ nor part of some multiplexer unit's apportionment, then Claim \ref{claim:macro-result} holds for $s$:
    \begin{equation*}
        \MO \not\in S \cup W^{\textbf{M}} \implies \varphi(s) = 0 ~ \vee ~ \NONT(s) ~ \vee ~ \ISO(i) \; \forall i.
    \end{equation*}
\end{claim}
\begin{proof}
    Assume $\MO \not\in S \cup W^{\textbf{M}}$. It is possible that $\MO \in W^{\REG{i}{SIM}}$ for exactly one $i$, and that $\CO \in W^{\REG{i}{}}$ for exactly one $i$. Thus, there exists at least one $W^{\REG{i}{}}$ such that $(\MO \cup \CO) \cap W^{\REG{i}{}} = \varnothing$. Assume without loss of generality that it is $i = 0$ (neither $\MO$ nor $\CO$ are in $W^{\REG{0}{}}$). 
    
    Consider the possible assignments of $\REG{0}{AND}$ (it could be a constituent of the system, an apportionment of some other unit, etc.). By Claim \ref{claim:macro1}, if $\REG{0}{AND} \in W \setminus W^S$ or $\REG{0}{AND} \in W^{\REG{0}{SIM}}$, then $\varphi_s(s) = 0$. If $\REG{0}{AND} \in W^J$for some $J \in S$ with $U^J \cap \REG{0}{} = \varnothing$ (a mediator for a unit outside $\REG{0}{}$), then:
    \begin{itemize}
        \item Case 1: $\exists k$ such that $\MUX{k} \in S$. Then the outputs of $\MUX{k}$ can be cut without loss (since $\MO \not\in S$ and $\MO \not\in W^{\MUX{k}}$) and $\varphi_s(s) = 0$.        
        \item Case 2: $\not\exists k$ such that $\MUX{k} \in S$. Then we can cut the outputs of $\REG{0}{SIM}$ (if $\REG{0}{XOR} \not\in S$) or the outputs of $\REG{0}{XOR}$ (if $\REG{0}{XOR} \in S$) without loss, and $\varphi_s(s) = 0$. 
    \end{itemize}
    
    All of the possible assignments for $\REG{0}{AND}$ considered so far result in $\varphi_s(s) = 0$. The remaining possibilities are $\REG{0}{AND} \in S$, or $(\REG{0}{XOR} \in S ~ \wedge ~ \REG{0}{AND} \in W^{\REG{0}{XOR}})$. Consider a potential distinction that includes $\REG{0}{SIM}$ in its effect purview. Such a distinction  must include units that output to $\REG{0}{SIM}$ (possibilities are $\REG{0}{SIM}, \REG{0}{AND}, \REG{0}{XOR}$). Now consider whether such a distinction could also include another register output $\REG{i'}{SIM}$ ($i' \neq 0)$ in its effect purview. Units $\REG{0}{SIM}$ and $\REG{i'}{SIM}$ share no common inputs, so the only way for such a distinction with $\REG{0}{SIM}$ and $\REG{i'}{SIM}$ in its effect purview to be irreducible is to include an additional unit $V \not\in \REG{}{SIM}$ in the effect purview that integrates the effect. If such a $V$ is in the effect purview, then $\NONT(s)$, otherwise such a distinction does not appear in the cause--effect structure and $\ISO(0)^{e}$.
\end{proof}

Thus, Claim \ref{claim:macro-result} holds for any system for which $\MO$ is not a unit or a mediator of some $\MUX{k}$. 
The next set of claims considers the potential role of $\REG{i}{AND}$ within a system where $\MO \in S \cup W^{\textbf{M}}$. 

\begin{claim}
\label{claim:macro3}
    Let $S$ in state $s \in \mathbb{P}^{\mathrm{SIM}}(u)$ be a macro system. If the output of the multiplexer is a constituent of $S$ or part of some multiplexer unit's apportionment, and if some register's AND gate is a constituent of the system, then Claim \ref{claim:macro-result} holds for $s$:
    \begin{equation*}
        \MO \in S \cup W^{\MUX{k}} \wedge \REG{i}{AND} \in S \implies \varphi_s(s) = 0 ~\vee~ \NONT(s) ~\vee~ \ISO^e(i).
    \end{equation*}
\end{claim}
\begin{proof}
    Suppose $\MO \in S$ or $\exists k : \MUX{k} \in S \wedge \MO \in W^{\MUX{k}}$. Also suppose $\exists i : \REG{i}{AND} \in S$. Assume without loss of generality that $\REG{1}{AND} \in S$. 
    
    The only inputs to $\REG{1}{SIM} \in S$ are $\REG{1}{SIM}$ and $\REG{1}{AND}$. Therefore any distinction with $\REG{1}{SIM}$ in its effect purview must have $\REG{1}{SIM}$ and/or $\REG{1}{AND}$ in its mechanism. Consider whether some $\REG{i'}{SIM}$ could also be in the effect purview ($i' \neq i$). Then the mechanism must include some unit $J$ that outputs to $\REG{i'}{SIM}$ (otherwise it is reducible).  
    \begin{itemize}
        \item Case 1: $J = \REG{1}{AND}$. Then $\CA{k} \in W^{\REG{1}{AND}}$, in order to create the necessary output pathway to $\REG{i'}{SIM}$. Also, $\REG{1}{XOR} \in W^{\REG{1}{SIM}}$, providing an output pathway from $\REG{1}{SIM}$ to $\REG{1}{AND}$, otherwise the distinction's cause would be reducible. However, this means that we can cut the inputs to $\REG{i}{}$ without loss when partitioning $s$ to test for system integration, so $\varphi_s(s) = 0$. 
        \item Case 2: $J \neq \REG{1}{AND}$. Then $J \not\in \REG{1}{}$, and $J$ does not share any outputs with $\REG{1}{SIM}$ or $\REG{1}{AND}$. Therefore the distinction is reducible (i.e, $\ISO^e(1)$), or includes some unit outside $\REG{}{SIM}$ in its effect purview (i.e., $\NONT(s)$).  
    \end{itemize}
\end{proof}

Based on Claim \ref{claim:macro2}, Claim \ref{claim:macro-result} holds for systems where $\MO$ is neither a unit nor a mediator of $\MUX{k}$. Based on Claims \ref{claim:macro1} and \ref{claim:macro3}, Claim \ref{claim:macro-result} also holds in systems where $\MO$ is a unit itself, or mediator of some $\MUX{k}$, and any $\REG{i}{AND}$ are units, background, or mediators of $\REG{i}{SIM}$. It remains to be shown for systems where $\MO$ is a unit itself, or mediator of some $\MUX{k}$, and each $\REG{i}{AND}$ is a mediator for some $J \neq \REG{i}{SIM}$. We now focus on the potential distinctions that such a system could support, and whether or not they entail $\NONT(s)$ or $\ISO(i)$ for some $i$. 

\begin{claim}
\label{claim:macro4}
    Let $S$ in state $s \in \mathbb{P}^{\mathrm{SIM}}(u)$ be a macro system.
    Let $d = (m, z^*=\{z^*_c, z^*_e\}, \varphi_d) \in C(s)$ be a distinction in the cause--effect structure of $s$, with mechanism $M$ in state $m$, cause purview $Z^*_c$ in state $z^*_c$, effect purview $Z^*_e$ in state $z^*_e$, and irreducibility $\varphi_d$.
    Let $\textbf{X} = \textbf{T} \cup \textbf{B}$ be the set of all meso units in the buffer and timekeeping chain. 

    If (1) $\MO$ is a unit itself, or a mediator of some multiplexer unit, and (2) each $\REG{i}{AND}$ is a mediator for some $J \neq \REG{i}{SIM}$, and (3) any mechanism $m \in d \in C(s)$ includes a member of $\textbf{X}$, then (4) Claim \ref{claim:macro-result} holds for $s$:
    \begin{equation*}
        \bigl( \MO \in S \cup W^{\textbf{M}} \bigr)
        ~ \wedge ~
        \bigl( \exists J \neq \REG{i}{SIM} : \REG{i}{AND} \in W^J, ~ \forall i \bigr)
        ~ \wedge ~
        \bigl( U^m \cap \textbf{X} \neq \varnothing \bigr) 
        \implies \NONT(s) 
    \end{equation*}
\end{claim}
\begin{proof}
    Assume all the premises, and consider any $i \in \{1, \ldots, n\}$. 
    Since $\MO \in S$ or $\exists k : \MUX{k} \in S \wedge \MO \in W^{\MUX{k}}$, and $\REG{i}{AND} \not\in W^{\REG{i}{SIM}}, ~ \forall i$, there is no path consisting only of units in $W^{\REG{i}{SIM}}$ from $\REG{i}{SIM}$ to any unit in $\textbf{X}$, including $U^m \cap \textbf{X}$. Without such a path, it is possible to cut $U^m \cap \textbf{X}$ away from $Z^*_c$ without loss ($\varphi_d = 0$) unless there is another unit $J \not\in \REG{}{SIM}$ that integrates the cause (i.e., $\NONT(s)$). 
\end{proof}
    
\begin{claim}
\label{claim:macro5}
    Let $S$ in state $s \in \mathbb{P}^{\mathrm{SIM}}(u)$ be a macro system. 
    Let $d = (m, z^*=\{z^*_c, z^*_e\}, \varphi_d) \in C(s)$ be a distinction in the cause--effect structure of $s$, with mechanism $M$ in state $m$, cause purview $Z^*_c$ in state $z^*_c$, effect purview $Z^*_e$ in state $z^*_e$, and irreducibility $\varphi_d$.
    Let $\textbf{X} = \textbf{P}$ be the set of all meso units in the program. 

    If (1) $\MO$ is a unit itself, or a mediator of some multiplexer unit, and (2) each $\REG{i}{AND}$ is a mediator for some $J \neq \REG{i}{SIM}$, and (3) any mechanism $m \in d \in C(s)$ includes a member of $\textbf{X}$, then (4) Claim \ref{claim:macro-result} holds for $s$:
    \begin{equation*}
        \bigl( \MO \in S \cup W^{\textbf{M}} \bigr)
        ~ \wedge ~
        \bigl( \exists J \neq \REG{i}{SIM} : \REG{i}{AND} \in W^J, ~ \forall i \bigr)
        ~ \wedge ~
        \bigl( U^m \cap \textbf{X} \neq \varnothing \bigr)
        \implies \NONT(s) 
    \end{equation*}
\end{claim}
\begin{proof}
    Assume all the premises, and consider any $i \in \{1, \ldots, n\}$. 
    Since $\MO \in S$ or $\exists k : \MUX{k} \in S \wedge \MO \in W^{\MUX{k}}$, there is no path consisting only of units in $W^{\textbf{X}}$ from $\textbf{X}$ to any $\REG{i}{SIM}$. Without such a path, it is possible to cut $U^m \cap \textbf{X}$ away from $Z^*_e$ without loss ($\varphi_d = 0$) unless there is another unit $J \not\in \REG{}{SIM}$ that integrates the effect (i.e., $\NONT(s)$).   
\end{proof}

Based on Claim \ref{claim:macro2}, Claim \ref{claim:macro-result} holds for systems where $\MO$ is neither a unit itself, nor a mediator of some $\MUX{k}$. Based on Claims \ref{claim:macro1} and \ref{claim:macro3}, Claim \ref{claim:macro-result} also holds in systems where $\MO$ is a unit or mediator of some $\MUX{k}$, and any $\REG{i}{AND}$ are units, background, or mediators of $\REG{i}{SIM}$. Moreover, based on Claims \ref{claim:macro4}-\ref{claim:macro5}, for systems where $\MO$ is a unit or mediator of some $\MUX{k}$, and each $\REG{i}{AND}$ is a mediator for some $J \neq \REG{i}{SIM}$, no unit in $\mathbf{B} \cup \mathbf{T} \cup \mathbf{P}$ can contribute to a mechanism. We continue to focus on the potential distinctions that such a system could support, and whether or not they entail $\NONT(s)$ or $\ISO(i)$ for some $i$.

\begin{claim}
\label{claim:macro6}
        Let $S$ in state $s \in \mathbb{P}^{\mathrm{SIM}}(u)$ be a macro system. If any multiplexer unit connects to every register's SIM gate, then Claim \ref{claim:macro-result} holds for $s$:
    \begin{equation*}
        \exists J \in \textbf{M} \cap S : J \out \REG{i}{SIM}, ~ \forall i \implies \NONT(s) ~ \vee ~ \varphi_s(s) = 0.  
    \end{equation*}
\end{claim}
\begin{proof}
    Assume $\MO \out \REG{i}{SIM}, \; \forall i$ (a similar argument holds for the case where $\exists k : \MO \in W^{\MUX{k}})$. 
    It follows from the irreducibility of $s$ that (1) $\exists d = (m, z^*, \ldots) \in C(s) : m = \REG{}{SIM}$, and (2) $z^*_e = \MO$, and thus (3) $\NONT(s)$. If this distinction did not exist (i.e., $\varphi_d(d) = 0$), it would mean that the mechanism could be partitioned along some $\theta \in \Theta(M,Z)$ (i.e., $\theta \in \Theta(\REG{}{SIM}, \{\MO\})$) without loss. 
    However, this cut would also be available at the system level (i.e., $\theta \in \Theta(S)$, though this is not necessarily true in general), and thus $\varphi_s(s) = 0$. Said another way, a cut that makes $d$ reducible cuts one or more connections from $\REG{}{SIM}$ without loss. The particular connectivity structure of $s$ is such that the same cut is in $\Theta(S)$, meaning that the same connections that can be cut without loss are available assessing system integrated information, and thus $\varphi_s(s) = 0$.  
\end{proof}

\begin{claim}
\label{claim:macro7}
    Let $S$ in state $s \in \mathbb{P}^{\mathrm{SIM}}(u)$ be a macro system. If the multiplexer's output is either a constituent of $S$ or is part of some multiplexer unit's apportionment, then Claim \ref{claim:macro-result} holds for $s$:
    \begin{equation*}
        \MO \in S \cup W^{\textbf{M}} \implies \varphi_s(s) > 0 ~ \vee ~ \NONT(s) ~ \vee ~ \exists i : \ISO(i)
    \end{equation*}
\end{claim}
\begin{proof} Assume the premises. 

    It follows from Claims \ref{claim:macro1}-\ref{claim:macro3} that either Claim \ref{claim:macro-result} holds, or $\exists J \neq \REG{i}{SIM} : \REG{i}{AND} \in W^J, ~ \forall i$.
    
    It follows from Claims \ref{claim:macro4} and \ref{claim:macro5} that no units in $\mathbf{T} \cup \mathbf{P} \cup \mathbf{B}$ contribute to any mechanism in $C(s)$.
    
    It follows from Claim \ref{claim:macro6} that there exists a unit $\REG{i}{SIM} \in S : \neg\wp(\MO, \REG{i}{SIM}) \vee \neg\wp(\MUX{k}, \REG{i}{SIM})$. Assume, without loss of generality, that this unit is $\REG{1}{SIM}$. 
    
    The only possible mechanism units that output to $\REG{1}{SIM}$ are itself and $\REG{1}{XOR}$:
    \begin{equation*}
        \REG{i}{SIM} \in z^*_e \implies \{\REG{i}{SIM}, \REG{i}{XOR}\} \subseteq m.
    \end{equation*}
    These units do not output to any $\REG{i'}{SIM}, \; i' \neq 0$. For $\REG{i'}{SIM}$ to be in the effect purview ($\REG{i'}{SIM} \in Z_e^*$), there must be some additional $J \in m$ such that $\wp(J, \REG{i'}{SIM})$. However, $\wp(J, \REG{i'}{SIM}) \implies \neg\wp(J, \REG{1}{SIM})$. So either there is an additional unit $V \in Z_E^*$ that integrates the distinction (i.e., $\NONT(s)$) or the effect is reducible by $\theta = \{(\{\REG{1}{SIM}, \REG{1}{XOR}\}, \{\REG{1}{SIM}\}), (\{J\}, \{\REG{i'}{SIM}\})\}$ and $\ISO(1)$.
\end{proof}

\newpage
\section*{Applying IIT's postulates to candidate macro units within the computer}
\label{sec:applying_postulates_to_macro_computer}

For a more complete and general discussion of IIT's postulates as they apply to macro systems, see \cite{Marshall2024}. Here, we present a simplified summary of some key concepts, as they apply to the computer in Fig. \ref{fig:macro_computer}. 

A complex and its intrinsic units must comply with IIT's postulates of physical existence \cite{Albantakis2023}. Thus, like complexes, macro units can only qualify as units if their cause--effect power is irreducible to that of independent subsets (satisfying integration). Otherwise, one could literally build something (a macro unit) out of nothing (non-interacting micro units; Fig. \ref{fig:micro_macro}C). And, just as complexes must be definite, which translates into maximally irreducible cause--effect power (satisfying exclusion), so must macro units. Otherwise, one could build something (a macro unit) out of nearly nothing (weakly interacting sets of micro units; Fig. \ref{fig:micro_macro}D). 

Exclusion has the further consequence that complexes cannot overlap in their constituent units, implying that their cause--effect power should not be counted multiple times. For the same reason, intrinsic units cannot overlap, and their cause--effect power should not be counted multiple times (Fig. \ref{fig:intrinsic_unit_constraints}A). Thus, the multiplexer micro units cannot simultaneously belong to $\sigma$, $\varsigma$, or $\upsilon$ (though an alternative mapping could reassign them, say from $\omega$ to $\rho$). In order not to count intrinsic cause--effect power multiple times, a further constraint for intrinsic units is that they cannot serve as mediators for other intrinsic units within the same complex. For example, one might include the multiplexer units into a macro unit (here, $\omega$) separate from $\rho$, $\sigma$, $\varsigma$, $\upsilon$. From an extrinsic perspective, one might observe that a change in the state of $\rho$ can affect the states of $\rho$, $\sigma$, $\varsigma$, and $\upsilon$ through a pathway that includes the multiplexer. However, this causal power already ``belongs'' to the multiplexer's macro unit $\omega$ (Fig. \ref{fig:intrinsic_unit_constraints}B). (This is somewhat akin to ``screening-off'' in other causal inference frameworks) \cite{Pearl2000, Salmon1971, Brandon1982, Reichenbach1999}. By a similar logic, if $\omega$ were not treated as a macro unit and the multiplexer micro units were reassigned to $\rho$ under an alternative mapping (not shown in Fig. \ref{fig:macro_computer}), then the other macro units $\sigma$, $\varsigma$, and $\upsilon$ would not have direct causal power among themselves, because that would require the contribution of micro units that now belong to $\rho$. 

Similar logic dictates that although micro units constituting the background conditions of a complex can mediate interactions among its intrinsic units, their mediation must be partitioned into causally independent subsets, each associated with a different intrinsic unit. Otherwise, overlaps in the mediating background conditions would allow for counting intrinsic cause--effect power multiple times. For example, consider the subsystem constituted of the four units $\rho$, $\sigma$, $\varsigma$, and $\upsilon$---the read-out units of the computer. In this case, the multiplexer units that belonged to $\omega$ in Fig. \ref{fig:macro_computer} do not belong to any unit within the system, but instead are left out of the system entirely, and treated as background units. From the extrinsic perspective of an observer, a change in the state of $\rho$, $\sigma$, $\varsigma$, and $\upsilon$ can still affect each other's future state through a pathway that includes the multiplexer. However, this extrinsic integration assumes that multiplexer units in the background mediate multiple interactions, doing multiple duty in terms of cause--effect power. This causal overlap can be discounted by apportioning the background conditions among the intrinsic units \cite{Marshall2024}.

\section*{Why the computer is unlikely to support a complex with high structure integrated information ($\Phi$)}
\label{subsubsec:macro_likely_low_phi}

In \hyperref[sec:analysis_of_computer]{``Analysis of the computer''} we analyzed all possible micro-grain systems, in all states that might be visited during simulation. The resulting cause--effect structures were stereotyped, lacking higher-order distinctions or complex relations, and their $\Phi$ grew either linearly with the number of units being simulated, or was a fixed constant. In \hyperref[subsubsec:macro_analysis_of_scc]{``Macro grain analysis of the weakly and strongly connected computers''}, the cause--effect structures of macro systems were characterized for all possible grains. The resulting cause--effect structures could not replicate the cause--effect structure of an arbitrary simulated system. This was sufficient to establish that functional equivalence between two systems does not imply phenomenal equivalence, contradicting a central tenet of computational-functionalism. 

An outstanding question is, can the computer still support a macro complex with high $\Phi$ (i.e., can the computer be ``highly conscious,'' with rich content of \emph{any} sort)? Perhaps there is ``something it is like'' to be a subset of the computer, albeit quite different from the simulated system, and not in virtue of the simulation \textit{per se}. While we do not yet have a formal proof against this possibility under IIT 4.0, such a proof was possible under IIT 3.0 \cite{Findlay2019}, and the advancements incorporated into the latest version of the theory only offer further reason to doubt that the computer can support a highly integrated macro complex, though they make obtaining a formal proof more difficult (for several reasons, one of which is described below). However, even if we completely set aside these intuitions and the previous result from IIT 3.0, there are several other reasons to conjecture that the computer cannot support a macro complex with high $\Phi$, outlined in the following. 

It is possible that macro grain systems can have more integrated information than their corresponding micro-grain systems \cite{Hoel2016, Marshall2018, Marshall2024}. Generally, there are two ways that a macro grain system can accomplish this: (1) increasing its intrinsic information by enhancing determinism (or reducing degeneracy) and thus improving effect (or cause) selectivity; or (2) enhancing integration by increasing interconnectivity within the system at the macro grain. The first scenario arises when groups of homogeneous micro units are averaged together to form a ``coarse-grain'’ macro unit, while the latter scenario arises when specialized and heterogeneous units (and the fault lines between them) are concealed within a ``blackbox’' macro unit. 

The model computer (whether weakly or strongly connected) at the micro grain does not have the right unit functions or connectivity for macro groupings to increase integrated information. For a macro unit to improve effect selectivity, it should aggregate functionally homogeneous micro units. However, the micro-grain computer is already deterministic, so further improvements in selectivity are not achievable through coarse graining. Similarly, macro units improve integration when they combine sparsely interconnected micro units into larger, more interconnected macro systems. Due to the modular architecture of our system, especially bottlenecks in the time-keeping chain and multiplexer-based CPU, macro grain systems will struggle to achieve the necessary interconnectivity. This limitation is reflected in the macro grain results, which show that register units remain isolated in either the cause or effect purviews, regardless of how micro units are grouped into macro units. Isolated units have low fan-in and fan-out, representing fault lines within the system, and leading to low $\varphi_s$.

As mentioned above, all possible micro-grain systems were exhaustively evaluated to identify potential complexes. Such an analysis was possible because, at the micro grain, the computer is sparsely connected and contains externally determined units, which both decreases $\varphi_s$ and simplifies its computation by eliminating the need to evaluate all possible partitions to identify the minimum partition (MIP) \cite{Albantakis2023}. A similar analysis is not feasible at macro grains, due to the way mappings between micro and macro units, and background apportionments, can inject uncertainty into the TPM. This uncertainty makes analytically identifying the MIP difficult, even if $\varphi_s$ is vanishingly small. Thus, it is difficult to rule out potential macro systems as having $\varphi_s = 0$ (and, by extension, $\Phi = 0$). However, this same uncertainty that makes finding an analytical solution difficult generally decreases cause--effect power at the macro grain. Specifically, uncertainty in the macro TPM reduces the selectivity of the macro system, and thus both its intrinsic information and integrated information \cite{Marshall2023}. 

\section*{How the computer compares to modern computer architectures.}
\label{sec:modern_architectures}

An important question is to what extent the results for our simple computer can be extended to other computers. Of course, there is a tremendous diversity of computing machines, which include cell phones, tablets, mainframes, supercomputers, microcontrollers, video game consoles, enterprise servers, and more. All serve different purposes and face different design constraints. It should be no surprise, then, that there is no such thing as a canonical computer architecture.

Nevertheless, most modern computers have some important things in common. For example, common high-level components include: a clock, a data memory, an instruction memory (often the two memories are shared), and a processor containing both an arithmetic logic unit (ALU) and an instruction register. The clock generates a periodic signal that is used to coordinate and synchronize different parts of the computer. The instruction memory stores code which determines the specific actions---chosen from a larger repertoire of possible actions---that a computer will take. The processor repeatedly fetches one of these instructions from memory, determines (``decodes'') the corresponding action, and then executes that action. Executing the action often entails using the ALU to perform some simple operation (e.g., addition, subtraction) on one or more pieces of data fetched from the data memory. In some cases, the result of this operation can influence which instruction is fetched from memory next. This process repeats indefinitely in what is called the ``fetch-decode-execute'' loop. All of the sophisticated things that a computer does are achieved by cleverly exploiting this loop. 

Many implementations of the fetch-decode-execute loop result in a computer that operates sequentially (the order of instructions matters), suffers some sort of computational bottleneck (e.g., the processor can only execute instructions as fast as the ALU can operate), and is organized in a highly modular fashion. This is perhaps partially due to the nature of the computations being performed, but is also for practical reasons. Computers are designed and built by engineers, and sequential, modular systems are easier to analyze than parallel, integrated systems; by contrast, as neuroscientists are well aware, it is difficult to reason about vast, stochastic systems that perform complex functions in parallel. The brain did not evolve to be understood.

So how much is our computer like a ``real'' computer, given that the latter is somewhat ill-defined? Ours is certainly a toy system, small and unsuitable for practical use. It is missing some common features of modern computers, such as an addressable memory. The instruction it fetches during each fetch-decode-execute loop is never a function of a previous ``execute'' step. Some of the authors have designed conventional computers made from transistors, and we would be the first to admit that what is presented here is only analogous to those machines. 

It is important to point out that our computer is presumed to exist at the ``micro grain,'' where units are assumed to be indivisible and update their states synchronously at discrete intervals. Real computers are designed and fabricated at a level far above the smallest physical units, are subject to thermal noise, and receive inputs that are temporally jittered and passed on to outputs after variable propagation delays. In most systems, this desynchronization of units due to internal variability is one of the primary reasons for having a clock. We, on the other hand, only need a clock to coordinate steps of the computation. Our target system, PQRS, does not need a clock at all. 

There are certain aspects of our toy computer that are not fundamental aspects of all computers, but only apply to a subset of computers. As examples: the program in our computer is not malleable at runtime, as with ROMs (c.f. RAMs) and FPGAs; and the data and program in our computer are separate modules, as with Harvard architectures (c.f. Princeton architectures). 

That said, our computer captures many of a computer's essential aspects. It has a clock, a program memory, a data memory, and a processing unit. It is a clocked sequential circuit with a convergent datapath and a modular architecture. It uses design motifs found in modern architectures, such as T-latches and a multiplexer. The processing unit even resembles an ALU whose opcode (a signal that indicates which operation to perform) doubles as one of its operands. As far as its functional properties are concerned, we demonstrate that our computer can easily be made generally programmable and Turing-complete without affecting our results. 

Ultimately, the reasons that our computer cannot support complexes with high $\Phi$ or rich cause--effect structures---much less replicate the cause--effect structure of a richly conscious simulandum---are sufficiently general and high-level as to be mostly abstracted from the ways in which our computer differs from other common architectures. Hence, our results are likely to generalize to most current computer architectures, insofar as they rely on a fetch-decode-execute loop, a central processing unit, a high degree of datapath convergence, and modules which are themselves largely unintegrated. 

It is an open question whether certain parallel computing architectures that are very different from more traditional sequential (e.g., Harvard and Princeton) architectures, such as systolic array variants, may have substrates more favorable for supporting complexes of high $\Phi$. It is important to note that many classes of parallel architecture may have relatively little interdependence between the parallel streams of computation, resembling the way in which the human cerebellum is thought to implement segregated parallel streams of computation that appear to be entirely unconscious. On the other hand, a parallel architecture that more closely resembled a grid-like lattice of internally integrated compute nodes, resembling the way in which the human visual cortex is thought to be organized, might stand a good chance of supporting one or more high-$\Phi$ complexes. 

An interesting direction for future research would be to apply the tools of IIT to analyze computing machines with radically different architectures than those described above. For example, an intriguing question is whether truly neuromorphic computers that mimic the physical organization of our brain and do not rely on a CPU could be designed to achieve both functional \textit{and} phenomenal equivalence to human brains. It is also unclear whether quantum computers, with entangled qubits, might be capable of supporting complexes with high integrated information and rich cause--effect structures \cite{Albantakis2023-Quantum}.  

Finally, we note that our most fundamental and important result, the dissociation in principle between sophisticated functions and consciousness, holds regardless of whether or not a high-$\Phi$ architecture ever becomes used in digital computing machines: what we have demonstrated is that such a complex would \emph{not} have high $\Phi$ in virtue of the function being performed, and that the content of any experience---what the unfolded substrate is---may be dissociated from what the substrate does. 

\section*{Replication vs. Simulation.}
\label{sec:replication_vs_simulation}

To better understand our results, it may be useful to consider a thought experiment in which various parts of a brain are replaced with functionally-equivalent, silicon substitutes. This experiment is much like the gradual neural replacement scenario often used to discuss organization invariance and the question of ``fading qualia'' \cite{Chalmers1995}. Consider the following functionally equivalent brains:
\begin{enumerate}
    \item A normal, organic, biological brain in a particular state, with a particular set of neurons and a particular set of directed connections between them. For the sake of the thought experiment, we assume that this system can be described by a causal model as in \cite{Albantakis2023}. In principle, units of the causal model can be complete neurons, dendritic compartments, individual synapses, etc. For simplicity, let us say that each unit is a neuron, and that it is interactions at the level of neurons that are directly responsible for the quantity and quality of consciousness. The existence of a directed ``connection'' between neurons simply implies that the state of one neuron can directly affect the state of the other, without needing a third neuron to mediate this effect.  
    \item A brain just like (1), but where each neuron has been replaced by a silicon chip and wiring, such that there is a 1:1 mapping between the biological neurons in (1) and silicon chips in (2), and such that there is a connection between two chips in (2) if-and-only-if there is a connection between the two corresponding neurons in (1). Each silicon chip is perfectly functionally equivalent to the neuron that it replaces. There are two subcases to consider: 
    \begin{enumerate}
        \item The silicon chips use a (possibly neuromorphic) physical architecture such that each chip is integrated (to be precise, ``maximally irreducible within'' \cite{Marshall2024}).
        \item The silicon chips are unintegrated. 
    \end{enumerate}
    \item A brain like in (2), but where each silicon neuron is physically separated from its neighbors, and instead communicates with other neurons via transceivers. Each input to a given neuron is replaced by a single transceiver dedicated to relaying that input, and the neuron's output is also replaced by a single dedicated transceiver. Each output transceiver is hardcoded to wirelessly pass its signal along to the appropriate input transceivers. For the sake of argument, assume that each transceiver is a perfect drop-in replacement for whatever physical connection it replaced, and operates entirely independently of all other transceivers (i.e., their signals are not multiplexed in the same frequency bands, etc.). The resulting system is perfectly functionally equivalent to (1) and (2). 
    \begin{enumerate}
        \item The silicon chips use a (possibly neuromorphic) physical architecture such that each chip is integrated (to be precise, ``maximally irreducible within'' \cite{Marshall2024}).
        \item The silicon chips are unintegrated. 
    \end{enumerate}
    \item A brain like in (3), but where transceivers no longer communicate directly with each other. Instead, each output transceiver sends its signal to a central router, and each input transceiver receives its signal from this router. The implementation of the routing circuitry matters, of course, but for our purposes we should only need to assume that it is non-trivial; That is, outputs are not merely connected directly to inputs, but rather there is some kind of computation (checking a lookup table, say) that needs to be done by the router in order to determine which outputs need to be routed to which inputs. There is only one central router servicing all transceivers in the brain (i.e., simulating all of that brain's connections), but it runs fast and its I/O is coordinated so that the resulting system is perfectly functionally equivalent to (1), (2), and (3). 
    \item A brain like in (2), but where we now replace silicon neurons, rather than the connections between them, with transceivers. Each silicon neuron is replaced by a transceiver, such that there is a 1:1 mapping between the silicon neurons in (2) and transceivers in (5), and such that there is connection between two transceivers in (5) if-and-only-if there is a connection between the two corresponding silicon neurons in (2). Each transceiver is a dumb relay, forwarding the inputs that it receives to a central processing unit (CPU) that computes the next state of the replaced silicon neuron. The CPU sends this state back to the same transceiver, which relays this output to any connected transceivers at exactly the moment when the replaced neuron would have. There is only one CPU servicing all transceivers in the brain (i.e., simulating all of that brain's neurons), but it runs fast and its I/O is coordinated so that the resulting system is perfectly functionally equivalent to (1), (2), (3), (4), and (5). There are many schemes according to which this central processing unit might schedule its workload. For instance, the CPU might compute the next state of each transceiver and the neuron that it represents one-at-a-time, sequentially, and always in the same order. This would provide a deterministic way for the computer to ``know'' exactly, at any given moment, which neuron it needs to be simulating the function of. Or it may be that each transceiver stamps its transmissions to the CPU with a barcode that identifies what neuron the CPU should simulate when processing that data. The exact scheme does not matter.
    \item A brain in which both the local function of each neuron and the connections between neurons are simulated entirely on a single CPU, as was done in this paper. The system is functionally equivalent to (1), (2), (3), (4), and (5). 
\end{enumerate}
Our claim, based on the results obtained in this paper, is that systems (1), (2a), and (3a) are similarly conscious, while systems (2b), (3b), (4), (5), and (6) are unconscious as a whole, possibly supporting several small complexes with low $\Phi$ and simple cause--effect structures dissociated from the larger system's function. 

\clearpage
{\centering\Huge\sffamily\bfseries Supplementary Figures\par}
\vspace{0.5cm}
\hrule


\begin{figure}
\centering
\includegraphics[width=7in]{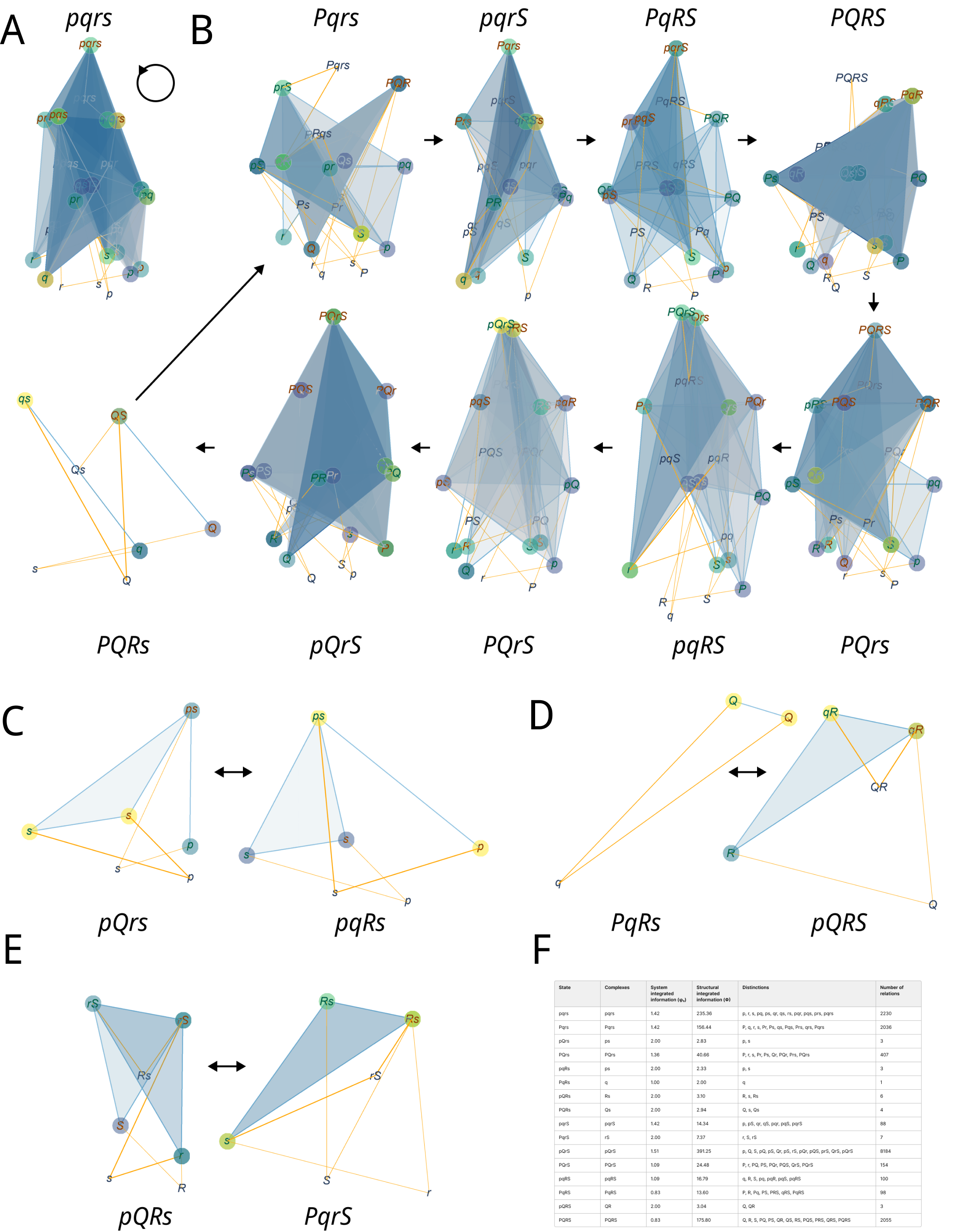}
    \caption[\textbf{cause--effect structure of PQRS at every state.}]{\textbf{cause--effect structure of PQRS at every state.} Dynamical evolution of the unfolded cause--effect structure of the target system PQRS as it transitions through the 16 possible states in its state space. Arrows between cause--effect structures indicate the trajectory of the system from one state to another. Depending on its initial state, PQRS may exhibit any of five cycles. 
    \textbf{(A)} In the first cycle, PQRS is trapped in the fixed-point state $pqrs$, with a single maximally irreducible complex of 4 units. 
    \textbf{(B)} In the second cycle, PQRS will transit through 9 states: $Pqrs$, $pqrS$, $PqRS$, $PQRS$, $PQrs$, $pqRS$, $PQrS$, $pQrS$, and $PQRs$. In the first eight states, PQRS will be a single, maximally-irreducible complex of four units. In the final state ($PQRs$), only $Qs$ will form a complex. 
    \textbf{(C-E)} The final three cycles are simple back-and-forth transitions between states where there is never more than a single, two-unit complex.  
    \textbf{(F)} Summary of all the complexes and associated cause--effect structures that appear in every possible state of PQRS.}
\label{fig:pqrs_cycle}
\end{figure}

\begin{figure}
    \centering
    \includegraphics[width=4.51in]{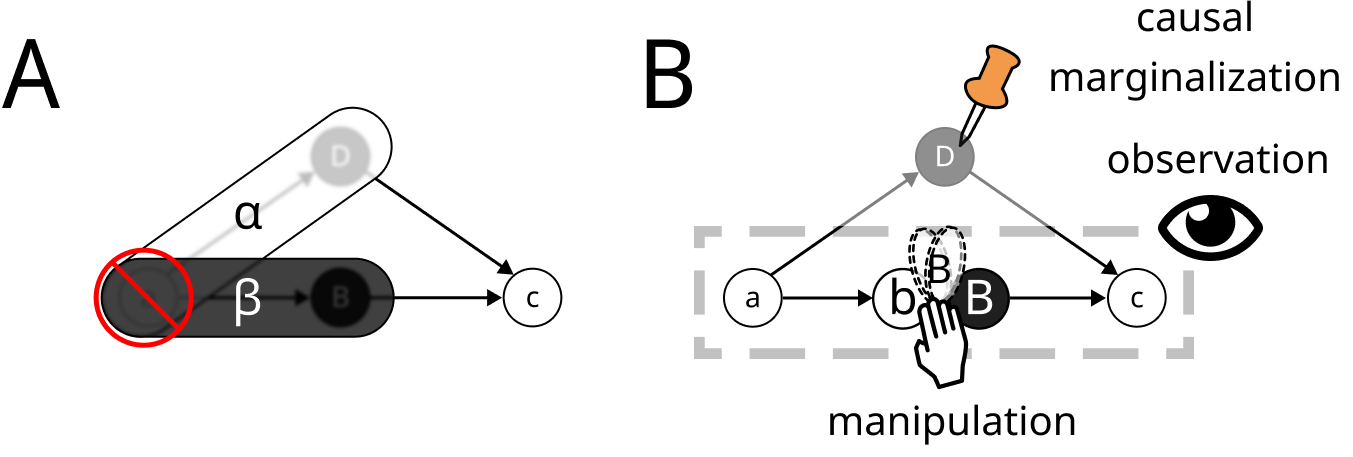}
    \caption{\textbf{Constraints on intrinsic units.}
    A complex and its intrinsic units must comply with IIT's postulates of physical existence \cite{Albantakis2023}.
    (\textbf{A}) Units (here, $\alpha$ and $\beta$) cannot overlap in their micro constituents, because one should not count cause--effect power multiple times. This also means that it must be possible to manipulate each macro unit (change its state) independently and observe the result. This prohibits $\rho$, $\sigma$, $\varsigma$, and $\upsilon$ from sharing any of the units that constitute $\omega$ in Fig. \ref{fig:macro_computer}.  
    (\textbf{B}) A complex's units cannot serve as mediators for other units. From an extrinsic perspective, one might observe that a change in the state of $a$ can affect the state of $c$ through a pathway that includes $b$ and is entirely intrinsic to the system. However, from the intrinsic perspective, this causal power ``belongs'' to $b$ and not $a$. This is somewhat akin to ``screening-off'' in other causal inference frameworks \cite{Pearl2000, Salmon1971, Brandon1982, Reichenbach1999}. Similarly, the causal power of $\omega$ must be accounted for separately from the causal power of $\rho$, $\sigma$, $\varsigma$, and $\upsilon$ in Fig. \ref{fig:macro_computer}.} 
    \label{fig:intrinsic_unit_constraints}
\end{figure}

\begin{figure}
\centering
\includegraphics[width=7.5in]{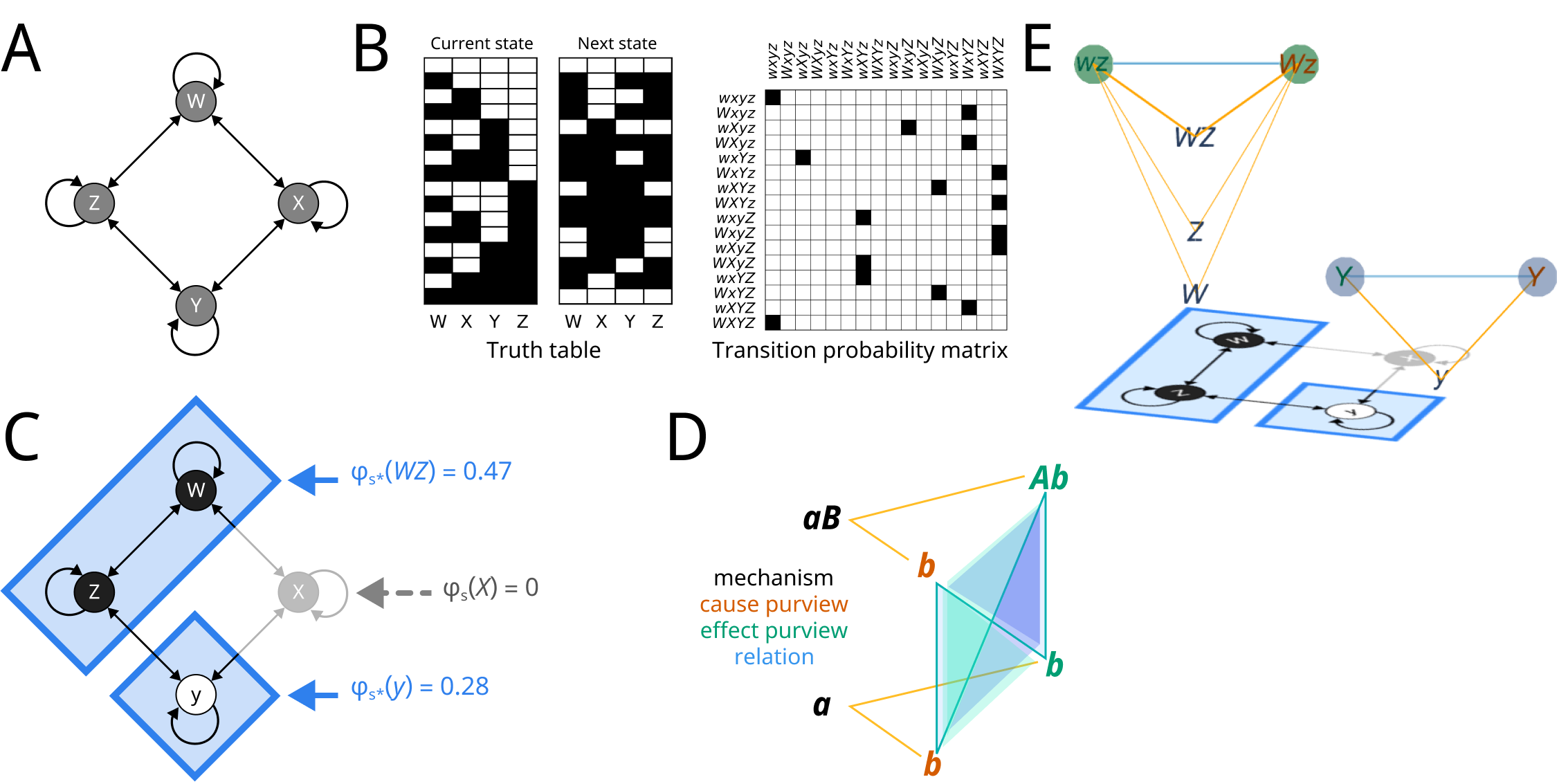}
\caption[\textbf{A four-cell elementary cellular automaton implementing Rule 110.}]{\textbf{A four-cell elementary cellular automaton implementing Rule 110.} The computer is generally programmable, thus can simulate any four-unit system. 
\textbf{(A)} WXYZ is a substrate of four units with binary states that update synchronously at discrete intervals. Here, the system is shown in state 1101, also written as $WXyZ$. 
\textbf{(B)} The units W, X, Y, and Z follow the rule 110 of elementary cellular automata, as defined in the truth table. The left table contains all 16 possible states of the system at update $t$. The right contains the state of the system at $t + 1$ for every state at $t$. Since all state transitions in WXYZ's dynamics are deterministic, this truth table is also equivalent to WXYZ's Transition Probability Matrix (TPM). 
\textbf{(C)} $WXyZ$ does meet IIT's postulates for consciousness. By applying IIT's postulates to $WXyZ$ and its TPM, the system is revealed to split into two irreducible complexes ($\varphi_s(WZ) = 0.47$, $\varphi_s(y) = 0.28$). 
\textbf{(D)} The first unfolded cause--effect structure, $WX$, consists of 3 distinctions ($W$, $Z$, $WZ$) and 7 relations among distinctions. The second unfolded cause--effect structure, $y$, consists of 1 distinction ($y$) and 1 self-relation.}
\label{fig:WXyZ}
\end{figure}

\begin{figure}
\centering
\includegraphics[width=7.25in]{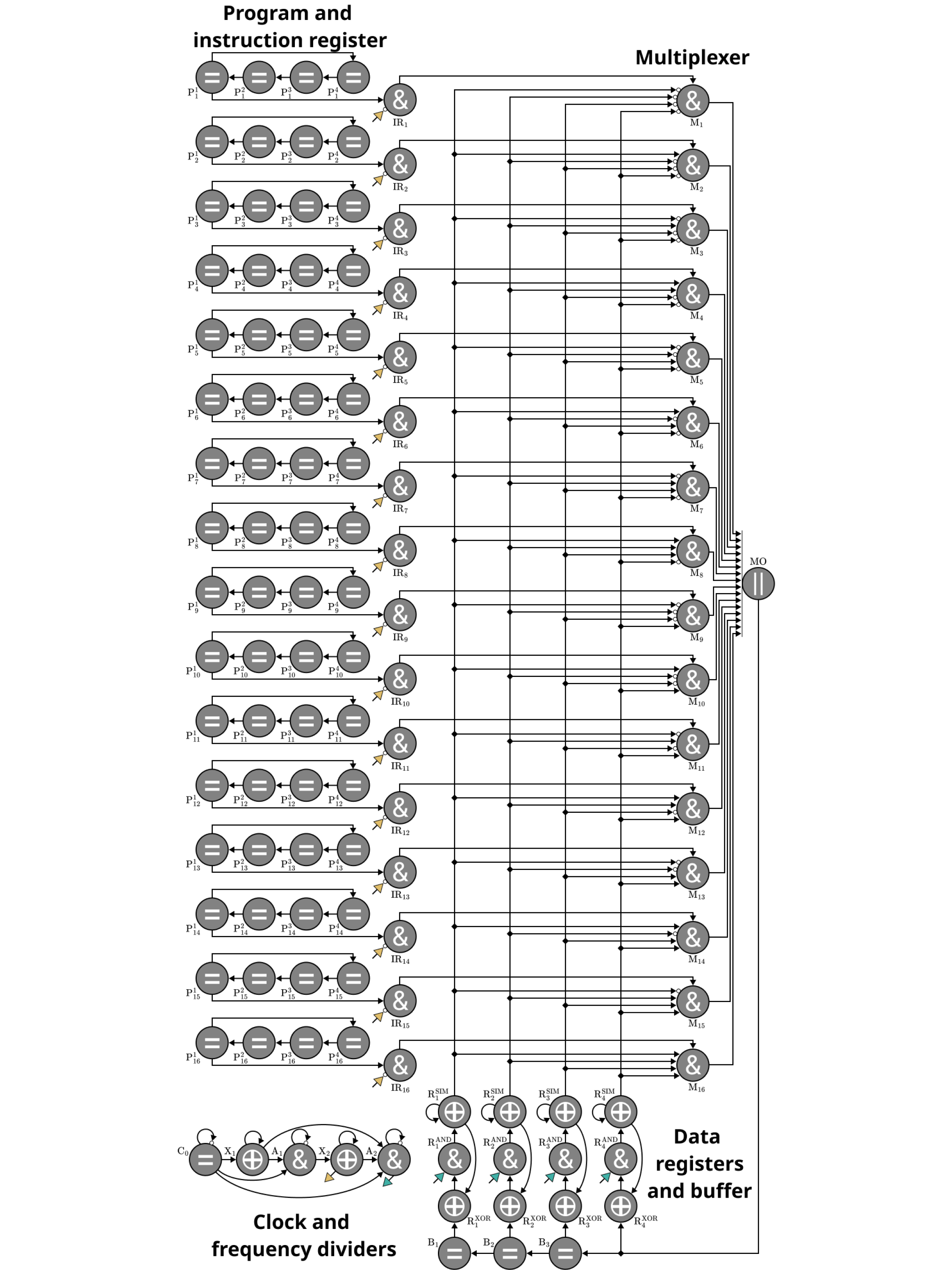}
\caption[\textbf{The four-bit computer with labeled units.}]{\textbf{The four-bit computer with labeled units.} Unit labels match those used in the supplementary code and proofs.}
\label{fig:computer_with_labeled_units}
\end{figure}

\begin{figure}
\centering
\includegraphics[width=3.6in]{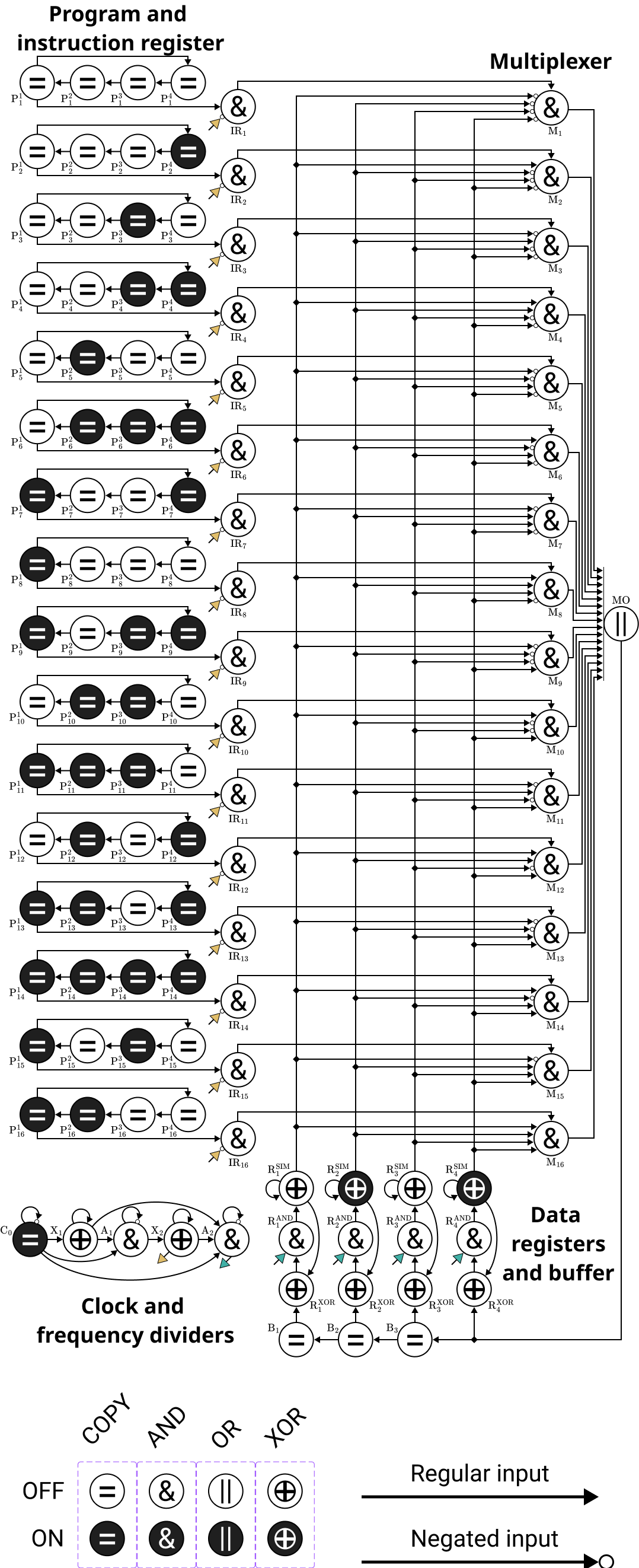}
\caption[\textbf{Update 0: Initialization.}]{\textbf{Update 0: Initialization.} First, the initial states of units $\REG{1}{SIM}$, $\REG{2}{SIM}$, $\REG{3}{SIM}$, and $\REG{4}{SIM}$ are set to reflect the state of the target PQRS (here, $0101$). Next, the state of the clock is set as shown. Note that $\CX{2}$ is 0, which will enable (because of the negated input) the instruction register to load values at the next update. Similarly, $\CA{2}$ is 0, which will disable the data registers from loading new values. Finally, the states of the computer's program units are set in such a way that they encode the target's state transition rules (see main text Fig. 2B).}
\label{fig:update0}
\end{figure}

\begin{figure}
\centering
\includegraphics[width=3.6in]{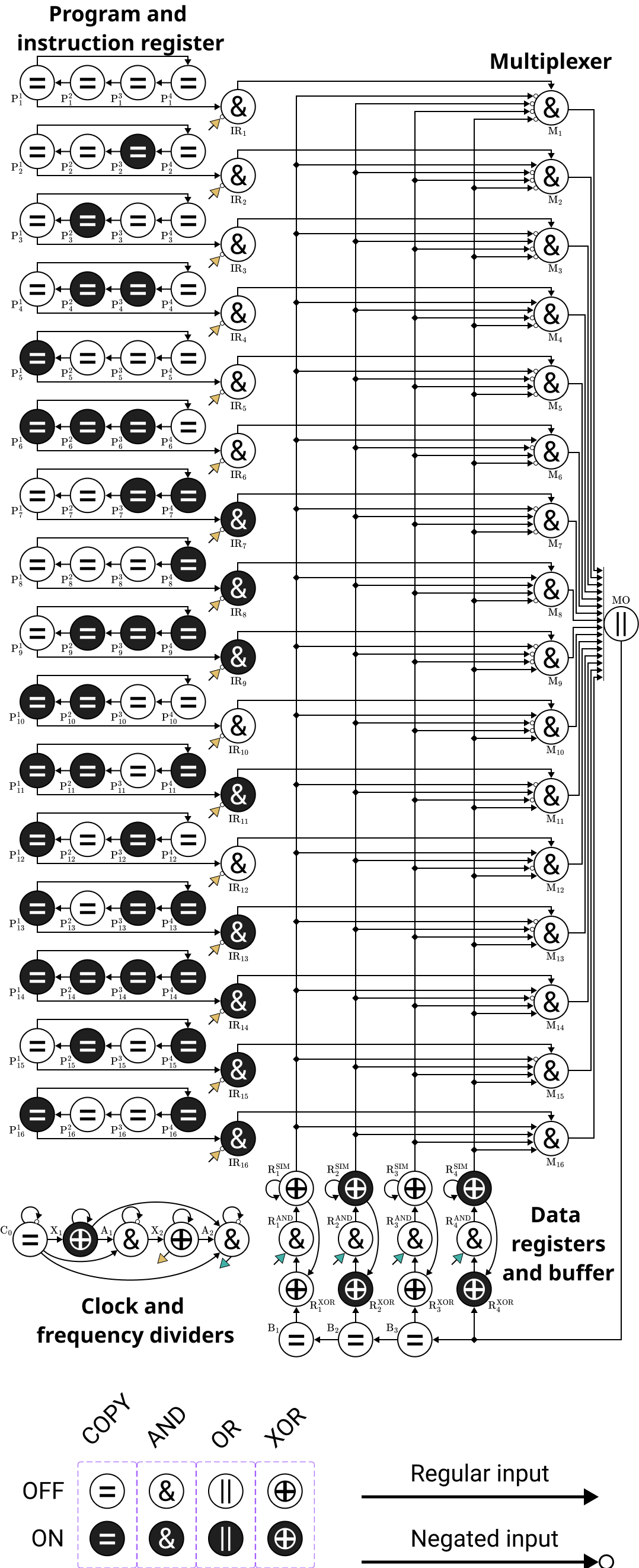}
\caption[\textbf{Update 1: The instruction register loads P's truth table, and current state selects a multiplexer input.}]{\textbf{Update 1: The instruction register loads P's truth table, and current state selects a multiplexer input.} At update 1, the computer is ready to compute the next state of P by using the current state of P to select a value from its truth table.}
\label{fig:update1}
\end{figure}

\begin{figure}
\centering
\includegraphics[width=3.6in]{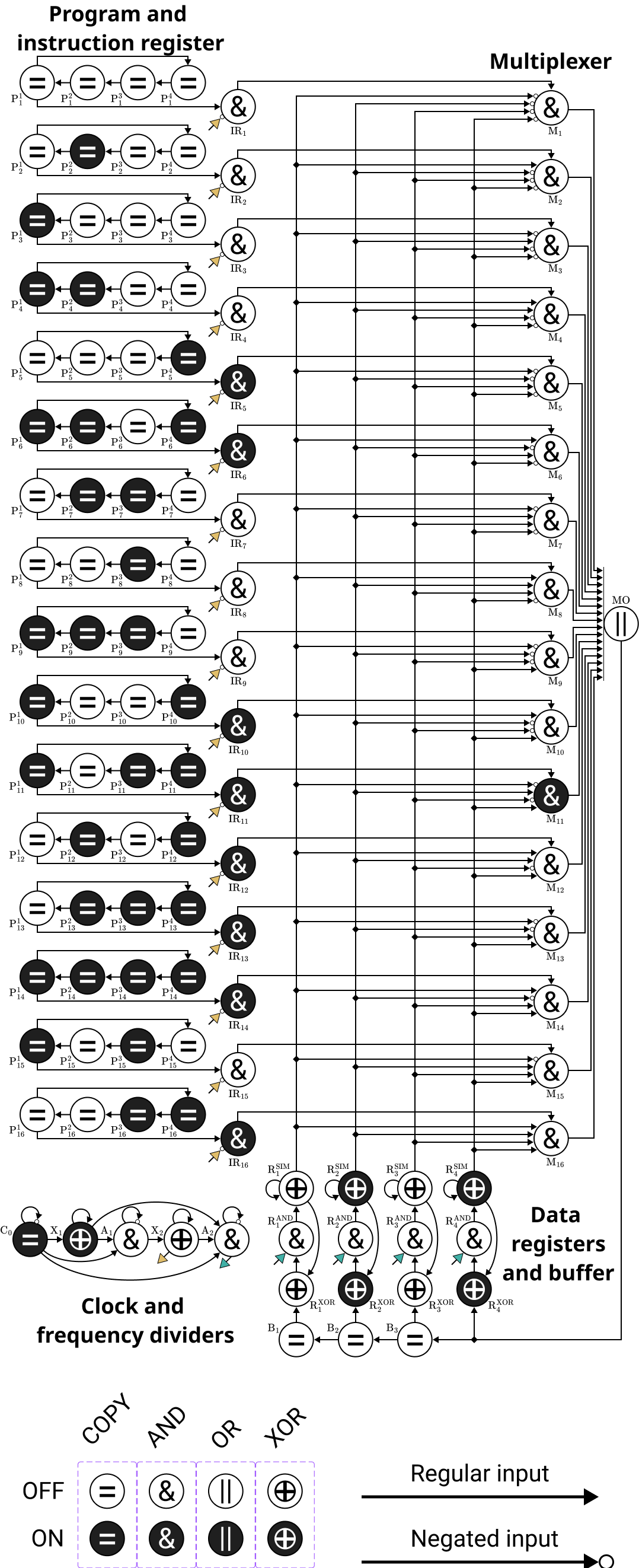}
\caption[\textbf{Update 2: The next state of P is computed.}]{\textbf{Update 2: The next state of P is computed.} The next state of P is computed, while Q's truth table and current state arrive at the multiplexer's inputs.}
\label{fig:update2}
\end{figure}

\begin{figure}
\centering
\includegraphics[width=3.6in]{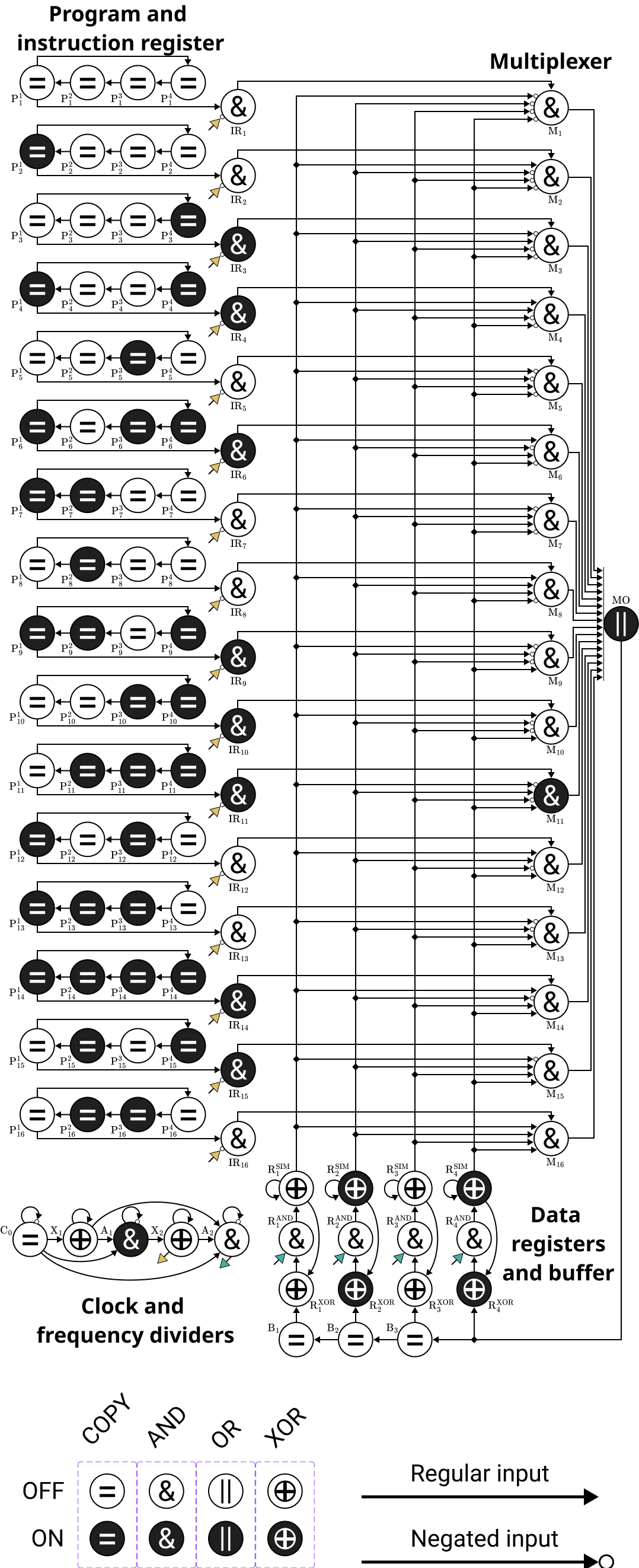}
\caption[\textbf{Update 3: The next state of Q is computed.}]{\textbf{Update 3: The next state of Q is computed.} The next state of P propagates towards the registers, the next state of Q is computed, while R's truth table and current state arrive at the multiplexer's inputs.}
\label{fig:update3}
\end{figure}

\begin{figure}
\centering
\includegraphics[width=3.6in]{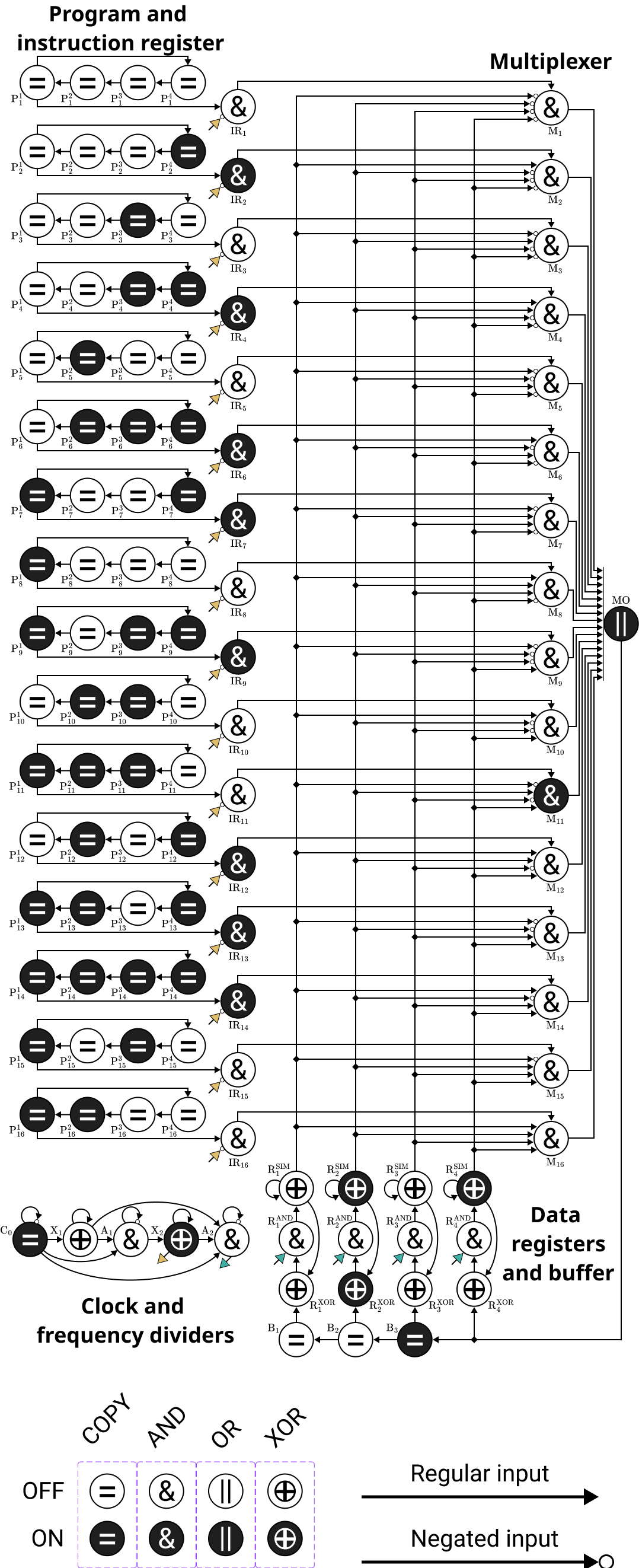}
\caption[\textbf{Update 4: The next state of R is computed.}]{\textbf{Update 4: The next state of R is computed.} P and Q's next state propagate towards the registers, the next state of R is computed, while S's truth table and current state arrive at the multiplexer's inputs.}
\label{fig:update4}
\end{figure}

\begin{figure}
\centering
\includegraphics[width=3.6in]{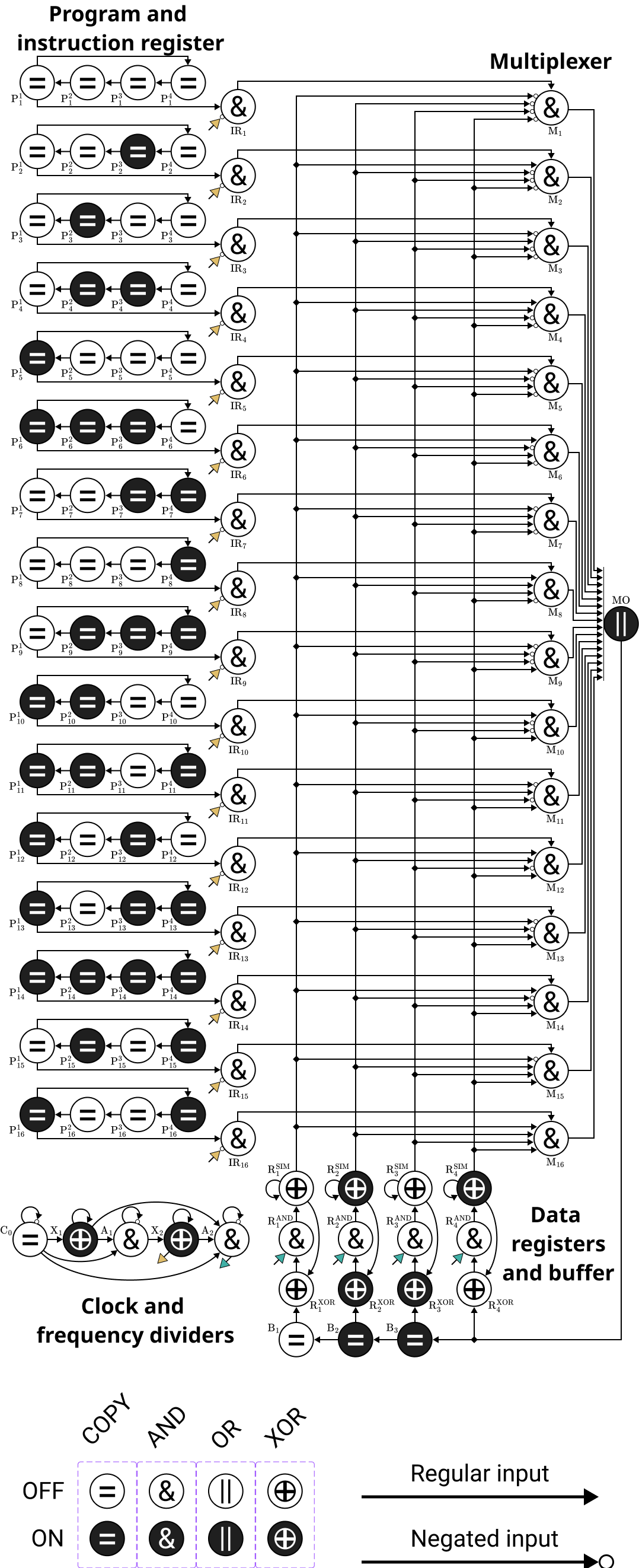}
\caption[\textbf{Update 5: The next state of S is computed.}]{\textbf{Update 5: The next state of S is computed.} P, Q, and R's next state propagate towards the registers, and the next state of S is computed. $\CX{2}$ is 1, which means that the instruction register will be cleared at the next update.}
\label{fig:update5}
\end{figure}

\begin{figure}
\centering
\includegraphics[width=3.6in]{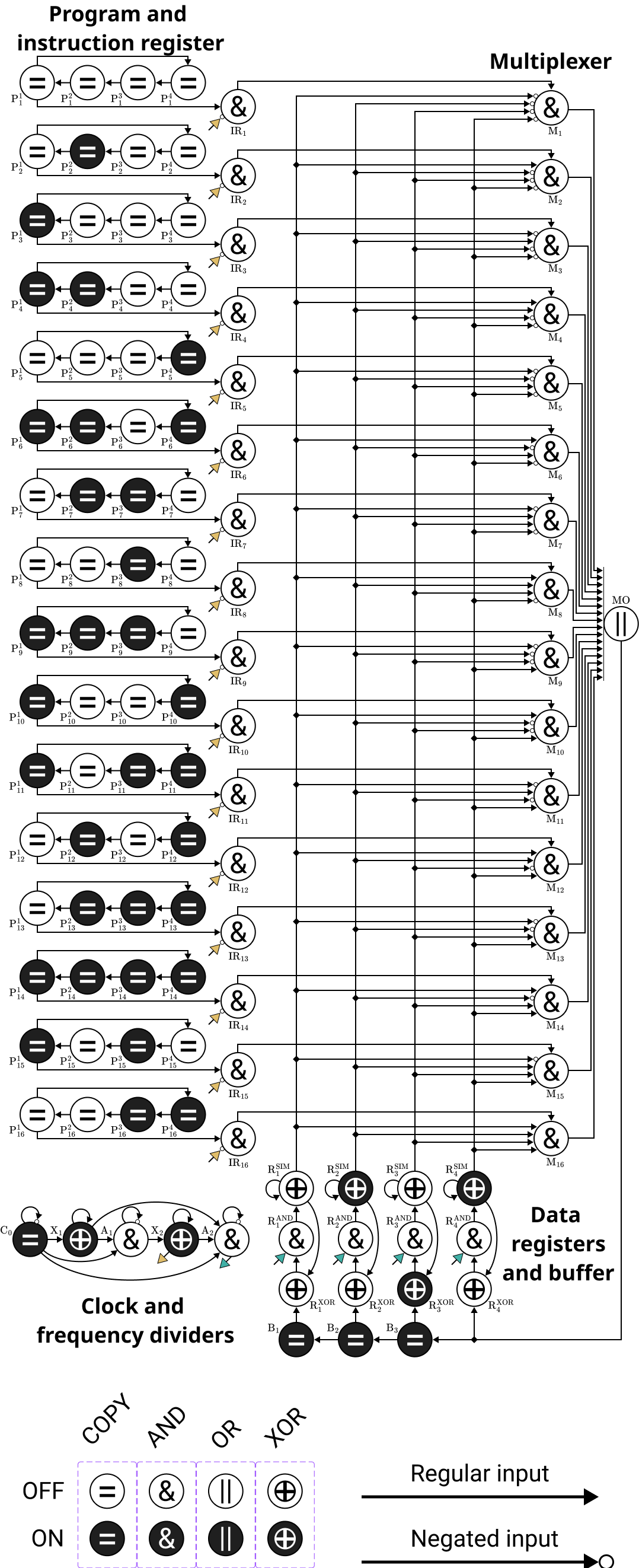}
\caption[\textbf{Update 6: Each simulated unit's next state arrives at its respective data register.}]{\textbf{Update 6: Each simulated unit's next state arrives at its respective data register.} P, Q, R, and S's next state arrive at their respective registers. The next state of each unit is compared with its current state, to determine whether or not the registers' outputs need to toggle.}
\label{fig:update6}
\end{figure}

\begin{figure}
\centering
\includegraphics[width=3.6in]{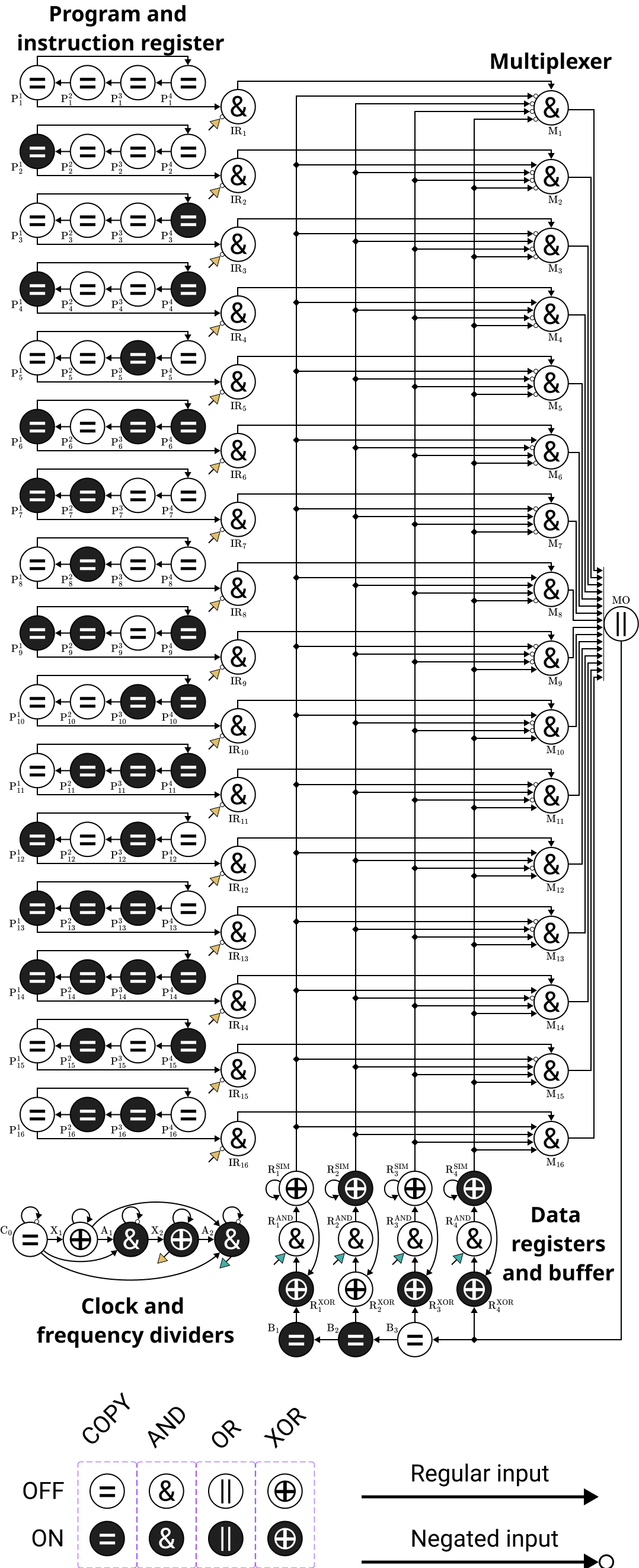}
\caption[\textbf{Update 7: The clock enables each register.}]{\textbf{Update 7: The clock enables each register.} Each data register computes a ``change'' signal (i.e., whether or not that register's output needs to toggle), while the clock enables each register so that this toggle signal can propagate towards the output at the next update.}
\label{fig:update7}
\end{figure}

\begin{figure}
\centering
\includegraphics[width=3.6in]{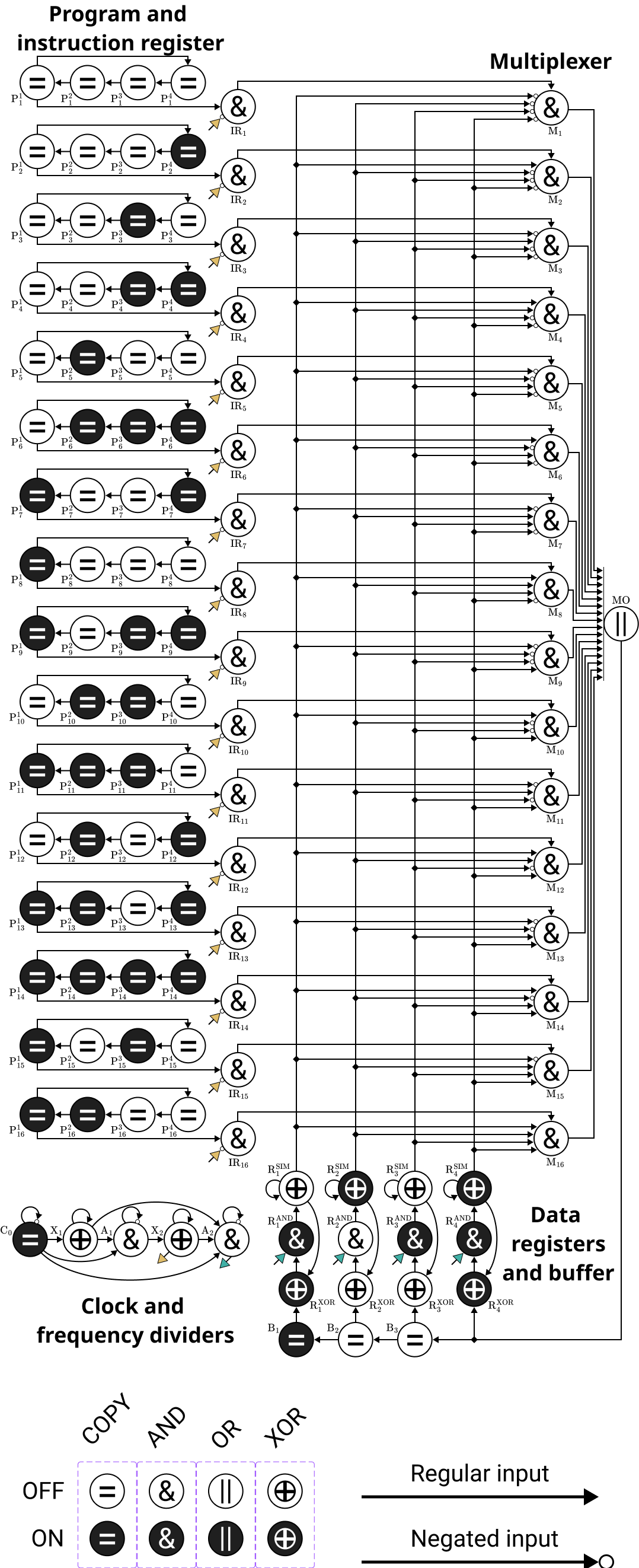}
\caption[\textbf{Update 8: Each register's toggle signal arrives at its output.}]{\textbf{Update 8: Each register's toggle signal arrives at its output.} Each register's AND unit reflects whether or not that register's output needs to toggle. This value is compared against the current value of the output, to determine what the next state of the output should be. The program has reset, and $\CX{2}$ enables the instruction register to begin loading from the program in anticipation of the next simulation iteration. }
\label{fig:update8}
\end{figure}

\begin{figure}
\centering
\includegraphics[width=3.6in]{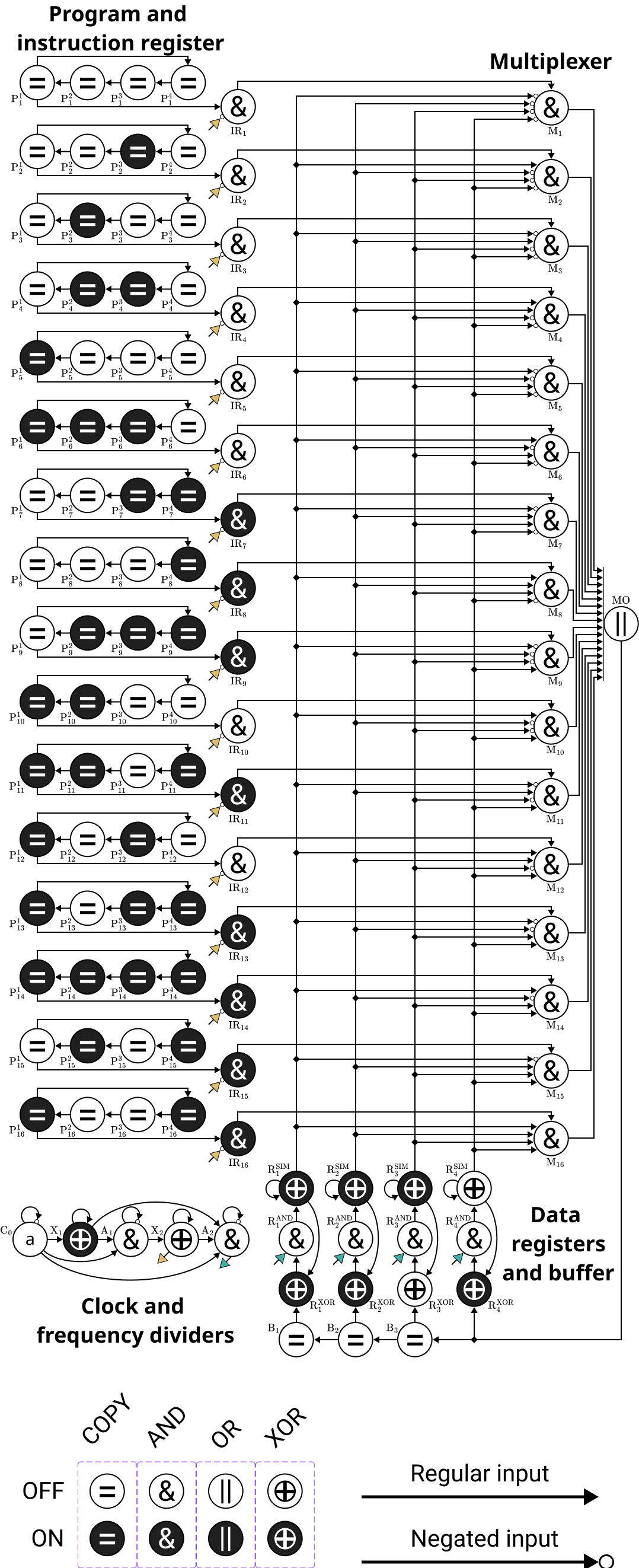}
\caption[\textbf{Update 9: The registers adopt the next state of PQRS, and the cycle repeats.}]{\textbf{Update 9: The registers adopt the next state of PQRS, and the cycle repeats.} $\REG{1}{SIM}$, $\REG{2}{SIM}$, $\REG{3}{SIM}$, and $\REG{4}{SIM}$ update synchronously to state 1110, the instruction register load's P's truth table, and the next simulation iteration begins.}
\label{fig:update9}
\end{figure}

\end{document}